\documentclass{article} %
\usepackage{iclr2027_conference,times}

\newif\ifarxiv
\arxivtrue %

\usepackage{fix-cm}       %
\usepackage{graphicx}     %
\usepackage{booktabs}     %
\usepackage{subcaption}
\usepackage{caption}
\usepackage{adjustbox}
\usepackage{dcolumn}
\usepackage{xspace}
\usepackage{multirow}
\usepackage{makecell}
\usepackage{tikz}
\usetikzlibrary{arrows.meta,backgrounds,calc,fit, positioning, shapes.geometric, decorations.pathmorphing, decorations.pathreplacing, external, spy, shapes.symbols}
\usepackage{algorithm}
\usepackage{algpseudocode}
\usepackage{xcolor}
\usepackage{colortbl}
\usepackage{pifont}
\usepackage[normalem]{ulem}   %
\usepackage{enumitem}
\usepackage{wrapfig}        %
\usepackage{listings}

\usepackage{amsmath,amsfonts,bm}

\def\eqref#1{equation~\ref{#1}}

\def\1{\bm{1}}

\DeclareMathAlphabet{\mathsfit}{\encodingdefault}{\sfdefault}{m}{sl}
\SetMathAlphabet{\mathsfit}{bold}{\encodingdefault}{\sfdefault}{bx}{n}

\newcommand{\R}{\mathbb{R}}

\usepackage{hyperref}
\usepackage{url}
\usepackage{cleveref}

\AtEndPreamble{%
  \crefname{proposition}{Prop.}{Props.}%
  \Crefname{proposition}{Prop.}{Props.}%
  \crefname{algorithm}{Alg.}{Algs.}%
  \Crefname{algorithm}{Alg.}{Algs.}%
  \crefname{figure}{Fig.}{Figs.}%
  \Crefname{figure}{Fig.}{Figs.}%
  \crefname{subfigure}{Fig.}{Figs.}%
  \Crefname{subfigure}{Fig.}{Figs.}%
  \crefname{table}{Tab.}{Tabs.}%
  \Crefname{table}{Tab.}{Tabs.}%
  \crefname{section}{Sec.}{Secs.}%
  \Crefname{section}{Sec.}{Secs.}%
  \crefname{subsection}{Sec.}{Secs.}%
  \Crefname{subsection}{Sec.}{Secs.}%
  \crefname{subsubsection}{Sec.}{Secs.}%
  \Crefname{subsubsection}{Sec.}{Secs.}%
  \crefname{equation}{Eq.}{Eqs.}%
  \Crefname{equation}{Eq.}{Eqs.}%
  \crefname{appendix}{App.}{Apps.}%
  \Crefname{appendix}{App.}{Apps.}%
}

\title{SoftCache: Adaptive Step Size Caching for Diffusion Transformers}

\let\cite\citep

\newcommand{\method}{SoftCache\xspace}

\newcommand{\ignorethis}[1]{}

\newcommand{\ie         }     {{i.e.}}

\newcommand{\Reals      }     {{\textrm{I\kern-0.18em R}}}

\newcommand{\change     } [1] {\mbox{{\footnotesize $\Delta$} \kern-3pt}#1}

\newlength{\x}

\definecolor{darkred}{rgb}{0.7,0.1,0.1}
\definecolor{darkgreen}{rgb}{0.1,0.6,0.1}
\definecolor{cyan}{rgb}{0.7,0.0,0.7}
\definecolor{otherblue}{rgb}{0.1,0.4,0.8}
\definecolor{maroon}{rgb}{0.76,.13,.28}
\definecolor{burntorange}{rgb}{0.81,.33,0}
\definecolor{cyan_real}{rgb}{0.0,1.0,1.0}
\definecolor{purple_real}{rgb}{0.5,0.0,0.5}
\definecolor{pixelblue}{HTML}{1B6898}   %
\definecolor{latentred}{HTML}{C91D22}   %
\definecolor{sleeperwood}{HTML}{A97F5B}
\definecolor{sleeperdark}{HTML}{7A5A3C}
\definecolor{heatfresh}{HTML}{C44E52}   %
\definecolor{heatcached}{HTML}{3D5A6C}  %

\ifdefined\ShowNotes
  \newcommand{\colornote}[3]{{\color{#1}\textbf{#2} #3\normalfont}}
\else
  \newcommand{\colornote}[3]{}
\fi

\newcommand\todosilent[1]{}

\newcommand{\probP}{\text{I\kern-0.15em P}}

\ifdefined\RevisedVersion

\else

\fi

\newcommand{\filmstriph}{4.5}%
\newcommand{\filmholew}{3.2pt}%
\newcommand{\filmholeh}{2.2pt}%
\newcommand{\filmnholes}{15}%
\newcommand{\filmstripcol}{black!85}%
\newcommand{\filmsprocket}[1]{%
  \edef\filmstripw{\the\dimexpr#1\relax}%
  \expandafter\filmsprocketat\expandafter{\filmstripw}}
\newcommand{\filmsprocketat}[1]{%
  \begin{tikzpicture}[x=#1,y=1pt,baseline=0pt]
    \fill[\filmstripcol] (0,0) rectangle (1,\filmstriph);
    \foreach \i in {1,...,\filmnholes}{%
      \node[fill=white,inner sep=0pt,rounded corners=0.35pt,
            minimum width=\filmholew,minimum height=\filmholeh]
        at ({(\i-0.5)/\filmnholes},{\filmstriph/2}) {};}%
  \end{tikzpicture}}

\newcommand{\gridprompt}[1]{{\usefont{T1}{ppl}{m}{it}\strut
  {\usefont{T1}{ptm}{m}{n}\textquotedblleft}#1{\usefont{T1}{ptm}{m}{n}\textquotedblright}}}

\providecommand{\zoominsetfrac}{0.42}%
\providecommand{\zoomcolor}{zoomcolor@default}%
\definecolor{zoomcolor@default}{HTML}{F2A900}%
\providecommand{\zoommarks}{}%
\definecolor{markcolor}{HTML}{D81B60}%
\tikzset{defectmark/.style={markcolor,line width=0.6pt}}%
\newcommand{\defectellipse}[4]{\draw[defectmark] (#1,{1-(#2)}) ellipse [x radius=#3, y radius=#4];}%
\newcommand{\marktile}[2]{\ifx\zoommarks\empty\includegraphics[width=#1]{#2}\else
  \begin{tikzpicture}
    \node[anchor=south west,inner sep=0] (img) at (0,0) {\includegraphics[width=#1]{#2}};
    \begin{scope}[x={(img.south east)},y={(img.north west)}]\zoommarks\end{scope}
  \end{tikzpicture}\fi}%
\makeatletter
\newlength{\zt@tilew}
\newlength{\zt@roih}
\newlength{\zt@insetw}
\newlength{\zt@inseth}
\newlength{\zt@bigw}
\newlength{\zt@margin}
\newlength{\zoompop}
\newlength{\zoompopx}
\newlength{\zoompopy}
\newlength{\zt@mx}
\newlength{\zt@my}
\newif\ifzt@pop

\newcommand{\zoomtile}[7]{%
  \edef\zt@tilewd{\the\dimexpr#1\relax}\setlength{\zt@tilew}{\zt@tilewd}%
  \begin{tikzpicture}
    \node[anchor=south west,inner sep=0] (img) at (0,0) {\includegraphics[width=\zt@tilew]{#2}};
    \ifdim\dimexpr\zoompopx+\zoompopy\relax>0pt
      \zt@poptrue \setlength{\zt@mx}{-\zoompopx}\setlength{\zt@my}{-\zoompopy}%
      \useasboundingbox (img.south west) rectangle (img.north east);
    \else
      \zt@popfalse \setlength{\zt@mx}{\zt@margin}\setlength{\zt@my}{\zt@margin}%
    \fi
    \begin{scope}[x={(img.south east)},y={(img.north west)}]
      \coordinate (roi-nw) at (#3,{1-(#4)});
      \coordinate (roi-ne) at ({(#3)+(#5)},{1-(#4)});
      \coordinate (roi-sw) at (#3,{1-(#4)-(#6)});
      \coordinate (roi-se) at ({(#3)+(#5)},{1-(#4)-(#6)});
    \end{scope}
    \pgfpointdiff{\pgfpointanchor{roi-sw}{center}}{\pgfpointanchor{roi-nw}{center}}%
    \edef\zt@roihd{\the\pgf@y}\setlength{\zt@roih}{\zt@roihd}%
    \pgfmathsetmacro{\ztmag}{\zoominsetfrac/(#5)}%
    \setlength{\zt@insetw}{\zoominsetfrac\zt@tilew}%
    \setlength{\zt@inseth}{\ztmag\zt@roih}%
    \setlength{\zt@bigw}{\ztmag\zt@tilew}%
    \edef\zt@corner{#7}%
    \def\zt@NW{north west}\def\zt@NE{north east}%
    \def\zt@SW{south west}\def\zt@SE{south east}%
    \ifx\zt@corner\zt@NW
      \coordinate (cw-nw) at ($(img.north west)+(\zt@mx,-\zt@my)$);
      \coordinate (cw-se) at ($(cw-nw)+(\zt@insetw,-\zt@inseth)$);
      \def\zt@roinear{roi-ne}\def\zt@cwnear{cw-se}%
    \else\ifx\zt@corner\zt@NE
      \coordinate (cw-ne) at ($(img.north east)+(-\zt@mx,-\zt@my)$);
      \coordinate (cw-nw) at ($(cw-ne)+(-\zt@insetw,0pt)$);
      \coordinate (cw-se) at ($(cw-nw)+(\zt@insetw,-\zt@inseth)$);
      \def\zt@roinear{roi-nw}\def\zt@cwnear{cw-sw}%
    \else\ifx\zt@corner\zt@SW
      \coordinate (cw-sw) at ($(img.south west)+(\zt@mx,\zt@my)$);
      \coordinate (cw-nw) at ($(cw-sw)+(0pt,\zt@inseth)$);
      \coordinate (cw-se) at ($(cw-nw)+(\zt@insetw,-\zt@inseth)$);
      \def\zt@roinear{roi-se}\def\zt@cwnear{cw-ne}%
    \else %
      \coordinate (cw-se) at ($(img.south east)+(-\zt@mx,\zt@my)$);
      \coordinate (cw-nw) at ($(cw-se)+(-\zt@insetw,\zt@inseth)$);
      \def\zt@roinear{roi-sw}\def\zt@cwnear{cw-nw}%
    \fi\fi\fi%
    \coordinate (cw-sw) at (cw-nw |- cw-se);
    \coordinate (cw-ne) at (cw-se |- cw-nw);
    \coordinate (roi-near) at (\zt@roinear);
    \coordinate (cw-near) at (\zt@cwnear);
    \coordinate (zt-vec) at ($(roi-nw)-(img.south west)$);
    \coordinate (zt-p2) at ($(cw-nw)-\ztmag*(zt-vec)$);
    \ifzt@pop\fill[white] (cw-nw) rectangle (cw-se);\fi
    \begin{scope}
      \clip (cw-nw) rectangle (cw-se);
      \node[anchor=south west,inner sep=0] at (zt-p2) {\includegraphics[width=\zt@bigw]{#2}};
    \end{scope}
    \draw[\zoomcolor,line width=0.5pt,opacity=0.7] (roi-near) -- (cw-near);
    \draw[white,line width=1.2pt] (roi-nw) rectangle (roi-se);
    \draw[\zoomcolor,line width=0.6pt] (roi-nw) rectangle (roi-se);
    \ifzt@pop %
      \draw[white,line width=1.6pt] (cw-nw) rectangle (cw-se);
      \draw[\zoomcolor,line width=0.9pt] (cw-nw) rectangle (cw-se);
    \else
      \draw[white,line width=1.2pt] (cw-nw) rectangle (cw-se);
      \draw[\zoomcolor,line width=0.6pt] (cw-nw) rectangle (cw-se);
    \fi
    \begin{scope}[x={(img.south east)},y={(img.north west)}]\zoommarks\end{scope}
  \end{tikzpicture}%
}
\makeatother

\algnewcommand{\IfThen}[2]{\State \algorithmicif\ #1\ \algorithmicthen\ #2}

\InputIfFileExists{local_config.tex}{}{} %

\hypersetup{colorlinks=true, linkcolor=blue, citecolor=blue, urlcolor=blue, breaklinks=true}

\author{Shimon Malnick$^{1}$, Nadav Orenstein$^{1}$, Ohad Fried$^{2}$, Shai Avidan$^{1}$ \\
$^{1}$Tel Aviv University \quad $^{2}$Reichman University
\ifarxiv \\ Project page: \href{https://malnick.net/softcache}{\texttt{malnick.net/softcache}}\fi
}

\ifarxiv
\iclrfinalcopy
\fi
\begin{document}

\maketitle
\ifarxiv\lhead{Preprint}\fi

\begin{abstract}
    Diffusion and flow models generate a sample over iterative steps, each a full
    pass through a large network, and that pass dominates generation cost.
    Training-free caching accelerates them by reusing activations from recent
    evaluations instead of recomputing them.
    Dynamic caching methods decide when to refresh by accumulating a cheap
    probe of the network input into a staleness signal and spending a fresh
    evaluation once it crosses a threshold. They perform \emph{hard caching}:
    the signal is a binary switch between refresh and reuse, and every step,
    fresh or cached, advances by the prescribed step size.
    We observe that the signal carries much more than this switch. How far it
    sits below the threshold, its headroom, predicts the error induced by reuse.
    We propose \method, which performs \emph{soft caching}: instead of a binary
    switch, the headroom sets how far each cached step goes.
    Where the headroom is large, the sampler jumps past several scheduled nodes, and as a refresh nears, its steps shrink back toward the original step.
    These adaptive jumps may land between the scheduled nodes, taking the sampler off its fixed grid.
    \method requires no training and adds no per-backbone settings: on FLUX for images and on
    HunyuanVideo and Wan2.1 for video, at speedups from $3.5\times$ to
    $6.9\times$, it is ahead of every baseline on all four fidelity
    metrics.\ifarxiv\else\footnote{Our code will be released publicly.}\fi
\end{abstract}

\section{Introduction}
Diffusion and flow models set the state of the art in image
and video synthesis~\citep{flux2024,kong2024hunyuanvideo,wan2025wan,esser2024scaling}.
These models are slow at inference: producing a single output requires
tens of full network evaluations, each a pass through the same large transformer.
Yet the latent changes only slightly between steps, so successive
passes largely recompute what the previous one already produced.
This redundancy is the resource that acceleration methods draw on, and they
differ in what they are willing to give up to reach it.
In practice, this is done using
distillation~\citep{pmlr-v202-song23a,salimans2022progressive}, fewer-step
solvers~\citep{song2021denoising,dpmsolver,zhao2023unipc}, or cheaper
attention~\citep{yuan2024ditfastattn,xi2025sparse,zhang2025fast,zhang2025spargeattn},
each with its own cost in training, calibration, or fidelity.

A different lever is to reuse the model's own output, \ie, \emph{caching},
instead of solving faster or attending more cheaply.
Successive steps can then share computation directly rather than recompute
it.
Every cache must still decide when to refresh, and many settle this before
sampling begins, with a refresh pattern that is fixed in advance and shared by every
prompt~\citep{ma2024deepcache,selvaraju2024fora,liu2025taylorseer,cui2026dpcache,aliev2026recache}.
\emph{Dynamic} caches instead make refresh timing their central question and
decide it during sampling: a cheap probe of the network predicts how much its
output is about to change, and this signal decides, per prompt, when the
cached computation has gone stale~\citep{liu2025teacache,chung2026seacache}.

Yet these dynamic caches spend the signal on a single binary decision, a
form of \emph{hard caching}: at every node of the sampler's fixed grid, the
signal is compared with a threshold, and the sampler either reuses the
cached computation or refreshes it.
How far below the threshold the signal sits plays no role.
A node where the cache has barely aged and one
on the verge of refresh are treated identically, and both advance by exactly
the step the schedule prescribed.

\begin{wrapfigure}[16]{r}{0.5\linewidth}
  \centering
  \definecolor{trajectorypurple}{HTML}{8172B2}
  \definecolor{vectorblue}{HTML}{0072B2}
  \adjustbox{max width=\linewidth}{%
    \input{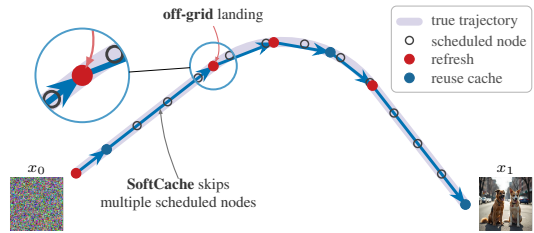}}
  \caption{\textbf{\method: how far, not just whether.}
  Hard caching walks the fixed grid of scheduled nodes one node
  at a time, deciding only whether to \textcolor{latentred}{refresh} or
  \textcolor{pixelblue}{reuse}. \method also sets the step size: a
  \textcolor{pixelblue}{reused} step the cache still supports skips several
  nodes and lands between them (inset).}
  \label{fig:teaser}%
  \tikzexternalenable
\end{wrapfigure}

We observe that this ignored distance is informative. Below the threshold,
the signal closely tracks the reuse error, the gap between the cached
computation and a fresh one at the current node. The
distance from the threshold, which we call the \emph{headroom}, measures
how safely the cache can be trusted, and we use it to set the step size
instead of only the refresh decision: this is \emph{soft caching}.
When the headroom is large, the cache is still reliable, so the sampler takes
a long step that skips several scheduled nodes and can land between them
(\cref{fig:teaser}). When the headroom is small, refresh is near, so the step
shrinks back toward the scheduled one. We call this method \method: the headroom sets the step continuously rather
than picking how many nodes to skip, and every node a step passes over is
one the sampler no longer visits.

Set against both kinds of cache, this is a different form of adaptivity. A
fixed schedule refreshes at the same nodes for every prompt. A dynamic
cache refreshes at nodes that vary with the prompt, but at scheduled nodes
all the same. Here the steps vary with the prompt too, and they also land
where no node was scheduled. This freedom translates into fidelity: on
FLUX, HunyuanVideo and Wan2.1, against caches of both kinds at comparable
speed, \method reproduces the uncached model most faithfully on every
metric at every operating point we test.
Adaptivity, then, does not stop at deciding when to refresh. It also
decides where the sampler goes next.

Our main contributions are:
(1) We identify that the refresh signal of a dynamic cache carries more than a
binary decision: its headroom below the threshold tracks the reuse error, so it
measures how far the cached computation can still be trusted.
(2) We propose \method, which reads the headroom as a step size: cached steps
jump off the scheduler's grid in proportion to it, the tail is re-spaced so
that every node passed is dropped from the trajectory, and the cached residual is
extrapolated linearly to the landing, at no extra network cost.
(3) On image and video backbones, \method adds no per-backbone settings
and is ahead of every cache we compare on all four fidelity metrics in
every speed band from $3.5\times$ to $6.9\times$.

\section{Related Work}
\label{sec:related_cache}

\subsection{Diffusion sampling acceleration}
\label{sec:related_cost}
Methods that cut sampling cost differ in what they modify. Some change the network or a
single evaluation of it: distillation trains a student that needs fewer
steps~\citep{pmlr-v202-song23a,salimans2022progressive,luo2023lcm}, while sparse
attention~\citep{yuan2024ditfastattn,xi2025sparse,zhang2025fast,zhang2025spargeattn} and token
reduction~\citep{bolya2023tome} make each step cheaper. Others change the grid: higher-order
solvers reach a given quality in fewer steps~\citep{song2021denoising,dpmsolver,zhao2023unipc},
and schedule design places the steps where they matter
most~\citep{karras2022elucidating,esser2024scaling,align_your_steps}.
We work in a third family, caching, which changes neither the network nor an evaluation
of it: it exploits the redundancy between neighboring states along the trajectory and
reuses computation from earlier steps, without training.

\subsection{Feature caching}
\label{sec:related_what_when}
Caching methods differ in what they store and when they refresh it. The stored
quantity ranges from intermediate activations of U-Net or transformer
blocks~\citep{ma2024deepcache,selvaraju2024fora,liu2025taylorseer} to the residual
of a span of blocks or of the entire
stack~\citep{chen2024deltadit,liu2025teacache,chung2026seacache}, and it may be
refreshed for only a subset of tokens~\citep{zou2025accelerating,zou2024DuCa,ClusCa}.
Fixed schedules refresh periodically~\citep{ma2024deepcache,selvaraju2024fora,liu2025taylorseer}
or at steps chosen offline, by thresholding a reuse error calibrated on a few
samples~\citep{ma2025magcache,peng2026ertacache} or by
optimizing for a fixed budget of full evaluations~\citep{cui2026dpcache,aliev2026recache}.
Online rules refresh once an accumulated cheap proxy of the output change crosses a
threshold~\citep{liu2025teacache,chung2026seacache,nam2026oricache,bu2025dicache}.
Such rules therefore refreshe at different nodes for different prompts,
whereas a fixed schedule applies one refresh pattern to all of them.
Either way, each node, or token, makes one binary choice between a full evaluation
and reuse. \method refreshes on the same spectral signal~\citep{chung2026seacache},
but also reads its headroom below the threshold to make reuse soft rather than binary.

\subsection{Predictive caching}
\label{sec:related_how}
Between refreshes, most caches hold the stored quantity
fixed~\citep{ma2024deepcache,selvaraju2024fora,liu2025teacache,chung2026seacache},
so the reuse error grows with the time since the last refresh. Predictive caches
extrapolate from recent history instead. Some forecast what the network returns,
the output of the block stack~\citep{cui2026dpcache},
its increment over the input~\citep{bu2025dicache}, or the velocity itself~\citep{wang2026zeus}.
Others forecast the increment each attention and MLP
branch adds to its input, which the current node applies to its own activations,
with Taylor~\citep{liu2025taylorseer} or Hermite~\citep{feng2026hicache}
coefficients built from several stored past values.
Multistep formulas~\citep{yu2025abcache,zheng2025foca} and a check on the
prediction~\citep{liu2025speca} extend how far it can be carried. All of them
predict only at the nodes of the sampler's schedule, so the schedule itself is
untouched. We extrapolate an increment as well, a single one for the entire stack,
along the straight line through the two most recent fresh evaluations. Written in
the flow time, that line is defined between nodes, which is what lets the sampler
leave the schedule rather than only skip evaluations along it.

\subsection{Adaptive step size control}
\label{sec:related_solvers}
Adaptive ODE solvers set each step at runtime from an estimate of the local error,
obtained by comparing two solutions of different order~\citep{dormand1980family},
and the same control has been applied to diffusion sampling in both its
SDE~\citep{jolicoeur2021gotta} and ODE~\citep{dpmsolver} forms.
The estimate costs several network evaluations per step, which is the computation a
cache exists to avoid. \method takes its step size from the refresh signal the cache
already computes, so the schedule adapts at no evaluations beyond those the cache spends
anyway.

\section{Preliminaries}
\label{sec:prelim}

Diffusion and flow models generate a sample by transporting noise to data
along a differential equation. In diffusion models this is a reverse-time SDE
or, equivalently, its probability-flow ODE, whose drift is built from a learned
score~\citep{song2021scorebased}, while in flow matching the network regresses the
transporting velocity directly~\citep{lipman2022flow,liu2022flow,albergo2022building}.
We present the method in the ODE form, which covers the rectified-flow
transformers we evaluate~\citep{esser2024scaling}. Writing $x_t\in\R^d$ for the state at time $t\in[0,1]$, with $x_0\sim p_0$ a sample
from the prior and $x_1$ the generated data sample, sampling integrates
\begin{equation}
    \frac{d x_t}{d t} = v_\theta(x_t, t),
    \qquad v_\theta:\R^d\times[0,1]\to\R^d,
    \label{eq:ode}
\end{equation}
from $t=0$ to $t=1$ over a grid $0 = t_0 < t_1 < \cdots < t_N = 1$ that the
scheduler fixes before sampling, independently of the prompt and of the
trajectory. With the explicit Euler solver, one step
reads
\begin{equation}
    x_{i+1} = x_i + \Delta_i\, v_\theta(x_i, t_i),
    \qquad \Delta_i := t_{i+1} - t_i ,
    \label{eq:euler}
\end{equation}
so a sample costs $N$ evaluations of $v_\theta$, one at every node of the grid.

\subsection{Residual caching}
\label{sec:cache}

A cache removes most of the cost of a step by reusing what an earlier evaluation
produced. Which part is reused follows from how a diffusion
transformer~\citep{peebles2023scalable,esser2024scaling,flux2024,kong2024hunyuanvideo}
computes $v_\theta$, in three stages. A patch embedding
$\mathcal{E}:\R^d\to\R^{n\times c}$ maps the latent to a sequence of $n$ tokens of
width $c$, which we write $s_i = \mathcal{E}(x_i)$. A stack of transformer blocks
$\mathcal{B}$ transforms these tokens, conditioned on the time embedding $e_i$,
and an output layer $\mathcal{O}$ normalizes and projects them back to $\R^d$:
\begin{equation}
    v_\theta(x_i, t_i) = \mathcal{O}\big(\mathcal{B}(s_i, e_i),\ e_i\big).
    \label{eq:factorization}
\end{equation}
The prompt embedding is fixed for the whole trajectory and enters both
$\mathcal{B}$ and $e_i$. We suppress it from the notation, since nothing in
what follows reads it.
Evaluating $\mathcal{B}$ accounts for nearly all of the cost of a model
evaluation, whereas $\mathcal{E}$ and $\mathcal{O}$ are comparatively cheap, so
it is $\mathcal{B}$ that a cache skips.

Let $a < i$ be the anchor, the most recent node at which the stack was
evaluated. What is stored there is not its output but the change it made to its
input,
\begin{equation}
    r_a = \mathcal{B}(s_a, e_a) - s_a ,
    \label{eq:residual}
\end{equation}
which we call the residual~\citep{liu2025teacache,chung2026seacache}. It varies
slowly along the trajectory, so a cached node skips $\mathcal{B}$ and adds the
anchor's residual to its own patched input,
\begin{equation}
    \widehat r_i = r_a ,
    \qquad
    \widehat v_i = \mathcal{O}\big(s_i + \widehat r_i,\ e_i\big),
    \label{eq:zoh}
\end{equation}
a zero-order hold. The patched input $s_i$ is recomputed from the current latent
and the output layer is evaluated at the current time through $e_i$, so what is
stale is the stack's own contribution, computed at the anchor's state and under
the anchor's time conditioning.

\subsection{Dynamic caching}
A dynamic cache decides when to refresh from the trajectory
itself~\citep{liu2025teacache}. It reads a cheap probe $I_i$ of the network at
every node, available before any block runs, and scores staleness by the relative
change of the probe between consecutive nodes,
\begin{equation}
    \rho_i =
    \frac{\|I_i - I_{i-1}\|_1}{\|I_{i-1}\|_1 + \epsilon}
    \;\geq\; 0 ,
    \label{eq:proxy}
\end{equation}
with $\epsilon > 0$ a small constant. The score is accumulated since the anchor,
\begin{equation}
    A_i = \sum_{j=a+1}^{i} \rho_j ,
    \label{eq:accumulator}
\end{equation}
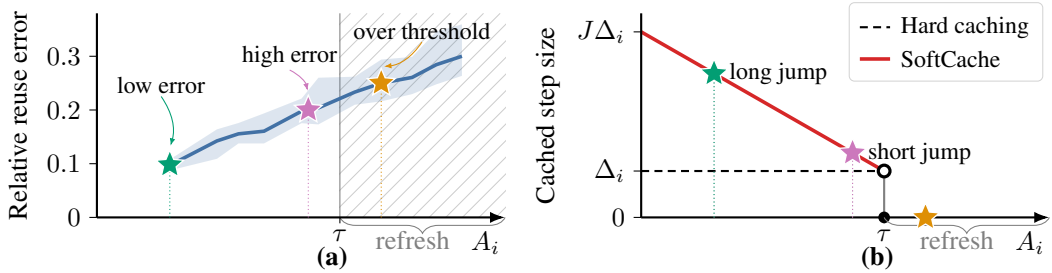
\begin{figure}[hb]
\setlength{\abovecaptionskip}{5pt}
\centering
\definecolor{driftblue}{HTML}{4271A5}
\definecolor{oursred}{HTML}{D62728}
\definecolor{ptgreen}{HTML}{029E73}
\definecolor{ptamber}{HTML}{DE8F05}
\definecolor{ptpurple}{HTML}{CC78BC}
\begin{subfigure}{0pt}\phantomsubcaption\label{fig:jump_size_a}\end{subfigure}%
\begin{subfigure}{0pt}\phantomsubcaption\label{fig:jump_size_b}\end{subfigure}%

\def\panelw{5.40}
\def\panelh{2.70}
\def\gap{1.80}
\def\tickgap{0.10}
\def\yaxisgapa{0.80}
\def\yaxisgapb{0.95}
\def\annotfontsize{\small}
\def\axisfontsize{\normalsize}

\pgfmathsetmacro{\xunit}{\panelw/1.68}
\pgfmathsetmacro{\yunita}{\panelh/0.38}
\pgfmathsetmacro{\yunitb}{\panelh/4.4}
\pgfmathsetmacro{\panelbx}{\panelw+\gap}
\pgfmathsetmacro{\capaY}{-1.3*(\yunitb/\yunita)}

\begin{tikzpicture}[
    >=Latex,
    every node/.style={font=\axisfontsize},
    tick/.style={thin},
    pointstar/.style={star, star points=5, star point ratio=2.2,
        minimum size=11pt, inner sep=0pt, draw=white, line width=0.5pt},
]

\def\refy{-0.05}
\def\refyleftdrop{-0.09}

\pgfmathsetmacro{\hstartmin}{\xunit-\panelh-0.3}
\pgfmathtruncatemacro{\hatchn}{ceil((\panelw-\hstartmin)/0.22)}
\begin{scope}
\clip (\xunit,0) rectangle (\panelw,\panelh);
\foreach \k in {0,...,\hatchn}{
    \draw[gray!40, line width=0.3pt]
        ({\hstartmin+\k*0.22},0) -- ({\hstartmin+\k*0.22+\panelh},\panelh);
}
\end{scope}

\begin{scope}[x=\xunit cm, y=\yunita cm]

\draw[gray] (1,0) -- (1,0.38);
\draw[gray, decorate, decoration={brace, amplitude=6pt, mirror}] (1,0) -- (1.68,0);
\node[gray, anchor=north, font=\axisfontsize, inner sep=1pt] (reflab) at (1.30,{(\refy+\refyleftdrop)/\yunita})
    {refresh};

\fill[driftblue, opacity=0.18]
    (0.2751,0.0825) -- (0.3223,0.0900) -- (0.4940,0.1091) -- (0.5827,0.1377) --
    (0.6877,0.1377) -- (0.8423,0.1754) -- (0.9115,0.1731) -- (1.0612,0.2097) --
    (1.1702,0.2160) -- (1.2972,0.2295) -- (1.3965,0.2477) -- (1.5023,0.2635) --
    (1.5023,0.3595) -- (1.3965,0.3459) -- (1.2972,0.2989) -- (1.1702,0.2939) --
    (1.0612,0.2620) -- (0.9115,0.2601) -- (0.8423,0.2213) -- (0.6877,0.1937) --
    (0.5827,0.1758) -- (0.4940,0.1640) -- (0.3223,0.1143) -- (0.2751,0.1022) --
    cycle;

\draw[driftblue, line width=1.35pt] plot coordinates {
    (0.2751,0.0933) (0.3223,0.1022) (0.4940,0.1424) (0.5827,0.1556)
    (0.6877,0.1605) (0.8423,0.1977) (0.9115,0.2039) (1.0612,0.2333)
    (1.1702,0.2500) (1.2972,0.2606) (1.3965,0.2843) (1.5023,0.3001)};

\draw[thick] (0,0) -- (0,0.38);
\draw[-{Latex[length=2mm]}, thick] (0,0) -- (1.68,0);

\foreach \v/\l in {1/\tau}{
    \draw[tick] (\v,0) -- (\v,-0.011);
    \node[below, inner sep=2pt] at (\v,-0.011) {$\l$};
}
\foreach \v/\l in {0/0,0.1/0.1,0.2/0.2,0.3/0.3}{
    \draw[tick] (0,\v) -- ({-\tickgap/\xunit},\v);
    \node[left, inner sep=3pt] at ({-\tickgap/\xunit},\v) {$\l$};
}

\node[below, inner sep=3pt] at (1.6,-0.011) {$A_i$};
\node[rotate=90, anchor=south] at ({-\yaxisgapa/\xunit},0.19) {Relative reuse error};
\node[anchor=south west, inner sep=3pt] at (0.87,\capaY) {\textbf{\normalsize{(a)}}};

\foreach \x/\y/\c/\n in {0.30/0.0980/ptgreen/g, 1.1702/0.2500/ptamber/m, 0.87/0.2002/ptpurple/p}{
    \draw[\c, thin, densely dotted] (\x,0) -- (\x,\y);
    \node[pointstar, fill=\c] (sa-\n) at (\x,\y) {};
}

\end{scope}

\node[font=\annotfontsize, anchor=south, inner sep=1pt] (la-g) at (0.85,1.60) {low error};
\draw[ptgreen, -{Latex[length=1mm]}, thin, shorten >=1pt] (la-g.south) to[bend left=25] (sa-g.north);

\node[font=\annotfontsize, anchor=south, inner sep=1pt] (la-m) at (4.3,2.32) {over threshold};
\draw[ptamber, -{Latex[length=1mm]}, thin, shorten >=1pt] (la-m.south) to[bend left=25] (sa-m.north);

\node[font=\annotfontsize, anchor=south, inner sep=1pt] (la-p) at (2.55,1.95) {high error};
\draw[ptpurple, -{Latex[length=1mm]}, thin, shorten >=1pt] (la-p.south) to[bend left=25] (sa-p.north);

\begin{scope}[x=\xunit cm, y=\yunitb cm, shift={(\panelbx cm,0)}]

\draw[oursred, line width=1.35pt] (0,4) -- (1,1);

\draw[thick, densely dashed] (0,1) -- (1,1);

\draw[thick] (0,0) -- (0,4.4);
\draw[-{Latex[length=2mm]}, thick] (0,0) -- (1.68,0);

\draw[gray, thick] (1,1) -- (1,0);
\fill[white] (1,1) circle (2.2pt);
\draw[very thick] (1,1) circle (2.2pt);
\fill (1,0) circle (2.2pt);

\foreach \v/\l in {1/\tau}{
    \draw[tick] (\v,0) -- (\v,-0.126);
    \node[below, inner sep=2pt] at (\v,-0.126) {$\l$};
}
\foreach \v/\l in {0/0,1/\Delta_i,4/J\Delta_i}{
    \draw[tick] (0,\v) -- ({-\tickgap/\xunit},\v);
    \node[left, inner sep=3pt] at ({-\tickgap/\xunit},\v) {$\l$};
}

\node[below, inner sep=3pt] at (1.6,-0.126) {$A_i$};
\node[rotate=90, anchor=south] at ({-\yaxisgapb/\xunit},2.2) {Cached step size};
\node[anchor=south west, inner sep=3pt] at (0.87,-1.3) {\textbf{\normalsize{(b)}}};

\draw[gray, decorate, decoration={brace, amplitude=6pt, mirror}] (1,0) -- (1.68,0);
\node[gray, anchor=north, font=\axisfontsize] at (1.30,{\refy/\yunitb}) {refresh};

\draw[gray!50, thin, densely dotted] (0.30,0) -- (0.30,3.10);
\draw[gray!50, thin, densely dotted] (0.87,0) -- (0.87,1.39);

\foreach \x/\y/\c/\n in {0.30/3.10/ptgreen/g, 1.1702/0/ptamber/m, 0.87/1.39/ptpurple/p}{
    \draw[\c, thin, densely dotted] (\x,0) -- (\x,\y);
    \node[pointstar, fill=\c] (sb-\n) at (\x,\y) {};
}
\node[font=\annotfontsize, anchor=west, inner sep=1pt] at ([xshift=1pt]sb-g.east) {long jump};
\node[font=\annotfontsize, anchor=west, inner sep=1pt] at ([xshift=1pt]sb-p.east) {short jump};

\end{scope}

\def\legleftx{0.90}
\def\legtopy{4.10}
\def\legrowgap{0.42}
\def\leglinelen{0.40}
\def\leggap{0.10}
\def\legpad{4pt}
\pgfmathsetmacro{\legLx}{\panelbx+\legleftx*\xunit}
\pgfmathsetmacro{\legTy}{\legtopy*\yunitb}

\coordinate (legline1a) at (\legLx,\legTy);
\coordinate (legline1b) at ({\legLx+\leglinelen},\legTy);
\node[anchor=west, font=\annotfontsize, inner sep=1pt] (legtxt1) at ({\legLx+\leglinelen+\leggap},\legTy) {Hard caching};
\coordinate (legline2a) at (\legLx,{\legTy-\legrowgap});
\coordinate (legline2b) at ({\legLx+\leglinelen},{\legTy-\legrowgap});
\node[anchor=west, font=\annotfontsize, inner sep=1pt] (legtxt2) at ({\legLx+\leglinelen+\leggap},{\legTy-\legrowgap}) {\method};
\begin{pgfonlayer}{background}
\node[draw=gray!35, fill=white, fill opacity=0.85, rounded corners=2pt, line width=0.35pt,
      fit=(legtxt1)(legtxt2)(legline1a)(legline2a), inner sep=\legpad] {};
\end{pgfonlayer}
\draw[thick, densely dashed] (legline1a) -- (legline1b);
\draw[oursred, line width=1.35pt] (legline2a) -- (legline2b);

\end{tikzpicture}

\caption{\textbf{The refresh signal carries more than a binary decision.}
\textbf{(a)} On uncached FLUX trajectories, the error $E_i$ of reusing the cached
residual (binned median, interquartile band) grows steadily with the refresh signal $A_i$.
\textbf{(b)} Hard caching only compares $A_i$ with $\tau$ and takes the scheduled step
$\Delta_i$ at every cached node. \textcolor{oursred}{\method} also reads how far below
$\tau$ it sits: the lower $A_i$, the longer the step, up to $J\Delta_i$ (\cref{eq:jump}).
Stars mark the same $A_i$ in both panels: a \textcolor{ptgreen}{low error buys a long jump},
a \textcolor{ptpurple}{high one only a short jump}, and past $\tau$
\textcolor{ptamber}{the node refreshes}.}
\label{fig:jump_size}
\end{figure}

so $A_i$ is nonnegative and nondecreasing until the next refresh, and the node is
cached while it stays below a threshold $\tau$. 
Rules of this family differ in how the probe is prepared.
The timestep-modulated input at the entry of the stack tracks
how much the output is about to change, and is read either unfiltered, with
$\rho_i$ rescaled through a polynomial fitted per
backbone~\citep{liu2025teacache}, or through a closed-form spectral low-pass that
needs no such fit~\citep{chung2026seacache}. We use the filtered probe of SeaCache
throughout. \method assumes nothing of $\rho_i$ beyond $\rho_i \geq 0$ and the reset
of $A_i$ at a refresh, so TeaCache's probe can replace it, at lower fidelity
(\cref{app:probe}).

\section{Method}
\label{sec:method}

\subsection{Headroom as a step size}
\label{sec:jump}

Hard caching spends $A_i$ on one decision: compare it with $\tau$ and
either refresh or reuse~\citep{liu2025teacache,chung2026seacache}.
We ask whether there is more to this signal, not \emph{whether} the cached residual can
still be used, but for \emph{how long}.
Instead of reading $A_i$ against $\tau$ as a hard threshold, we read how far below
it the accumulator sits, which says how much of the reuse budget is left, and we let
that quantity set the length of the next step. We define the \emph{headroom}
\begin{equation}
    H_i = 1 - \frac{A_i}{\tau},
    \label{eq:headroom}
\end{equation}
which lies in $[0,1]$ at a cached node.
Reusing the anchor's residual costs a relative reuse error
$E_i = \|r_i - r_a\| / \|r_a\|$, between the residual the stack would produce
at node $i$ and the one \cref{eq:zoh} substitutes for it. We measure it on
uncached FLUX trajectories, where $r_i$ is available at every node
(\cref{fig:jump_size_a}): the error rises as the accumulator fills, and a node
just past a refresh differs from one about to trigger the next by $2.2\times$
in median error, although the threshold treats the two alike
(\cref{app:headroom_drift}).
The measurement says which nodes can carry the residual further, not how much
further, so any rule that takes longer steps where the error is lower is
consistent with it. We choose the simplest one.
Freezing the same enlarged steps at even, back-loaded or random positions instead does worse at
the slower operating point (\cref{tab:ablation_placement}).

With $\Delta_i$ the scheduled spacing at node $i$ (\cref{eq:euler}) and $J > 1$
a cap on how many scheduled steps a single step may span, we set the step from
the headroom:
\begin{equation}
    \bar\Delta_i =
    \begin{cases}
        \Delta_i, & A_i \geq \tau \quad\text{(fresh)} \\
        \min\bigl\{\,1 - t_i,\;\; \Delta_i + (J - 1)\,H_i\,\Delta_i\,\bigr\}, & A_i < \tau \quad\text{(cached).}
    \end{cases}
    \label{eq:jump}
\end{equation}
At a cached node the step exceeds the scheduled one in proportion to the
headroom (\cref{fig:jump_size_b}): approaching $\Delta_i$ as a refresh nears,
at most $J\Delta_i$, and never past the end of the schedule.
After a jump, the probe change is measured from the node the step left,
so a long step fills the accumulator more than a short one.

We disable the jump in two cases, at the start of the trajectory and at a node where the probe
change is itself large, and the cached node then takes the scheduled step $\Delta_i$. Both are
thresholds on quantities the sampler already holds, with the same values in every setting
(\cref{app:gate}).

\subsection{Sampling between nodes}
\label{sec:offgrid}

A jump ends at a time the grid does not contain, and the node the sampler lands
on is one no schedule placed. Like every cached node it reuses $r_a$,
and the further a node sits from the anchor the worse that reuse gets. A jump
puts it further than any scheduled step does. Instead of holding $r_a$, we
extend it along the line through the two most recent fresh residuals, the
\emph{residual motion} of the cache:
\begin{equation}
    \widehat r_i = r_a + \beta\,\lambda_i\,(r_a - r_{a-1}),
    \qquad
    \lambda_i = \frac{t_i - t_a}{t_a - t_{a-1}} ,
    \label{eq:residual_motion}
\end{equation}
a straight line through the last two anchors, followed a fraction $\beta$ of
the way. The ratio $\lambda_i$ is built from time gaps, so it is defined
wherever the sampler is, on the grid or off it. Predictive caches that count
steps instead have a residual one node later, two nodes later, and nothing in
between. We apply \cref{eq:residual_motion} at every cached node, whether or
not the node was reached by a jump. It costs one more stored tensor, $r_{a-1}$,
and no evaluation.

The nodes ahead of the landing also need attention. The jump has passed several
of them, and the first it has not passed may be a fraction of a scheduled step
away, so following the old grid would spend a full node on a fraction of a
step. The time the step ended at becomes $t_{i+1}$, and the nodes after
it are replaced. Writing $m_i$ for the number of nodes of the old grid still
ahead of $t_{i+1}$, we spread that many nodes evenly over $[t_{i+1}, 1]$,
counting $t_{i+1}$ as the first of them,
\begin{equation}
    t_{i+1+k} = t_{i+1} + \left(1 - t_{i+1}\right)\frac{k}{m_i - 1},
    \qquad k = 1, \ldots, m_i - 1 .
    \label{eq:tail_regrid}
\end{equation}

\begin{wrapfigure}[15]{r}{0.5\linewidth}
  \centering
  \includegraphics[width=\linewidth]{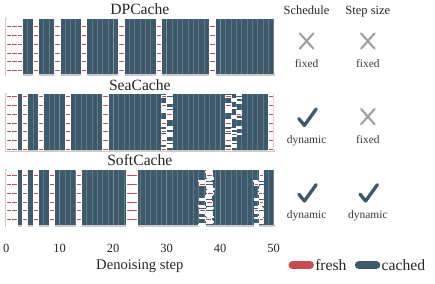}
  \caption{\textbf{Step placement on FLUX.}
  \textcolor{heatfresh}{\textbf{Fresh}} and \textcolor{heatcached}{\textbf{cached}}
  steps for 25 prompts, one per row.}
  \label{fig:schedule_heatmap}
\end{wrapfigure}

Sampling continues from $(x_{i+1}, t_{i+1})$ along this grid, which ends at
$t_{i+m_i} = 1$, and the construction repeats at every jump.
If $t = 1$ is the only node left ahead, the tail is already a single step and is kept.
The rebuilt tail is evenly spaced in $t$ since that choice has nothing to tune.
Re-spacing in the backbone's own coordinate instead changes little, and the
sign of the difference varies with speed (\cref{app:regrid_geometry}).

Until the first refresh there is a single anchor and no line to draw, so the
node holds $r_a$ as in \cref{eq:zoh}.
Every quantity in this section is fixed across all configurations, except $\beta$,
which is $1$ below $4\times$ and damped to $\tfrac{1}{2}$ above it~(\cref{sec:ablations}).

Collecting the pieces, a node is either fresh or cached, and the accumulator
decides which. A fresh node evaluates the stack and takes the scheduled step. A
cached node extrapolates the residual and takes a step set by its headroom. The
update is
\begin{equation}
    x_{i+1} = x_i + \bar\Delta_i\,\mathcal{O}\big(s_i + \widehat r_i,\ e_i\big).
    \label{eq:step}
\end{equation}
Two settings of \cref{eq:step} are known samplers. With $\beta = 0$ and
$\bar\Delta_i = \Delta_i$ at every node it is the hold of \cref{eq:zoh} on the
scheduled grid~\citep{chung2026seacache}. With every node fresh it is
\cref{eq:euler}. The loop, including its initialization, is in
\cref{alg:method}.

\expandafter\def\csname qfm@ours\endcsname{5.44}
\expandafter\def\csname qfm@seacache\endcsname{5.28}
\expandafter\def\csname qfm@dpcache\endcsname{4.74}
\expandafter\def\csname qfm@teacache\endcsname{5.28}
\expandafter\def\csname qfm@taylorseer\endcsname{5.59}

\newif\ifqualwrap
\qualwrapfalse
\ifqualwrap
  \newcommand{\qualpanelw}{0.115}%
\else
  \newcommand{\qualpanelw}{0.155}%
\fi
\ifqualwrap\begin{wrapfigure}{R}{0.6\linewidth}
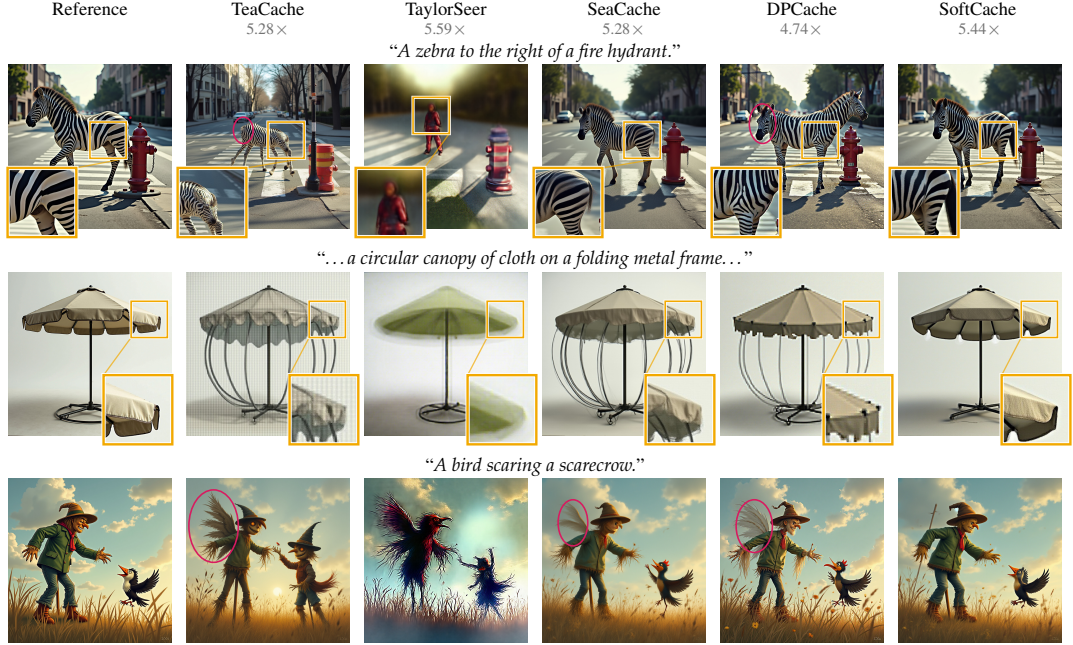
\vspace{0pt}\else\begin{figure}[ht]\fi
  \centering
  \def\gpanelw{\qualpanelw}%
  \newcommand{\gpromptfont}{\scriptsize}%
  \newcommand{\gpromptsep}{2pt}%
  \newcommand{\glabsize}{\scriptsize}%
  \newcommand{\gheadline}{0.5pt}%
  \newcommand{\gcolsep}{2.5pt}%
  \newcommand{\gheadsep}{10pt}%
  \newcommand{\growsep}{7.5pt}%
  \renewcommand{\zoominsetfrac}{0.42}%
  \setlength{\zoompop}{3pt}%
  \setlength{\tabcolsep}{\gcolsep}%
  \renewcommand{\arraystretch}{0}%
  \newcommand{\gzoomon}[2]{\expandafter\def\csname gzoomon@#1\endcsname{#2}}%
  \newcommand{\gzoomroi}[6]{\expandafter\def\csname gzoomroi@#1\endcsname{%
    \def\ztx{#2}\def\zty{#3}\def\ztw{#4}\def\zth{#5}\def\ztc{#6}}}%
  \newcommand{\gzoomat}[4]{\expandafter\def\csname gzoomat@#1@#2\endcsname{\def\ztx{#3}\def\zty{#4}}}%
  \gzoomon{drawbench_133}{1}%
  \gzoomroi{drawbench_133}{0.5}{0.35}{0.22}{0.22}{south west}
  \gzoomat{drawbench_133}{taylorseer}{0.3}{0.2}%
  \newcommand{\gmark}[6]{\expandafter\def\csname gmark@#1@#2\endcsname{\defectellipse{#3}{#4}{#5}{#6}}}%
  \gmark{drawbench_133}{teacache}{0.35}{0.40}{0.06}{0.08}%
  \gmark{drawbench_133}{dpcache}{0.26}{0.36}{0.08}{0.12}%
  \gzoomon{drawbench_079}{1}%
  \gzoomroi{drawbench_079}{0.75}{0.18}{0.22}{0.22}{south east}
  \gzoomon{drawbench_027}{0}%
  \gmark{drawbench_027}{teacache}{0.17}{0.29}{0.15}{0.22}%
  \gmark{drawbench_027}{seacache}{0.19}{0.27}{0.09}{0.14}%
  \gmark{drawbench_027}{dpcache}{0.21}{0.29}{0.11}{0.16}%
  \newcommand{\gcell}[2]{\def\garm{#2}\def\gref{reference}\def\gzoff{0}%
    \ifcsname gzoomon@#1\endcsname\edef\gzon{\csname gzoomon@#1\endcsname}\else\def\gzon{0}\fi
    \ifcsname gmark@#1@#2\endcsname\expandafter\let\expandafter\zoommarks\csname gmark@#1@#2\endcsname\else\def\zoommarks{}\fi
    \ifx\gzon\gzoff
      \marktile{\gpanelw\linewidth}{figures/qual_flux_main/#1/#2.jpg}%
    \else
      \csname gzoomroi@#1\endcsname
      \ifcsname gzoomat@#1@#2\endcsname\csname gzoomat@#1@#2\endcsname\fi
      \setlength{\zoompopy}{\zoompop}%
      \ifx\garm\gref\setlength{\zoompopx}{0pt}\else\setlength{\zoompopx}{\zoompop}\fi
      \zoomtile{\gpanelw\linewidth}{figures/qual_flux_main/#1/#2.jpg}{\ztx}{\zty}{\ztw}{\zth}{\ztc}%
    \fi}%
  \newcommand{\gprompt}[1]{\multicolumn{6}{@{}c@{}}{\parbox{\dimexpr6\dimexpr\gpanelw\linewidth\relax+10\dimexpr\gcolsep\relax\relax}{\centering\gpromptfont\gridprompt{#1}}} \\[\gpromptsep]}%
  \newcommand{\ghbox}[1]{\hbox to \gpanelw\linewidth{\hss#1\hss}}%
  \newcommand{\glabref}[1]{\vtop{\ghbox{\glabsize #1}}}%
  \newcommand{\glab}[2]{\vtop{\ghbox{\glabsize\strut #1}\kern\gheadline\ghbox{\tiny\color{black!55}\csname qfm@#2\endcsname$\times$}}}%
  \newcommand{\grow}[1]{%
    \gcell{#1}{reference} & \gcell{#1}{teacache} & \gcell{#1}{taylorseer} &
    \gcell{#1}{seacache} & \gcell{#1}{dpcache} & \gcell{#1}{ours}}%
  \ifdefined\tikzexternaldisable\tikzexternaldisable\fi
  \begin{tabular}{@{}cccccc@{}}
    \glabref{Reference} & \glab{TeaCache}{teacache} & \glab{TaylorSeer}{taylorseer} &
    \glab{SeaCache}{seacache} & \glab{DPCache}{dpcache} & \glab{\method}{ours} \\[\gheadsep]
    \gprompt{A zebra to the right of a fire hydrant.}
    \grow{drawbench_133} \\[\growsep]
    \gprompt{\dots a circular canopy of cloth on a folding metal frame\dots}
    \grow{drawbench_079} \\[\growsep]
    \gprompt{A bird scaring a scarecrow.}
    \grow{drawbench_027} \\
  \end{tabular}
  \ifdefined\tikzexternalenable\tikzexternalenable\fi
  \caption{\textbf{FLUX on DrawBench at $\approx\!5.2\times$ speedup.} The speedup under each
  method is its band mean. Red ellipses mark structural errors, and each inset enlarges one region
  of its tile where shown.}
  \label{fig:qual_flux_main}
\ifqualwrap\end{wrapfigure}\else\end{figure}\fi

\Cref{fig:schedule_heatmap} shows the schedules this produces on FLUX.
DPCache~\citep{cui2026dpcache} fixes its refresh steps before sampling, so
every prompt follows the same schedule. SeaCache~\citep{chung2026seacache}
refreshes dynamically, at different steps for different prompts, but each
step it takes is one scheduled interval. \method is dynamic in step size as
well: the headroom stretches each cached step (\cref{eq:jump}), and the
re-spaced tail (\cref{eq:tail_regrid}) moves the nodes after it, so its steps
vary in length and fall between grid lines.

\section{Experiments}
\label{sec:experiments}

\subsection{Experimental setup}
\label{sec:setup}

\newif\iffdwrap
\fdwraptrue
\iffdwrap
  \newcommand{\fdsize}{\small}%
  \newcommand{\fdcolsep}{2.5pt}%
\else
  \newcommand{\fdsize}{\small}%
  \newcommand{\fdcolsep}{5pt}%
\fi
\iffdwrap\begin{wraptable}[24]{R}{0.5\textwidth}\vspace{-12pt}\else\begin{table}[ht]\fi
\centering
\fdsize
\setlength{\tabcolsep}{\fdcolsep}
\renewcommand{\arraystretch}{1.15}
\iffdwrap
\caption{\textbf{FLUX DrawBench-200.} One block per speed band. DrmSim is DreamSim.}
\else
\caption{\textbf{FLUX DrawBench-200.} One block per speed band.}
\fi
\label{tab:flux_drawbench_main}
\begin{tabular}{lc*{4}{wc{23.16pt}}}%
\toprule
  \multirow{2}{*}{Method} & \multicolumn{1}{c}{Efficiency} & \multicolumn{4}{c}{Quality} \\
  \cmidrule(lr){2-2} \cmidrule(lr){3-6}
   & \scriptsize{Speedup} & \scriptsize{PSNR$\uparrow$} & \scriptsize{SSIM$\uparrow$} & \scriptsize{LPIPS$\downarrow$} & \scriptsize{DrmSim$\downarrow$} \\
\midrule
  TeaCache & $4.02\times$ & 16.02 & 0.660 & 0.396 & 0.140 \\
  TaylorSeer & $3.70\times$ & 15.50 & 0.637 & 0.397 & 0.128 \\
  SeaCache & $3.25\times$ & 22.41 & 0.830 & 0.173 & 0.037 \\
  DPCache & $3.49\times$ & 21.27 & 0.803 & 0.182 & 0.041 \\
  \cellcolor{gray!10}\method & \cellcolor{gray!10}$3.79\times$ & \cellcolor{gray!10}\textbf{23.20} & \cellcolor{gray!10}\textbf{0.847} & \cellcolor{gray!10}\textbf{0.146} & \cellcolor{gray!10}\textbf{0.028} \\
\midrule
  TeaCache & $5.28\times$ & 15.19 & 0.619 & 0.466 & 0.193 \\
  TaylorSeer & $5.59\times$ & 11.44 & 0.448 & 0.695 & 0.385 \\
  SeaCache & $5.28\times$ & 18.81 & 0.724 & 0.326 & 0.088 \\
  DPCache & $4.74\times$ & 18.05 & 0.700 & 0.305 & 0.086 \\
  \cellcolor{gray!10}\method & \cellcolor{gray!10}$5.44\times$ & \cellcolor{gray!10}\textbf{20.05} & \cellcolor{gray!10}\textbf{0.770} & \cellcolor{gray!10}\textbf{0.251} & \cellcolor{gray!10}\textbf{0.060} \\
\midrule
  TeaCache & $5.91\times$ & 14.73 & 0.596 & 0.516 & 0.244 \\
  TaylorSeer & $6.22\times$ & 10.34 & 0.363 & 0.824 & 0.579 \\
  SeaCache & $6.66\times$ & 17.46 & 0.673 & 0.409 & 0.131 \\
  DPCache & $6.50\times$ & 14.83 & 0.537 & 0.533 & 0.206 \\
  \cellcolor{gray!10}\method & \cellcolor{gray!10}$6.91\times$ & \cellcolor{gray!10}\textbf{18.58} & \cellcolor{gray!10}\textbf{0.710} & \cellcolor{gray!10}\textbf{0.327} & \cellcolor{gray!10}\textbf{0.087} \\
\bottomrule
\end{tabular}
\iffdwrap\end{wraptable}\else\end{table}\fi

We evaluate on one image and two video backbones. FLUX.1-dev~\citep{flux2024} generates
one $1024\times1024$ image for each of the $200$ DrawBench prompts~\citep{saharia2022photorealistic}.
HunyuanVideo~\citep{kong2024hunyuanvideo} generates one $65$-frame $640\times480$ video for each
of the $946$ VBench prompts~\citep{huang2023vbench}, and Wan2.1-T2V-1.3B~\citep{wan2025wan} one
$81$-frame $832\times480$ video for each of $100$ VBench prompts, stratified by dimension
(\cref{app:vbench_subset}). Every backbone samples
$50$ Euler steps.
On FLUX we compare against the dynamic caches TeaCache~\citep{liu2025teacache} and
SeaCache~\citep{chung2026seacache}, against the recent
DPCache~\citep{cui2026dpcache}, which fixes its refresh schedule before sampling from a calibrated
cost, and against TaylorSeer~\citep{liu2025taylorseer}, which forecasts features rather than
reusing them. Under our compute budget, on video we compare against SeaCache and DPCache, which
both report gains over TeaCache and TaylorSeer~\citep{chung2026seacache,cui2026dpcache} and are
the two strongest baselines on FLUX.

On each backbone all methods share the prompts, seeds, generation settings and hardware
(\cref{app:setup_details}). Methods are compared at matched wall-clock speedup. For each method
we sweep the parameter that sets its speed, the threshold for the dynamic caches and the number of
fresh steps or the refresh interval for DPCache and TaylorSeer, and take the run nearest each
speed band (\cref{tab:baseline_knobs}). PSNR, SSIM, LPIPS~\citep{zhang2018perceptual} and
DreamSim~\citep{fu2023dreamsim} measure fidelity to the uncached sample of the same prompt and
seed, since a cache approximates the uncached sampler.
We additionally report ImageReward~\citep{xu2023imagereward} for images and VBench for video,
which score a sample on its own and separate the methods far less than the fidelity metrics do
(\cref{sec:results_video,app:unused_metrics}).
\Cref{tab:prompts_image,tab:prompts_video} give the full prompts used in every qualitative figure.

\subsection{Image generation}
\label{sec:results_image}

\Cref{tab:flux_drawbench_main} reports four fidelity metrics at three matched speed bands on
FLUX, \cref{app:main_tables} adds latency, TFLOPs and peak memory, and \cref{app:scatter}
resolves each band per prompt. \method is ahead of every baseline on all four metrics in all
three bands, and at the fastest band it leads SeaCache by $1.12$~dB PSNR and $0.082$ LPIPS.
Because each method's speed is set by its own parameter, a lead in quality could come from
running slower. None of the leads above does. \method is faster than SeaCache and DPCache in
every band, and the only two runs faster than it, TeaCache at the slowest band by $0.23\times$
and TaylorSeer at the middle band by $0.15\times$, trail it by $7.2$ and $8.6$~dB.

\Cref{fig:qual_flux_main} shows generation for three DrawBench prompts at the $\approx\!5.2\times$~band, and illustrates what the gaps of
\cref{tab:flux_drawbench_main} look like in the image. In these examples the baselines lose the
structure of objects, not only their detail. In the first row two of them break the zebra itself,
leaving it without a head (TeaCache) or with a second one (DPCache, red ellipses), and even SeaCache,
the closest to \method, loses the stripe
pattern in the enlarged region, where black bleeds into the white stripes. The canopy of the second
row grows curved struts it does not have, and the scarecrow of the third row grows wings it does
not have (red ellipses). \method keeps every object the prompt names, with its shape intact, in all three rows.
\Cref{app:qualitative_grids} adds further prompts, including a set at the
$\approx\!6.6\times$ band (\cref{fig:qual_flux_6x}) where the same failures grow, and enlarged
examples at the $\approx\!5.2\times$ band (\cref{fig:supp_zoom_flux}).

\subsection{Video generation}
\label{sec:results_video}
\newif\ifhvwrap
\hvwraptrue
\ifhvwrap
  \newcommand{\hvsize}{\small}%
  \newcommand{\hvcolsep}{2.65pt}%
\else
  \newcommand{\hvsize}{\small}%
  \newcommand{\hvcolsep}{5pt}%
\fi
\ifhvwrap\begin{wraptable}[19]{R}{0.5\textwidth}\vspace{-10pt}\else\begin{table}[ht]\fi
\centering
\hvsize
\setlength{\tabcolsep}{\hvcolsep}
\renewcommand{\arraystretch}{1.15}
\ifhvwrap
\caption{\textbf{HunyuanVideo VBench 480p.} One block per matched speed band. DrmSim is DreamSim.}
\else
\caption{\textbf{HunyuanVideo VBench 480p.} One block per matched speed band.}
\fi
\label{tab:hunyuan_vbench_main}
\begin{tabular}{lc*{4}{wc{23.16pt}}}%
\toprule
  \multirow{2}{*}{Method} & \multicolumn{1}{c}{Efficiency} & \multicolumn{4}{c}{Quality} \\
  \cmidrule(lr){2-2} \cmidrule(lr){3-6}
   & \scriptsize{Speedup} & \scriptsize{PSNR$\uparrow$} & \scriptsize{SSIM$\uparrow$} & \scriptsize{LPIPS$\downarrow$} & \scriptsize{DrmSim$\downarrow$} \\
\midrule
  SeaCache & $3.53\times$ & 23.63 & 0.778 & 0.203 & 0.102 \\
  DPCache & $3.28\times$ & 23.70 & 0.786 & 0.182 & 0.092 \\
  \cellcolor{gray!10}\method & \cellcolor{gray!10}$3.52\times$ & \cellcolor{gray!10}\textbf{26.48} & \cellcolor{gray!10}\textbf{0.843} & \cellcolor{gray!10}\textbf{0.141} & \cellcolor{gray!10}\textbf{0.067} \\
\midrule
  SeaCache & $5.33\times$ & 21.82 & 0.713 & 0.292 & 0.162 \\
  DPCache & $5.42\times$ & 20.34 & 0.682 & 0.293 & 0.162 \\
  \cellcolor{gray!10}\method & \cellcolor{gray!10}$5.39\times$ & \cellcolor{gray!10}\textbf{22.36} & \cellcolor{gray!10}\textbf{0.742} & \cellcolor{gray!10}\textbf{0.246} & \cellcolor{gray!10}\textbf{0.129} \\
\midrule
  SeaCache & $6.11\times$ & 20.38 & 0.667 & 0.349 & 0.205 \\
  DPCache & $6.07\times$ & 20.06 & 0.660 & 0.310 & 0.172 \\
  \cellcolor{gray!10}\method & \cellcolor{gray!10}$6.05\times$ & \cellcolor{gray!10}\textbf{22.27} & \cellcolor{gray!10}\textbf{0.739} & \cellcolor{gray!10}\textbf{0.267} & \cellcolor{gray!10}\textbf{0.141} \\
\bottomrule
\end{tabular}
\ifhvwrap\end{wraptable}\else\end{table}\fi

\Cref{tab:hunyuan_vbench_main} repeats the comparison on HunyuanVideo against the two video
baselines. \method leads both on all four metrics in all three bands, by $0.54$ to $2.78$~dB
PSNR over the better one. As on FLUX, the leads do not come from running slower: no baseline
is more than $0.06\times$ (about $1\%$) faster in any band. The lead holds where the budget is
tightest.
From the middle to the fastest band \method loses $0.09$~dB PSNR and $0.021$ LPIPS and
SeaCache $1.44$~dB and $0.057$ LPIPS. Without its jump, \method, like SeaCache, can gain speed only by
refreshing less often, and it then falls furthest behind as the budget tightens
(\cref{sec:ablations}).
On VBench, scored with one sample per prompt, \method matches DPCache to within $0.003$ on
the Total in every band and scores above SeaCache in all three (\cref{app:unused_metrics}).
We extend the video comparison to Wan2.1-T2V-1.3B at two points, 
leading SeaCache and DPCache on all four metrics at both (\cref{tab:wan_vbench} in \cref{app:main_tables}).

\Cref{fig:grid_hunyuan_main} shows two VBench clips at the $\approx\!5.4\times$ band.
DPCache alters what the clips contain, adding a person the prompt does not ask for in the first and
a malformed duck in mid-air in the second (red ellipses).
Both baselines also lose detail, and in the last frame of the first clip SeaCache dissolves the
dog's head into speckle while DPCache keeps only a smeared outline of it (inset).
\method keeps the content of the reference in both clips, with
the dog's head clean down to the collar, the structure of the astronaut's gloves intact (insets),
and more background detail than either baseline.
The appendix repeats this comparison on two more clips at this band
(\cref{fig:supp_grid_hunyuan_5x}) and at other speeds: HunyuanVideo at $\approx\!3.5\times$
(\cref{fig:supp_grid_hunyuan}), FLUX at $\approx\!6.6\times$ (\cref{fig:qual_flux_6x}), and
Wan2.1 at $\approx\!3.7\times$ (\cref{fig:grid_wan}), and HunyuanVideo at its native 720p
resolution at $\approx\!4.5\times$ (\cref{fig:supp_grid_hunyuan720}).

\expandafter\def\csname gh@seacache\endcsname{5.33}
\expandafter\def\csname gh@dpcache\endcsname{5.42}
\expandafter\def\csname gh@ours\endcsname{5.39}

\newif\ifhunywrap
\hunywrapfalse
\ifhunywrap
  \newcommand{\hunypanelw}{0.27}%
\else
  \newcommand{\hunypanelw}{0.15}%
\fi
\ifhunywrap\begin{wrapfigure}[24]{r}{0.6\textwidth}\vspace{-10pt}\else\begin{figure}[ht]\fi
  \centering
  \newlength{\hpanelw}\setlength{\hpanelw}{\hunypanelw\linewidth}%
  \newlength{\hframesep}\setlength{\hframesep}{4pt}%
  \newcommand{\hlabsize}{\scriptsize}%
  \newcommand{\hlabline}{0.5pt}%
  \newcommand{\hlabsep}{3pt}%
  \newcommand{\hclipsep}{4pt}%
  \newcommand{\hrowsep}{4pt}%
  \renewcommand{\filmstriph}{4.5}%
  \renewcommand{\filmholew}{3.2pt}%
  \renewcommand{\filmholeh}{2.2pt}%
  \renewcommand{\filmnholes}{8}%
  \renewcommand{\filmstripcol}{black!85}%
  \renewcommand{\zoominsetfrac}{0.42}%
  \setlength{\zoompop}{2.8pt}%
  \setlength{\zoompopx}{\zoompop}\setlength{\zoompopy}{\zoompop}%
  \newlength{\htripw}\setlength{\htripw}{3\hpanelw}\addtolength{\htripw}{2\hframesep}%
  \setlength{\tabcolsep}{0pt}%
  \renewcommand{\arraystretch}{0}%
  \def\hroiA{vbad077}%
  \newcommand{\hroi}[1]{%
    \def\hroiid{#1}%
    \ifx\hroiid\hroiA %
      \def\ztf{2}\def\ztx{0.30}\def\zty{0.45}\def\ztw{0.18}\def\zth{0.24}\def\ztc{south east}%
    \else %
      \def\ztf{0}\def\ztx{0.445}\def\zty{0.55}\def\ztw{0.1297}\def\zth{0.1729}\def\ztc{north east}%
    \fi}%
  \newcommand{\hroiarm}[5]{\expandafter\def\csname hroiarm@#1@#2\endcsname{\def\ztx{#3}\def\zty{#4}\def\ztc{#5}}}%
  \hroiarm{vbad077}{dpcache}{0.51}{0.50}{north west}%
  \hroiarm{vbad687}{seacache}{0.425}{0.57}{north east}%
  \hroiarm{vbad687}{ours}{0.425}{0.57}{north east}%
  \newcommand{\hmark}[7]{\expandafter\def\csname hmark@#1@#2@#3\endcsname{\defectellipse{#4}{#5}{#6}{#7}}}%
  \hmark{vbad077}{dpcache}{1}{0.715}{0.62}{0.09}{0.36}%
  \hmark{vbad687}{dpcache}{1}{0.72}{0.60}{0.19}{0.08}%
  \newcommand{\hcell}[3]{\hroi{#1}\ifcsname hroiarm@#1@#2\endcsname\csname hroiarm@#1@#2\endcsname\fi%
    \ifcsname hmark@#1@#2@#3\endcsname\expandafter\let\expandafter\zoommarks\csname hmark@#1@#2@#3\endcsname\else\def\zoommarks{}\fi
    \ifnum#3=\ztf
      \zoomtile{\hpanelw}{figures/grid_hunyuan/#1/#2_f#3.jpg}{\ztx}{\zty}{\ztw}{\zth}{\ztc}%
    \else\marktile{\hpanelw}{figures/grid_hunyuan/#1/#2_f#3.jpg}\fi}%
  \newcommand{\htrip}[2]{\hcell{#1}{#2}{0}\hspace{\hframesep}%
    \hcell{#1}{#2}{1}\hspace{\hframesep}\hcell{#1}{#2}{2}}%
  \newlength{\hrowh}%
  \settoheight{\hrowh}{\includegraphics[width=\hpanelw]%
    {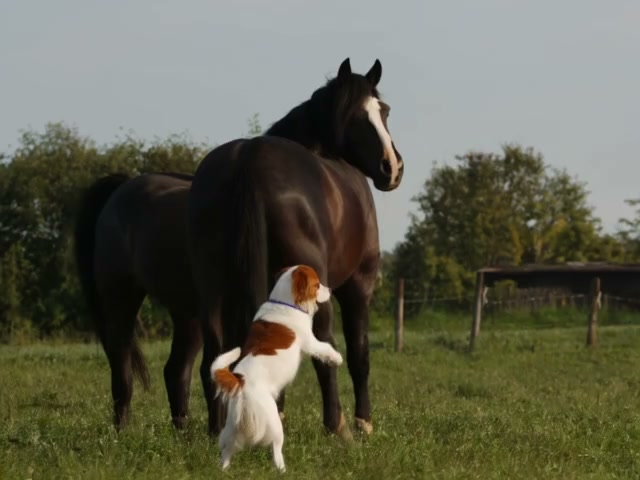}}%
  \newcommand{\hrot}[1]{\raisebox{\dimexpr0.5\hrowh-0.5\height\relax}%
    {\rotatebox{90}{{\hlabsize #1}}}}%
  \newcommand{\hlab}[2]{\hrot{\shortstack[c]{#1\\[\hlabline](#2)}}}%
  \newcommand{\hlabref}[1]{\hrot{\shortstack[c]{#1\\[\hlabline]\strut}}}%
  \newcommand{\hpromptfont}{\footnotesize}%
  \newcommand{\hprompt}[1]{\parbox{\htripw}{\centering\hpromptfont\gridprompt{#1}}}%
  \newcommand{\hfilmbar}{ & \filmsprocket{\htripw} & \filmsprocket{\htripw}}%
  \ifdefined\tikzexternaldisable\tikzexternaldisable\fi
  \begin{tabular}{@{}c@{\hspace{\hlabsep}}c@{\hspace{\hclipsep}}c@{}}
    & \hprompt{a dog and a horse} & \hprompt{An astronaut feeding ducks\dots} \\[\hrowsep]
    \hfilmbar \\[\hrowsep]
    \hlabref{Reference} & \htrip{vbad077}{reference} & \htrip{vbad687}{reference} \\[\hrowsep]
    \hlab{SeaCache}{$\csname gh@seacache\endcsname\times$} & \htrip{vbad077}{seacache} & \htrip{vbad687}{seacache} \\[\hrowsep]
    \hlab{DPCache}{$\csname gh@dpcache\endcsname\times$} & \htrip{vbad077}{dpcache} & \htrip{vbad687}{dpcache} \\[\hrowsep]
    \hlab{\method}{$\csname gh@ours\endcsname\times$} & \htrip{vbad077}{ours} & \htrip{vbad687}{ours} \\[\hrowsep]
    \hfilmbar \\
  \end{tabular}
  \ifdefined\tikzexternalenable\tikzexternalenable\fi
  \caption{\textbf{HunyuanVideo on VBench at $\approx\!5.4\times$ speedup.} Each clip shows its
  first, middle and last frame, and the speedup beside each method is its band mean. Insets
  enlarge one region, and red ellipses mark a person the prompt does not ask for and a malformed
  duck in mid-air.}
  \label{fig:grid_hunyuan_main}
\ifhunywrap\end{wrapfigure}\else\end{figure}\fi

\subsection{Ablations}
\label{sec:ablations}

We ablate on FLUX with $50$ prompts from DPG-Bench~\citep{hu2024ella}, kept separate from the
DrawBench-200 test prompts. Each variant disables one term of the sampler and keeps the rest of
\method. Removing the jump sets $J{=}1$ in \cref{eq:jump}, so every cached node takes the
scheduled step and the sampler stays on the grid. Removing headroom scaling replaces $H_i$ by $1$
in the same equation, so a cached node jumps the full $J\Delta_i$ whenever it jumps rather
than in proportion to the reuse budget left.
Removing tail re-spacing holds $m_i$ fixed in \cref{eq:tail_regrid}, so the rebuilt tail
retires one node per jump however far the jump lands, and the extra distance saves no
nodes. Removing residual motion sets $\beta{=}0$ in
\cref{eq:residual_motion}, so a cached node holds $r_a$ as in \cref{eq:zoh}.

\expandafter\def\csname qp@cap8\endcsname{2.92}
\expandafter\def\csname qp@cap1\endcsname{3.06}
\expandafter\def\csname qp@cap8fixed\endcsname{3.09}
\expandafter\def\csname qp@bincap8\endcsname{2.93}

\newif\ifqawrap
\qawraptrue
\ifqawrap
  \newcommand{\qpanelw}{0.19}%
  \newcommand{\qlabsize}{\fontsize{6}{7}\selectfont}%
\else
  \newcommand{\qpanelw}{0.192}%
  \newcommand{\qlabsize}{\fontsize{7}{8}\selectfont}%
\fi
\ifqawrap\begin{wrapfigure}[24]{R}{0.6\textwidth}\vspace{-10pt}\else\begin{figure}[ht]\fi
  \centering
  \newcommand{\qcolsep}{1pt}%
  \newcommand{\qrowsep}{1pt}%
  \setlength{\tabcolsep}{\qcolsep}%
  \renewcommand{\arraystretch}{0}%
  \newcommand{\qimg}[1]{\includegraphics[width=\qpanelw\linewidth]{figures/qual_ablation_pianist/#1}}%
  \newcommand{\qlab}[2]{\vtop{%
    \hbox to \qpanelw\linewidth{\hss\qlabsize\strut #1\hss}%
    \hbox to \qpanelw\linewidth{\hss\qlabsize\color{black!55}\ifx&#2&1.00\else\csname qp@#2\endcsname\fi$\times$\hss}}}%
  \newcommand{\qheadsep}{2pt}%
  \ifdefined\tikzexternaldisable\tikzexternaldisable\fi
  \begin{tabular}{@{}ccccc@{}}
    \qlab{Reference}{} & \qlab{w/o jumps}{cap1} & \qlab{w/o headroom}{bincap8} & \qlab{w/o re-spacing}{cap8fixed} & \qlab{\method}{cap8} \\[\qheadsep]
    \qimg{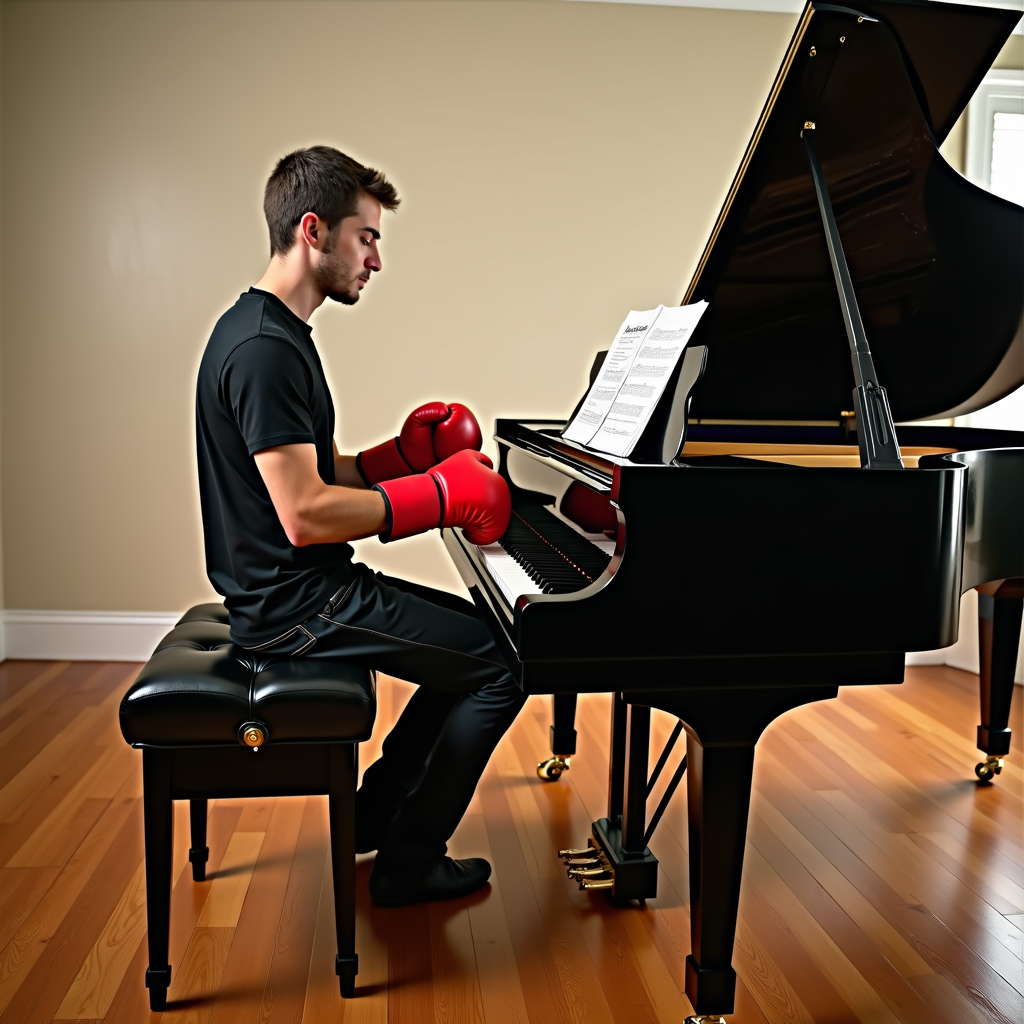} & \qimg{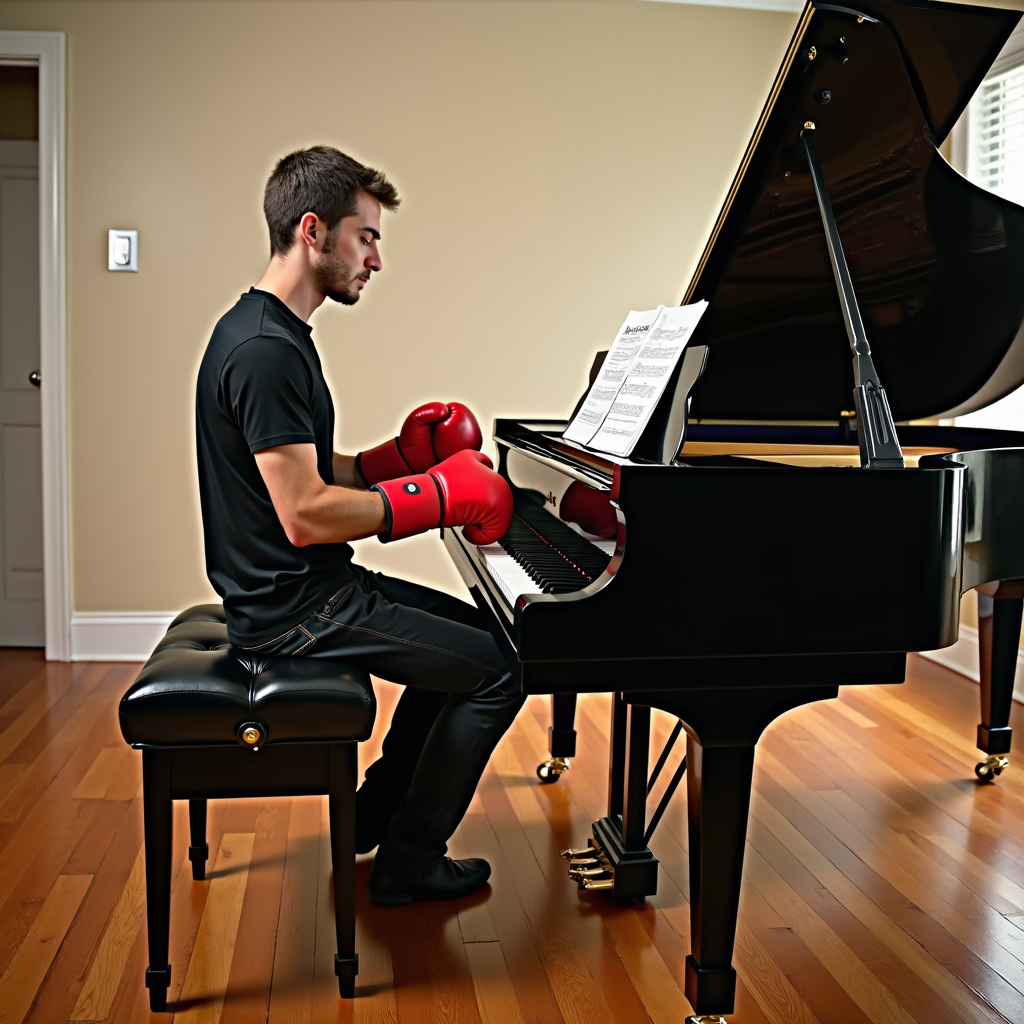} & \qimg{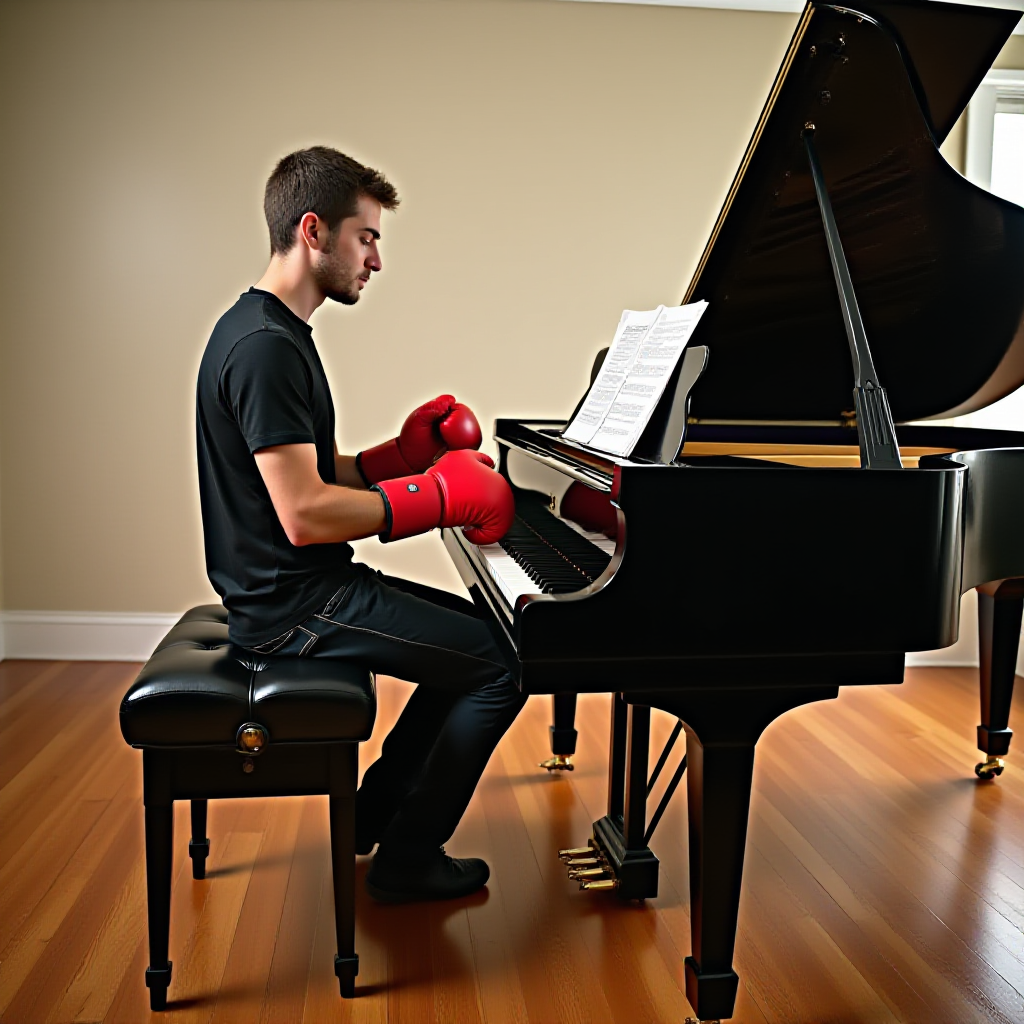} & \qimg{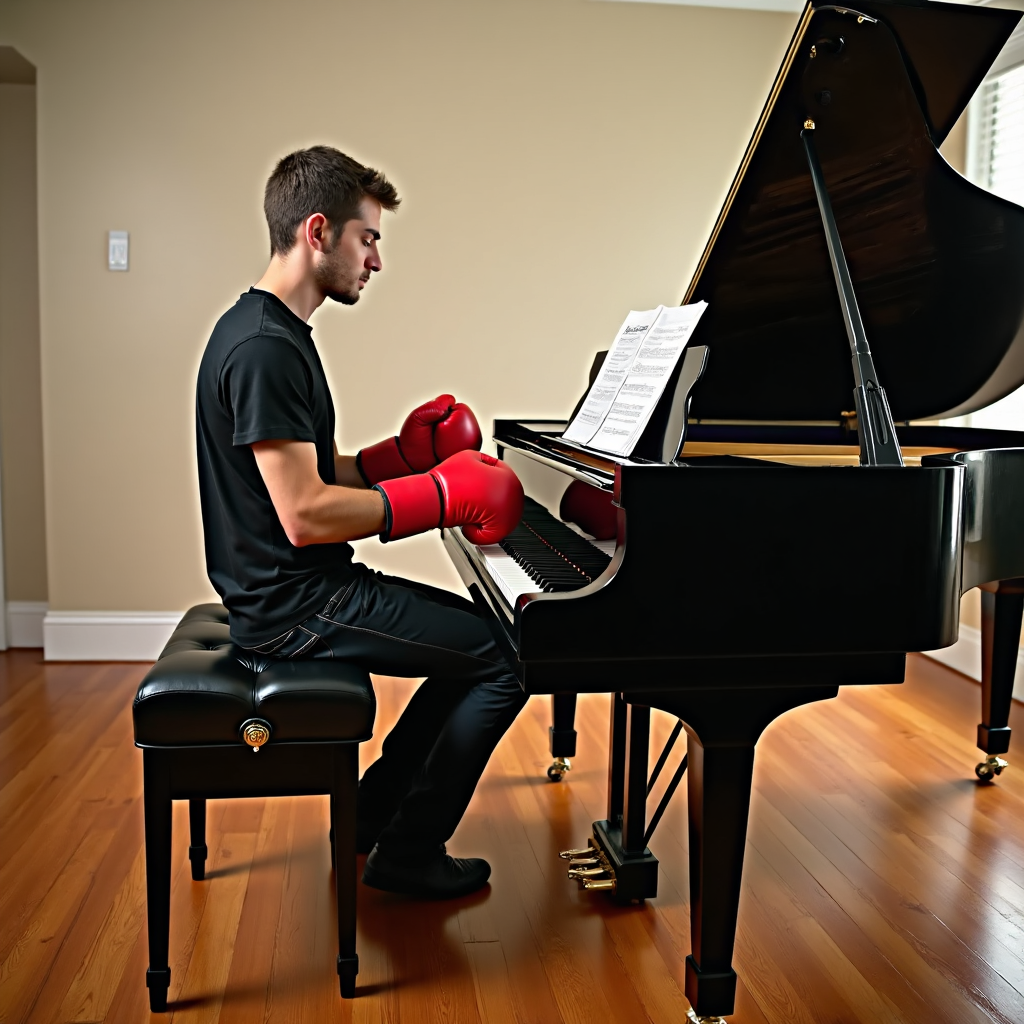} & \qimg{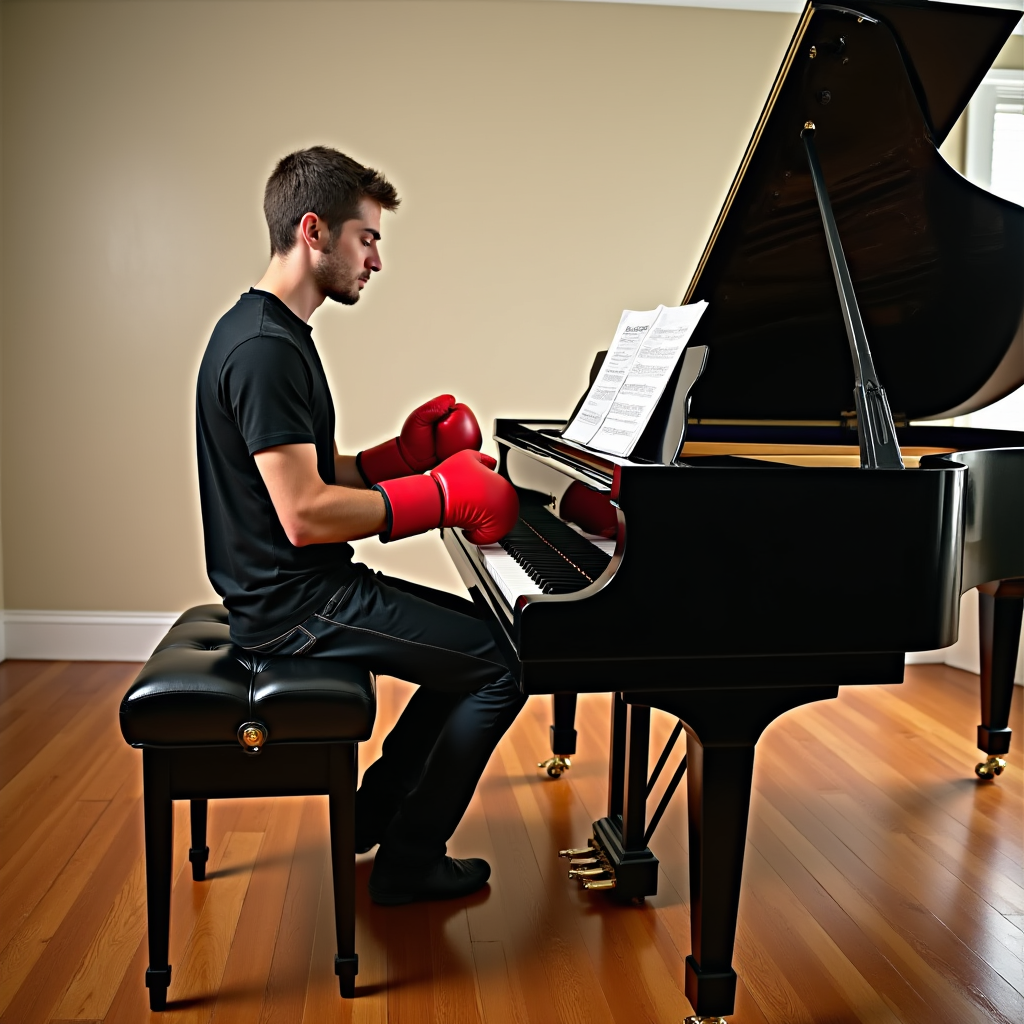} \\[\qrowsep]
    \qimg{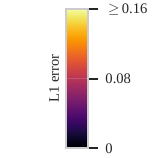} & \qimg{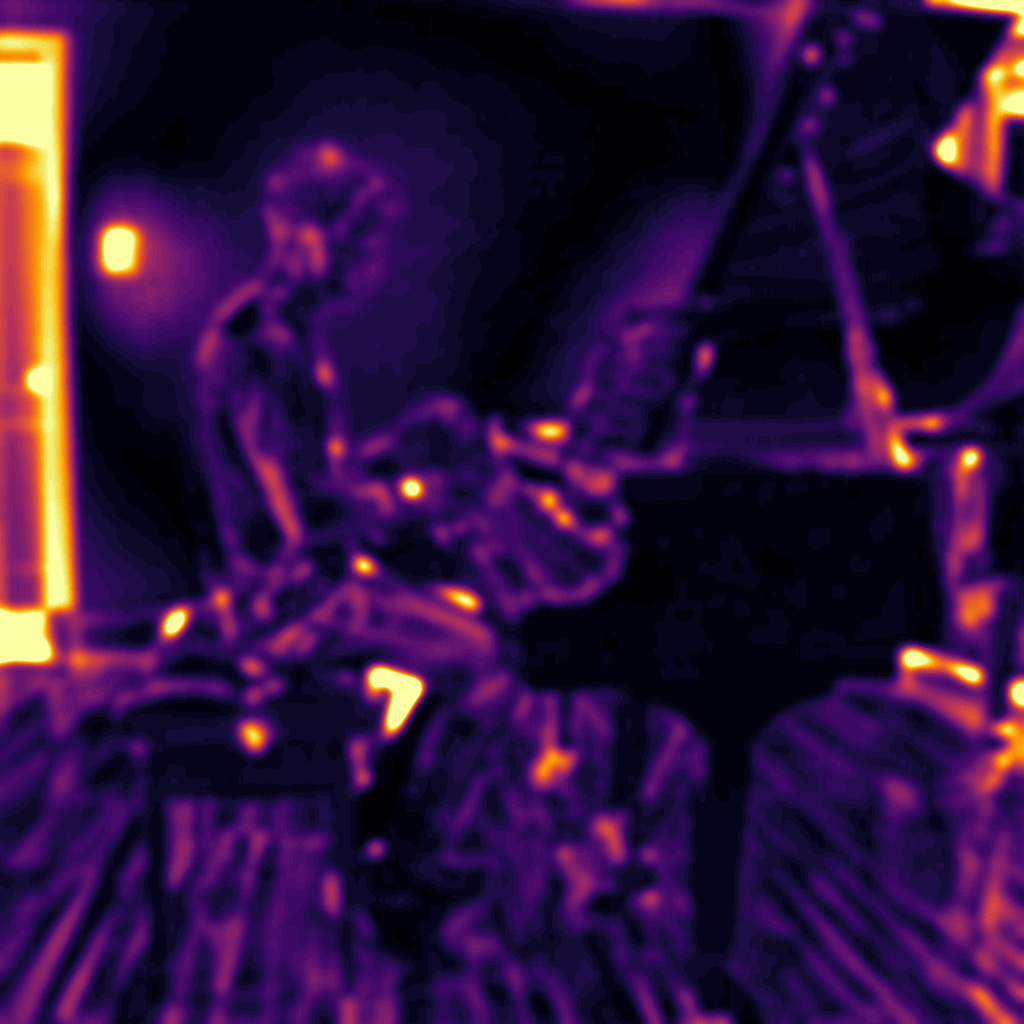} & \qimg{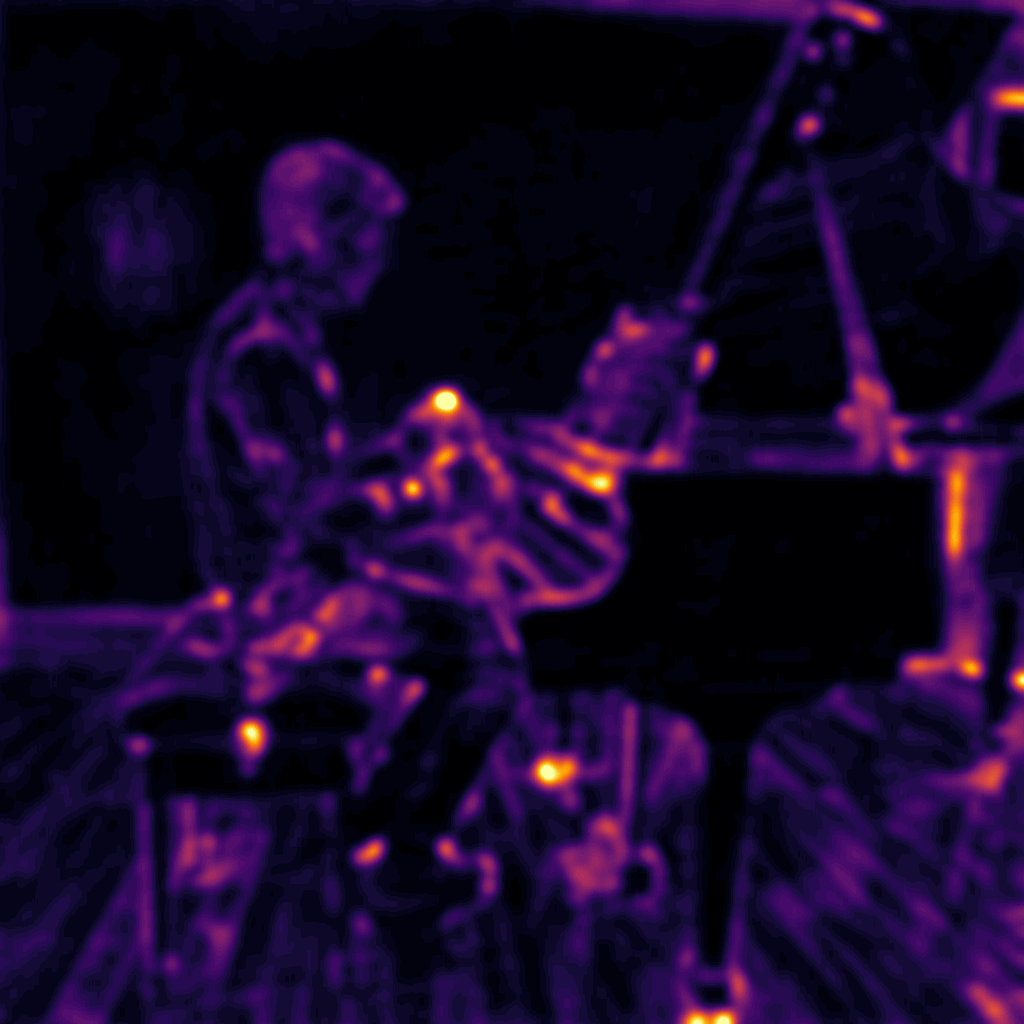} & \qimg{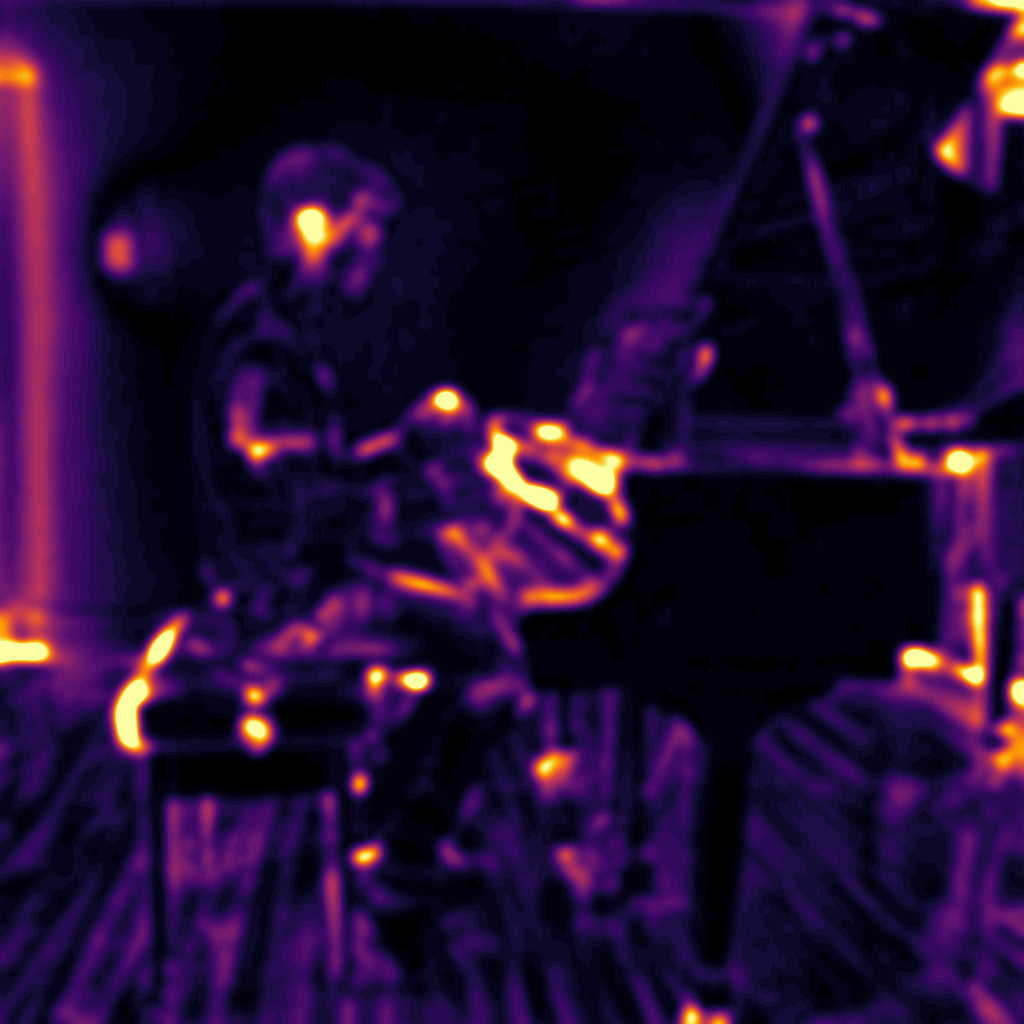} & \qimg{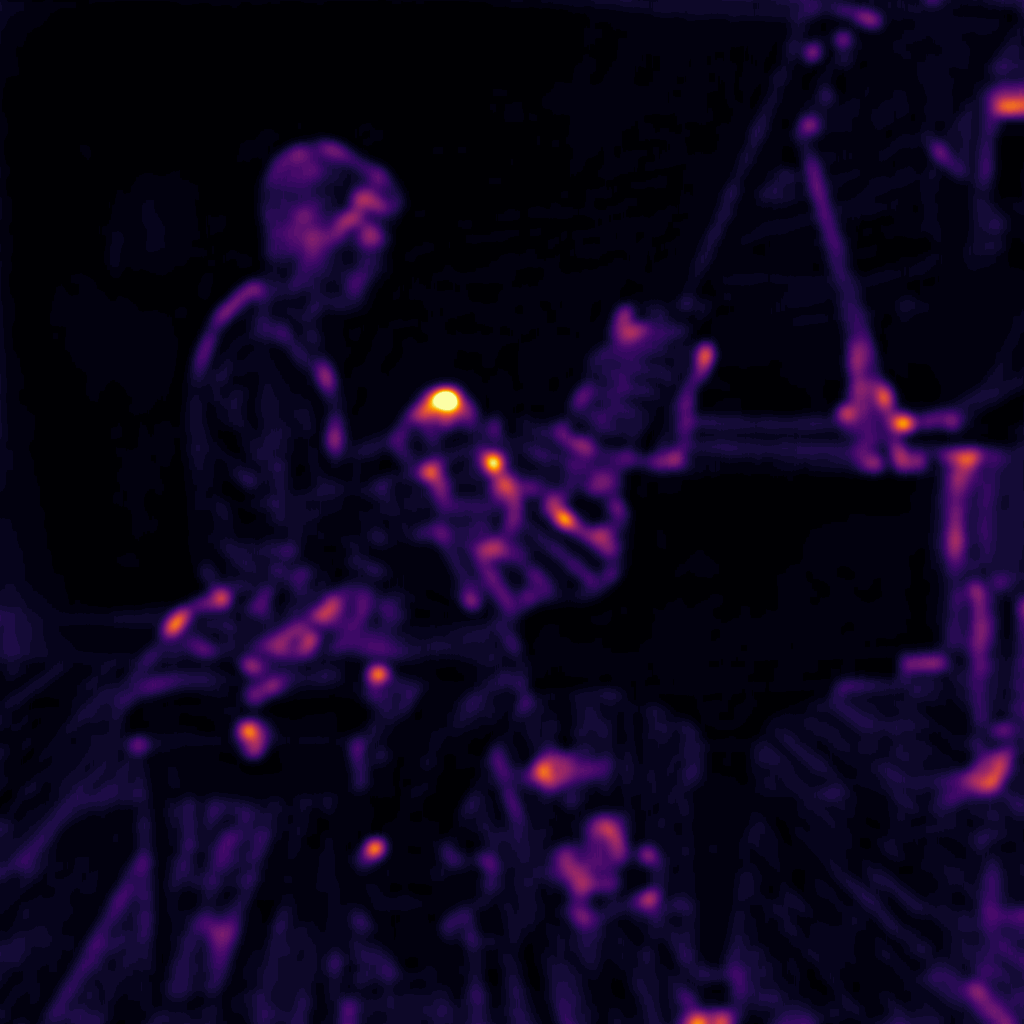} \\
  \end{tabular}
  \ifdefined\tikzexternalenable\tikzexternalenable\fi

  \vspace{4pt}
  {\centering
\scriptsize
\setlength{\tabcolsep}{1.5pt}
\renewcommand{\arraystretch}{1.15}
\begin{tabular}{lcccccc}
\toprule
  \multirow{2}{*}{} & \multicolumn{3}{c}{$3\times$} & \multicolumn{3}{c}{$7\times$} \\
  \cmidrule(lr){2-4} \cmidrule(lr){5-7}
   & LPIPS$\downarrow$ & DrmSim$\downarrow$ & PSNR$\uparrow$ & LPIPS$\downarrow$ & DrmSim$\downarrow$ & PSNR$\uparrow$ \\
\midrule
  w/o jumps ($J{=}1$) & 0.103 & 0.018 & 24.30 & 0.548 & 0.199 & 13.67 \\
  w/o headroom ($H_i{=}1$) & 0.122 & 0.016 & 25.70 & 0.330 & 0.082 & \textbf{18.17} \\
  w/o re-spacing ($m_i$ fixed) & 0.109 & 0.018 & 24.64 & 0.371 & 0.101 & 16.64 \\
  \cellcolor{gray!10}\method & \cellcolor{gray!10}\textbf{0.093} & \cellcolor{gray!10}\textbf{0.013} & \cellcolor{gray!10}\textbf{26.50} & \cellcolor{gray!10}\textbf{0.329} & \cellcolor{gray!10}\textbf{0.078} & \cellcolor{gray!10}17.85 \\
\bottomrule
\end{tabular}\par}

  \ifqawrap
  \caption{\textbf{Component leave-one-out (FLUX, $50$ DPG-Bench prompts).} Top: one prompt at
  $\approx\!3\times$ with its per-pixel $L_1$ error to the uncached image. Bottom: means at matched
  wall-clock speedup. Every removal trails \method at $3\times$, and removing the jump costs the
  most at $7\times$. \Cref{tab:ablation_loo} adds $5\times$ and residual motion.}
  \else
\caption{\textbf{Component ablations (FLUX, DPG-50).} Top: an example at $\approx\!3\times$
with its per-pixel error to the uncached image. Bottom: means over $50$ prompts at matched
speedup.}
  \fi
  \label{fig:qual_ablation_pianist}
\ifqawrap\end{wrapfigure}\else\end{figure}\fi

At $3\times$ every removal trails \method on all three metrics. At higher speeds the removals fall
into two groups. Removing the jump or the tail re-spacing trails on every metric at every band.
Without the jump the sampler stays on the scheduled grid, so it gains speed only by refreshing
less often, and at $7\times$ it trails \method by $4.18$~dB PSNR and $0.219$ LPIPS. Holding $m_i$
fixed costs less, $1.21$~dB, because the jump still lands where the headroom allows, but the tail
keeps its node count, so the distance the jump covers buys no speed. Removing headroom scaling or residual
motion behaves differently. 
\method stays ahead of both on LPIPS and DreamSim at every band, while both pass it on PSNR at
high speed (\cref{fig:qual_ablation_pianist,tab:ablation_loo}).
PSNR averages pixel error, so it rewards an output that blurs fine detail over one that keeps the detail but places it slightly
off, which the perceptual metrics prefer. \Cref{app:metric_reading} shows such a case.

The scale $\beta$ in \cref{eq:residual_motion} sets how far a cached node follows the
linear extrapolation. The extrapolation grows with distance from the last fresh evaluation,
through $\lambda_i$, so at high speed it reaches beyond the two evaluations it was
fitted to and becomes less reliable.
Damping it to $\beta{=}0.5$ loses to
$\beta{=}1$ at $3\times$ but beats it by $1.65$~dB PSNR and $0.080$ LPIPS at $7\times$
(\cref{tab:ablation_design}). We therefore use $\beta{=}1$ below $4\times$ and $\beta{=}0.5$ above it,
the same rule on all three backbones.
Raising the cap instead
gains less: at $7\times$ even $J{=}20$ gains $1.12$~dB and $0.050$ LPIPS,
and it loses fine texture that $\beta{=}0.5$ keeps (\cref{app:ablation_extra}).
Removing residual motion scores a higher PSNR than
$\beta{=}0.5$ at $7\times$ but a worse LPIPS (\cref{tab:ablation_loo}).
Comparing the images ourselves at that band, blind to the variant, we prefer $\beta{=}0.5$ to
$J{=}20$ and to removing residual motion (\cref{app:cap_dial}).

\section{Conclusion}
\label{sec:conclusion}
We presented \method, a training-free method that turns hard caching into
soft caching. A dynamic cache already measures how stale its cached
computation has become, but spends the measurement on a binary choice between
refresh and reuse. \method reads the headroom below the threshold as how far
the cache can still be trusted and lets it set the step size: cached steps
jump past scheduled nodes while the cache is reliable and shrink back to the
scheduled step as a refresh nears, the tail is re-spaced so that every node
passed is dropped from the trajectory, and the cached residual is extrapolated to each
off-grid landing at no extra network cost.
On FLUX, HunyuanVideo and Wan2.1, at wall-clock speedups from
$3.5\times$ to $6.9\times$, \method stays closest to the uncached sample on all
four fidelity metrics, with no per-model calibration.
Adaptivity in caching, then, does not stop at deciding when to refresh.

\ifarxiv
\section*{Acknowledgments}
This project was partially supported by ISF grants 1574/21 and 2132/23.

\else
\section*{AI usage disclosure}

We used generative AI tools to refine hypotheses, to give feedback on
experimental design, to interpret results, and to implement methods. The
implementation was developed with AI assistance throughout, covering the
method code, the integration of prior caching methods, the evaluation and
timing harness, and the analysis and plotting scripts, including code that
produced numbers reported in this paper. We did not use generative AI tools
to generate synthetic data, and the remaining tasks that require disclosure,
namely developing theoretical models, formulating or proving mathematical
claims, translation, dataset cleaning, and qualitative or thematic analysis,
are not applicable to this work. We additionally used generative AI tools to
draft and edit parts of the paper, to identify related work, to suggest a
structure for the paper, and to produce scripts for figures and tables.

All experiments were run by the authors on the authors' own hardware, and
every number reported in this paper comes from those runs.
An author reviewed every AI-assisted change to the manuscript
before it entered the source, all citations were checked against the papers
they cite, and bibliography entries were added only by an author. We take
responsibility for the final content of this work, including text, claims and
artifacts produced with the aid of generative AI.

\section*{Reproducibility Statement}

\method is training-free and runs on public checkpoints of FLUX, HunyuanVideo and Wan2.1.
The full sampler is given in \cref{alg:method}.
Hyperparameters, prompt sets, baseline configurations, hardware and the timing protocol
are listed in \cref{app:setup_details}.
Every design choice and ablation was made on DPG-50,
and DrawBench-200 was held out as the test set.
Code and evaluation scripts will be released upon acceptance.

\fi

\bibliography{main}
\bibliographystyle{iclr2027_conference}

\clearpage
\appendix
\section*{Appendices}
\setcounter{table}{0}\setcounter{figure}{0}\setcounter{algorithm}{0}
\counterwithin{table}{subsection}\counterwithin{figure}{subsection}\counterwithin{algorithm}{subsection}
\renewcommand{\theHtable}{A.\thesubsection.\arabic{table}}\renewcommand{\theHfigure}{A.\thesubsection.\arabic{figure}}\renewcommand{\theHalgorithm}{A.\thesubsection.\arabic{algorithm}}
In the following sections we provide additional setup and implementation details
(\Cref{app:grp_setup}), extended quantitative results (\Cref{app:grp_results}),
additional ablations (\Cref{app:grp_ablations}), and further qualitative results
and discussion (\Cref{app:grp_qual}).
\ifarxiv\else
\textbf{We encourage the reader to open \texttt{index.html}}, located at the top level of the supplementary ZIP archive,
which presents side-by-side image and video comparisons of \method and the baselines at matched speeds.
\fi

\section{Experimental setup and implementation}
\label{app:grp_setup}
\subsection{Setup details}
\label{app:setup_details}

\Cref{alg:method} is the full \method sampler, including its initialization.
The first node is always fresh, since no anchor exists before it,
and until the first refresh a cached node holds the single anchor rather than extrapolating.

\begin{algorithm}[H]
\caption{\method sampling}
\label{alg:method}
\label{app:algorithm}
\setlength{\lineskiplimit}{2pt}\setlength{\lineskip}{3pt}
\begin{algorithmic}[1]
\Require noise $x_0$, model $v_\theta = \mathcal{O}(\mathcal{B}(\mathcal{E}(\cdot), e),\ e)$ as in \cref{eq:factorization}, grid $0 = t_0 < \cdots < t_N = 1$, threshold $\tau$, cap $J$, scale $\beta$
\State $i \gets 0$
\While{$t_i < 1$}
    \State $s_i \gets \mathcal{E}(x_i)$
    \IfThen{$i > 0$}{$A_i \gets A_{i-1} + \rho_i$}
    \If{$i = 0$ \textbf{or} $A_i \geq \tau$} \Comment{fresh: evaluate the stack}
        \IfThen{$i > 0$}{$(r_{a-1}, t_{a-1}) \gets (r_a, t_a)$}
        \State $(r_a, t_a) \gets (\mathcal{B}(s_i, e_i) - s_i,\ t_i)$, \quad $A_i \gets 0$
        \State $\widehat r_i \gets r_a$, \quad $\bar\Delta_i \gets \Delta_i$
    \Else \Comment{cached: extrapolate the residual, stretch the step}
        \If{$r_{a-1}$ exists}
            \State $\widehat r_i \gets r_a + \beta\,\lambda_i\,(r_a - r_{a-1})$ \Comment{\cref{eq:residual_motion}}
        \Else
            \State $\widehat r_i \gets r_a$ \Comment{one anchor so far, \cref{eq:zoh}}
        \EndIf
        \State $H_i \gets 1 - A_i/\tau$, \quad $\bar\Delta_i \gets \min\{1 - t_i,\ \Delta_i + (J-1)H_i\Delta_i\}$ \Comment{\cref{eq:jump}}
        \IfThen{opening stretch \textbf{or} large probe change}{$\bar\Delta_i \gets \Delta_i$} \Comment{no jump, \cref{app:gate}}
    \EndIf
    \State $x_{i+1} \gets x_i + \bar\Delta_i\,\mathcal{O}(s_i + \widehat r_i,\ e_i)$, \quad $t_{i+1} \gets t_i + \bar\Delta_i$
    \IfThen{$\bar\Delta_i > \Delta_i$ \textbf{and} $t_{i+1} < 1$}{re-space the nodes still ahead by \cref{eq:tail_regrid}}
    \State $i \gets i + 1$
\EndWhile
\State \Return $x_i$
\end{algorithmic}
\end{algorithm}

FLUX uses all $200$ prompts of DrawBench~\citep{saharia2022photorealistic}.
HunyuanVideo uses all $946$ prompts of VBench~\citep{huang2023vbench} at every budget, and
Wan2.1 uses a subset of $100$ VBench prompts (\cref{app:vbench_subset}). The ablations of
\cref{sec:ablations} use $50$ prompts from DPG-Bench~\citep{hu2024ella}, disjoint from DrawBench
and subsampled deterministically from the released list.
The $25$ prompts of \cref{fig:schedule_heatmap} are drawn from these $200$ DrawBench prompts.

Every backbone samples $50$ steps with a first-order Euler step in bf16, one sample per prompt
from seed $0$ on a CPU generator, so all methods in a comparison start from the same initial
noise. FLUX loads \texttt{black-forest-labs/FLUX.1-dev} with guidance $3.5$, the checkpoint's
resolution-dependent shift, $s=e^{1.15}$ at $1024\times1024$, and $512$ T5 tokens.
HunyuanVideo loads \texttt{hunyuanvideo-community/HunyuanVideo} with guidance $6$, shift $s=7$
and $256$ Llama tokens. Wan2.1 loads \texttt{Wan-AI/Wan2.1-T2V-1.3B-Diffusers} with
classifier-free guidance $6$, shift $s=3$ and $226$ UMT5 tokens. FLUX and
HunyuanVideo use no negative prompt, since neither runs a classifier-free branch, and Wan2.1
runs classifier-free guidance with an empty negative prompt. On Wan2.1 the Euler step replaces the UniPC solver the checkpoint ships
with. FLUX generates $1024\times1024$ images,
HunyuanVideo $640\times480$ videos of $65$ frames and Wan2.1 $832\times480$ videos of $81$
frames.
Videos are saved as MP4 at $8$~fps on HunyuanVideo and $16$~fps on Wan2.1.
The 720p HunyuanVideo comparison of \cref{fig:supp_grid_hunyuan720} generates
$1280\times720$ videos of $129$ frames.
FLUX and HunyuanVideo run on one H100 NVL and Wan2.1 on one RTX A5000, so Wan2.1
latencies are not comparable with the other two backbones.

Every \method result uses the cap $J=8$, first-order residual motion
(\cref{eq:residual_motion}), the tail re-spacing of \cref{eq:tail_regrid} and the
two jump guards of \cref{app:gate}, with $\tau$ as the only parameter that sets
the speed.
The scale $\beta$ is $1$ below $4\times$
and $0.5$ above it (\cref{sec:ablations}).

All methods, \method and every baseline, are run by us on the hardware and settings above,
and every number in the paper comes from these runs.
\Cref{tab:baseline_knobs} gives, per method, the parameter that sets its speed, the values
swept on each backbone, and in bold the values that appear in \cref{app:main_tables}.
DPCache~\citep{cui2026dpcache} releases code for FLUX and for Wan2.1-I2V-14B, so we adapt its
official implementation to HunyuanVideo and Wan2.1-T2V-1.3B and recalibrate its cost tensor on
$10$ VBench prompts disjoint from the evaluation prompts. On FLUX its released cost tensor was
calibrated with fp16 Taylor states, so we recalibrate it at bf16 with its own protocol on $10$
of the $200$ DrawBench prompts. There, DPCache at $k=5$ is excluded
as degenerate ($8.64\times$, $8.46$~dB PSNR), and TeaCache and TaylorSeer are not run on video,
where each additional method costs a full VBench sweep per backbone and speed.
Every method is scored against the uncached sample of the same prompt and seed. \method,
SeaCache and TeaCache run in our sampler and share one reference on every backbone. DPCache
and TaylorSeer run their own drivers: DPCache's reference is byte-identical to ours, and
TaylorSeer's differs from it only by driver-level numerics.

\begin{table}[ht]
\centering
\small
\renewcommand{\arraystretch}{1.1}
\caption{\textbf{Swept speed parameter per method.} Bold values appear in \cref{app:main_tables}.}
\label{tab:baseline_knobs}
\begin{tabular}{lll}
\toprule
Method & Parameter & Swept values \\
\midrule
\multicolumn{3}{l}{\emph{FLUX}} \\
\method, $\beta{=}1$   & $\tau$ & $0.3, \mathbf{0.5}, 0.8, 1.1$ \\
\method, $\beta{=}0.5$ & $\tau$ & $\mathbf{0.8}, \mathbf{1.1}$ \\
SeaCache   & $\tau$ & $0.3, \mathbf{0.5}, 0.8, \mathbf{1.1}, \mathbf{1.5}, 2$ \\
TeaCache   & rel.\ L1 threshold & $0.5, \mathbf{0.8}, \mathbf{1.2}, \mathbf{1.6}$ \\
TaylorSeer & interval $N$ & $\mathbf{6}, 7, 10, \mathbf{16}, \mathbf{25}$ \\
DPCache    & fresh steps $k$ & $5, \mathbf{6}, 7, \mathbf{9}, \mathbf{13}$ \\
\midrule
\multicolumn{3}{l}{\emph{HunyuanVideo}} \\
\method, $\beta{=}1$   & $\tau$ & $\mathbf{0.5}, 0.6, 0.75, 0.9, 1.1, 1.3$ \\
\method, $\beta{=}0.5$ & $\tau$ & $\mathbf{0.9}, \mathbf{1.1}$ \\
SeaCache & $\tau$ & $0.5, \mathbf{0.6}, 0.75, 0.9, \mathbf{1.1}, \mathbf{1.3}, 1.6$ \\
DPCache  & fresh steps $k$ & $7, \mathbf{8}, \mathbf{9}, 11, 12, 13, \mathbf{15}$ \\
\midrule
\multicolumn{3}{l}{\emph{Wan2.1}} \\
\method, $\beta{=}1$   & $\tau$ & $0.32, \mathbf{0.38}, 0.45, 0.6, 0.73, 0.79, 1.07$ \\
\method, $\beta{=}0.5$ & $\tau$ & $0.32, 0.38, 0.6, \mathbf{0.73}$ \\
SeaCache & $\tau$ & $0.32, \mathbf{0.45}, 0.6, 0.73, \mathbf{0.79}, 1.07$ \\
DPCache  & fresh steps $k$ & $\mathbf{8}, 9, \mathbf{12}$ \\
\bottomrule
\end{tabular}
\end{table}

Settings that are not swept keep each method's released values.
TaylorSeer uses Taylor order $1$ and computes its first $3$ steps fresh.
TeaCache uses the rescaling polynomial of the released FLUX script and computes the first and
last steps fresh.
SeaCache uses its released filter, with exponent $2$ on FLUX and $3$ on the two video backbones, and the probe of \method uses the same.
DPCache uses Taylor order $2$.

No main-table result is interpolated. The runs of every method are grouped into speed bands,
and each method contributes to a band the run whose achieved speedup lies nearest it. Speedups
within a band therefore differ slightly, and each cell reports its own.
Speedup is the median uncached latency over prompts divided by the method's median
latency. The ablations match budgets by interpolation instead (\cref{app:ablation_extra}).

The ablations use the $50$ DPG-Bench prompts above. Every ablation is scored on a
seven-threshold grid, $\tau \in \{0.15, 0.3, 0.5, 0.8, 1.1, 1.5, 2.2\}$, run once per variant.
For each variant we interpolate its wall-clock speedup on the $50$ prompts, computed as for the
main results, along $\tau$ to find the threshold at which it runs at $3\times$, $5\times$ or
$7\times$. Each prompt's metrics are interpolated to that threshold and paired with \method
at its own matched threshold. A cell that would need extrapolation is dropped
rather than clamped. All variants run on an H100 80GB. The example images of
\cref{fig:qual_ablation_pianist} are rendered on an RTX 5090.

TFLOPs are counted with PyTorch's FLOP counter and include the VAE decode.
Peak memory is allocated, not reserved, GPU memory, except on Wan2.1, where it is
reserved memory.
DreamSim~\citep{fu2023dreamsim} uses version
$0.2.1$ with its default ensemble of DINO, CLIP and OpenCLIP ViT-B/16 features.
All fidelity metrics compare the decoded output with the uncached sample in RGB, and videos
are read back from their saved MP4 files.
PSNR and SSIM are computed on 8-bit values scaled to $[0,1]$ at full resolution, SSIM with the
default $7\times7$ uniform window of scikit-image on each color channel.
LPIPS~\citep{zhang2018perceptual} uses the AlexNet backbone, and DreamSim resizes its input to
$224\times224$. For video, every metric is computed per frame against the same frame of the
uncached clip, averaged over all frames ($65$ on HunyuanVideo, $81$ on Wan2.1), and then
averaged over prompts.
Samples are generated one at a time (batch size $1$). \method, SeaCache and TeaCache run in our
sampler, which times the denoising loop between two CUDA synchronizations, once per sample.
Every speedup divides the latency of an uncached run by the method's latency, both
measured the same way on the same GPU.
Attention runs through PyTorch SDPA and no model is compiled.

\subsection{VBench prompt subset}
\label{app:vbench_subset}

Wan2.1 (\cref{tab:wan_vbench}) is evaluated on $100$ of the $946$ VBench
prompts~\citep{huang2023vbench}, since running every method at every budget on the full set
was beyond our compute budget. The first $100$ prompts of the manifest are
not a usable subset: the manifest is grouped by dimension and opens with $48$ static-scene
prompts (``In a still frame\ldots''), against a $7.8\%$ static share in the full set, and
reuse across timesteps is nearly free on a static scene, so such a subset would flatter
every cache. We instead take the first $100$ prompts of a fixed low-discrepancy reordering
of the manifest, which spreads the draw evenly over the set with no seed to choose. The
subset has $8$ static prompts, and \cref{tab:vbench_dim_coverage} shows every VBench
dimension's share within two percentage points of its share in the full set.

\begin{table}[ht]
\centering
\small
\setlength{\tabcolsep}{5pt}
\caption{\textbf{VBench dimension coverage.} Number of prompts carrying each dimension
label in the $946$-prompt set and in our $100$-prompt subset. A prompt may carry several
labels, so columns do not sum to the prompt count.}
\label{tab:vbench_dim_coverage}
\begin{tabular}{lcc}
\toprule
VBench dimension & Population ($n{=}946$) & $n{=}100$ subset \\
\midrule
human\_action           & 100 (10.6\%) & 10 (10\%) \\
temporal\_style         & 100 (10.6\%) & 12 (12\%) \\
overall\_consistency    & 93 (9.8\%)   & 9 (9\%) \\
aesthetic\_quality      & 93 (9.8\%)   & 9 (9\%) \\
imaging\_quality        & 93 (9.8\%)   & 9 (9\%) \\
appearance\_style       & 90 (9.5\%)   & 9 (9\%) \\
scene                   & 86 (9.1\%)   & 8 (8\%) \\
background\_consistency & 86 (9.1\%)   & 8 (8\%) \\
color                   & 85 (9.0\%)   & 8 (8\%) \\
spatial\_relationship   & 84 (8.9\%)   & 10 (10\%) \\
multiple\_objects       & 82 (8.7\%)   & 9 (9\%) \\
object\_class           & 79 (8.4\%)   & 9 (9\%) \\
temporal\_flickering    & 75 (7.9\%)   & 8 (8\%) \\
subject\_consistency    & 72 (7.6\%)   & 8 (8\%) \\
dynamic\_degree         & 72 (7.6\%)   & 8 (8\%) \\
motion\_smoothness      & 72 (7.6\%)   & 8 (8\%) \\
\bottomrule
\end{tabular}
\end{table}

\subsection{Jump guards}
\label{app:gate}

\Cref{eq:jump} stretches the step at every cached node, since a cached node has $A_i < \tau$
and therefore $H_i > 0$. Two guards override it and make the node take the scheduled step
$\Delta_i$ instead: the opening stretch, $t_i < 0.1$, and a large probe change at the node
itself, $\rho_i > 0.18$ with $\rho_i$ as in \cref{eq:proxy}. The two thresholds are the same
for every model and resolution in the paper. On FLUX at $\tau = 0.8$ over $100$ DrawBench prompts, $1{,}103$ of the $5{,}000$ scheduled
nodes are cached but held to the scheduled step instead of jumping, $800$ by the opening
stretch, $302$ by the large-change guard and one whose stretched step would be less
than $1\%$ longer, which the sampler rounds to $\Delta_i$.
Turning both guards off in \cref{tab:ablation_design} leaves LPIPS within noise at every
matched speed and moves PSNR by $+0.59$~dB at $3\times$ and $-0.66$~dB at $7\times$.

\subsection{Re-spacing geometry}
\label{app:regrid_geometry}

The rebuilt tail of \cref{eq:tail_regrid} is spaced uniformly in $t$. The alternative is to space
it the way the backbone's scheduler spaces its own grid. Diffusers schedulers build a grid from
uniform points $u$ through the shift map $w_s(u) = su/\big(1+(s-1)u\big)$ on the noise level
$1-t$, with a shift $s$ fixed by the checkpoint ($s=e^{1.15}$ for FLUX at our
resolution). Applying the same map to the tail gives
\begin{equation}
    1 - t_{i+1+k} = w_s\!\left(\Big(1 - \frac{k}{m_i - 1}\Big)\, w_s^{-1}\big(1 - t_{i+1}\big)\right),
    \qquad k = 0, \ldots, m_i - 1 ,
    \label{eq:tail_regrid_shift}
\end{equation}
which unwarps the landing node, spaces $m_i$ nodes evenly between it and $t=1$, and warps them
back. At $s=1$ the map is the identity and \cref{eq:tail_regrid_shift} is \cref{eq:tail_regrid}.

On FLUX we compare the two on DPG-50 with the protocol of \cref{app:ablation_extra}. They differ
by at most $0.26$~dB PSNR, at $3\times$ in favour of \cref{eq:tail_regrid}, and the sign of the
difference flips between speeds on PSNR and on the perceptual metrics alike. With no consistent
winner we keep \cref{eq:tail_regrid} on every backbone.

\section{Additional quantitative results}
\label{app:grp_results}
\subsection{Full tables with efficiency columns}
\label{app:main_tables}

\Cref{tab:flux_drawbench,tab:hunyuan_vbench} add latency, TFLOPs and peak memory to
\cref{tab:flux_drawbench_main,tab:hunyuan_vbench_main}, and \cref{tab:wan_vbench} gives the
Wan2.1 comparison that \cref{sec:results_video} summarizes. The first row of each table is the
uncached reference: it carries efficiency figures and a dash in every quality column, since a
sample has no quality against itself. Peak memory is set by the VAE decode for every method
except TaylorSeer, which stores Taylor terms for every block and peaks $4.1$~GB above the rest
on FLUX.

Sampling with fewer uncached Euler steps does not reach the same fidelity. Scored against the
same 50-step reference, $13$, $9$ and $7$ steps, each within a few percent of the speed of a
\method band in \cref{tab:flux_drawbench}, reach $14.64$, $13.90$ and $13.58$~dB PSNR against
$23.20$, $20.05$ and $18.58$~dB for \method, and trail it on SSIM, LPIPS and DreamSim as well.
Even $15$ steps, slower than every \method run, reaches only $15.05$~dB.

\begin{table}[ht]
\centering
\small
\setlength{\tabcolsep}{5pt}
\renewcommand{\arraystretch}{1.15}
\caption{\textbf{FLUX DrawBench-200.} Each block groups the runs that land nearest one speed band,
and the speedup each method actually reached is given per row.
All four metrics measure fidelity to the uncached FLUX sample for the same prompt and seed.
The efficiency columns add the latency, TFLOPs and peak memory behind each speedup.}
\label{tab:flux_drawbench}
\resizebox{\linewidth}{!}{%
\begin{tabular}{lcccccccc}
\toprule
  \multirow{2}{*}{Method} & \multicolumn{4}{c}{Efficiency} & \multicolumn{4}{c}{Quality} \\
  \cmidrule(lr){2-5} \cmidrule(lr){6-9}
   & Speedup & Latency (s) ($\downarrow$) & TFLOPs ($\downarrow$) & Mem (GB) ($\downarrow$) & PSNR ($\uparrow$) & SSIM ($\uparrow$) & LPIPS ($\downarrow$) & DreamSim ($\downarrow$) \\
\midrule
  FLUX & $1.00\times$ & 15.49 & 3735 & 33.8 & $-$ & $-$ & $-$ & $-$ \\
\midrule
  TeaCache~\citep{liu2025teacache} & $4.02\times$ & 3.86 & 931 & 33.8 & 16.02 & 0.660 & 0.396 & 0.140 \\
  TaylorSeer~\citep{liu2025taylorseer} & $3.70\times$ & 4.18 & 760 & 37.9 & 15.50 & 0.637 & 0.397 & 0.128 \\
  SeaCache~\citep{chung2026seacache} & $3.25\times$ & 4.77 & 1132 & 33.8 & 22.41 & 0.830 & 0.173 & 0.037 \\
  DPCache~\citep{cui2026dpcache} & $3.49\times$ & 4.44 & 983 & 34.0 & 21.27 & 0.803 & 0.182 & 0.041 \\
  \cellcolor{gray!10}\method & \cellcolor{gray!10}$3.79\times$ & \cellcolor{gray!10}4.08 & \cellcolor{gray!10}983 & \cellcolor{gray!10}33.8 & \cellcolor{gray!10}\textbf{23.20} & \cellcolor{gray!10}\textbf{0.847} & \cellcolor{gray!10}\textbf{0.146} & \cellcolor{gray!10}\textbf{0.028} \\
\midrule
  TeaCache~\citep{liu2025teacache} & $5.28\times$ & 2.93 & 686 & 33.8 & 15.19 & 0.619 & 0.466 & 0.193 \\
  TaylorSeer~\citep{liu2025taylorseer} & $5.59\times$ & 2.77 & 388 & 37.9 & 11.44 & 0.448 & 0.695 & 0.385 \\
  SeaCache~\citep{chung2026seacache} & $5.28\times$ & 2.93 & 686 & 33.8 & 18.81 & 0.724 & 0.326 & 0.088 \\
  DPCache~\citep{cui2026dpcache} & $4.74\times$ & 3.27 & 686 & 34.0 & 18.05 & 0.700 & 0.305 & 0.086 \\
  \cellcolor{gray!10}\method & \cellcolor{gray!10}$5.44\times$ & \cellcolor{gray!10}2.85 & \cellcolor{gray!10}685 & \cellcolor{gray!10}33.8 & \cellcolor{gray!10}\textbf{20.05} & \cellcolor{gray!10}\textbf{0.770} & \cellcolor{gray!10}\textbf{0.251} & \cellcolor{gray!10}\textbf{0.060} \\
\midrule
  TeaCache~\citep{liu2025teacache} & $5.91\times$ & 2.62 & 579 & 33.8 & 14.73 & 0.596 & 0.516 & 0.244 \\
  TaylorSeer~\citep{liu2025taylorseer} & $6.22\times$ & 2.49 & 314 & 37.9 & 10.34 & 0.363 & 0.824 & 0.579 \\
  SeaCache~\citep{chung2026seacache} & $6.66\times$ & 2.32 & 537 & 33.8 & 17.46 & 0.673 & 0.409 & 0.131 \\
  DPCache~\citep{cui2026dpcache} & $6.50\times$ & 2.38 & 462 & 34.0 & 14.83 & 0.537 & 0.533 & 0.206 \\
  \cellcolor{gray!10}\method & \cellcolor{gray!10}$6.91\times$ & \cellcolor{gray!10}2.24 & \cellcolor{gray!10}536 & \cellcolor{gray!10}33.8 & \cellcolor{gray!10}\textbf{18.58} & \cellcolor{gray!10}\textbf{0.710} & \cellcolor{gray!10}\textbf{0.327} & \cellcolor{gray!10}\textbf{0.087} \\
\bottomrule
\end{tabular}}
\end{table}

\begin{table}[ht]
\centering
\small
\setlength{\tabcolsep}{5pt}
\renewcommand{\arraystretch}{1.15}
\caption{\textbf{HunyuanVideo VBench 480p.} Blocks are three increasing speed bands.
All four metrics measure fidelity to the uncached HunyuanVideo clip for the same prompt and seed,
averaged over all $65$ frames. The efficiency columns add the latency, TFLOPs and peak memory behind each speedup.}
\label{tab:hunyuan_vbench}
\resizebox{\linewidth}{!}{%
\begin{tabular}{lcccccccc}
\toprule
  \multirow{2}{*}{Method} & \multicolumn{4}{c}{Efficiency} & \multicolumn{4}{c}{Quality} \\
  \cmidrule(lr){2-5} \cmidrule(lr){6-9}
   & Speedup & Latency (s) ($\downarrow$) & TFLOPs ($\downarrow$) & Mem (GB) ($\downarrow$) & PSNR ($\uparrow$) & SSIM ($\uparrow$) & LPIPS ($\downarrow$) & DreamSim ($\downarrow$) \\
\midrule
  HunyuanVideo & $1.00\times$ & 212.99 & 30579 & 55.5 & $-$ & $-$ & $-$ & $-$ \\
\midrule
  SeaCache~\citep{chung2026seacache} & $3.53\times$ & 60.26 & 8906 & 55.5 & 23.63 & 0.778 & 0.203 & 0.102 \\
  DPCache~\citep{cui2026dpcache} & $3.28\times$ & 64.95 & 9744 & 56.2 & 23.70 & 0.786 & 0.182 & 0.092 \\
  \cellcolor{gray!10}\method & \cellcolor{gray!10}$3.52\times$ & \cellcolor{gray!10}60.52 & \cellcolor{gray!10}9114 & \cellcolor{gray!10}55.5 & \cellcolor{gray!10}\textbf{26.48} & \cellcolor{gray!10}\textbf{0.843} & \cellcolor{gray!10}\textbf{0.141} & \cellcolor{gray!10}\textbf{0.067} \\
\midrule
  SeaCache~\citep{chung2026seacache} & $5.33\times$ & 39.98 & 6172 & 55.5 & 21.82 & 0.713 & 0.292 & 0.162 \\
  DPCache~\citep{cui2026dpcache} & $5.42\times$ & 39.33 & 6172 & 56.2 & 20.34 & 0.682 & 0.293 & 0.162 \\
  \cellcolor{gray!10}\method & \cellcolor{gray!10}$5.39\times$ & \cellcolor{gray!10}39.49 & \cellcolor{gray!10}6170 & \cellcolor{gray!10}55.5 & \cellcolor{gray!10}\textbf{22.36} & \cellcolor{gray!10}\textbf{0.742} & \cellcolor{gray!10}\textbf{0.246} & \cellcolor{gray!10}\textbf{0.129} \\
\midrule
  SeaCache~\citep{chung2026seacache} & $6.11\times$ & 34.84 & 5577 & 55.5 & 20.38 & 0.667 & 0.349 & 0.205 \\
  DPCache~\citep{cui2026dpcache} & $6.07\times$ & 35.11 & 5577 & 56.2 & 20.06 & 0.660 & 0.310 & 0.172 \\
  \cellcolor{gray!10}\method & \cellcolor{gray!10}$6.05\times$ & \cellcolor{gray!10}35.20 & \cellcolor{gray!10}5574 & \cellcolor{gray!10}55.5 & \cellcolor{gray!10}\textbf{22.27} & \cellcolor{gray!10}\textbf{0.739} & \cellcolor{gray!10}\textbf{0.267} & \cellcolor{gray!10}\textbf{0.141} \\
\bottomrule
\end{tabular}}
\end{table}

\paragraph{Wan2.1.} At $3.83\times$ \method is faster than both SeaCache at $3.73\times$ and
DPCache at $3.60\times$ and leads all four metrics, by $1.06$~dB PSNR and $0.020$ DreamSim
over SeaCache and by $2.75$~dB and $0.033$ over DPCache. At the faster band \method again
leads all four metrics, by $0.34$~dB PSNR, $0.066$ LPIPS and $0.041$ DreamSim over SeaCache,
while its $6.12\times$ run is $1.15\times$ faster than SeaCache's $5.30\times$ and faster
still than DPCache's $5.07\times$. DPCache does not hold up at this budget, falling from
$23.20$~dB at $3.60\times$ to $15.05$~dB and $0.401$ DreamSim at $5.07\times$.

\begin{table}[ht]
\centering
\small
\setlength{\tabcolsep}{5pt}
\renewcommand{\arraystretch}{1.15}
\caption{\textbf{Wan2.1-T2V-1.3B, VBench 100-prompt subset.} Blocks are two increasing speed bands.
All four metrics measure fidelity to the uncached Wan2.1 clip for the same prompt and seed.}
\label{tab:wan_vbench}
\resizebox{\linewidth}{!}{%
\begin{tabular}{lcccccccc}
\toprule
  \multirow{2}{*}{Method} & \multicolumn{4}{c}{Efficiency} & \multicolumn{4}{c}{Quality} \\
  \cmidrule(lr){2-5} \cmidrule(lr){6-9}
   & Speedup & Latency (s) ($\downarrow$) & TFLOPs ($\downarrow$) & Mem (GB) ($\downarrow$) & PSNR ($\uparrow$) & SSIM ($\uparrow$) & LPIPS ($\downarrow$) & DreamSim ($\downarrow$) \\
\midrule
  Wan2.1 & $1.00\times$ & 440.02 & 28578 & 6.7 & $-$ & $-$ & $-$ & $-$ \\
\midrule
  SeaCache~\citep{chung2026seacache} & $3.73\times$ & 117.93 & 7771 & 7.2 & 24.89 & 0.838 & 0.139 & 0.058 \\
  DPCache~\citep{cui2026dpcache}$^\dagger$ & $3.60\times$ & 122.35 & 7209 & 5.9 & 23.20 & 0.791 & 0.155 & 0.071 \\
  \cellcolor{gray!10}\method & \cellcolor{gray!10}$3.83\times$ & \cellcolor{gray!10}115.03 & \cellcolor{gray!10}7770 & \cellcolor{gray!10}7.1 & \cellcolor{gray!10}\textbf{25.95} & \cellcolor{gray!10}\textbf{0.869} & \cellcolor{gray!10}\textbf{0.102} & \cellcolor{gray!10}\textbf{0.038} \\
\midrule
  SeaCache~\citep{chung2026seacache} & $5.30\times$ & 83.03 & 5522 & 7.2 & 21.17 & 0.736 & 0.262 & 0.144 \\
  DPCache~\citep{cui2026dpcache}$^\dagger$ & $5.07\times$ & 86.73 & 4960 & 5.9 & 15.05 & 0.566 & 0.470 & 0.401 \\
  \cellcolor{gray!10}\method & \cellcolor{gray!10}$6.12\times$ & \cellcolor{gray!10}71.91 & \cellcolor{gray!10}4958 & \cellcolor{gray!10}7.1 & \cellcolor{gray!10}\textbf{21.51} & \cellcolor{gray!10}\textbf{0.776} & \cellcolor{gray!10}\textbf{0.196} & \cellcolor{gray!10}\textbf{0.103} \\
\bottomrule
\end{tabular}}
\vspace{2pt}
\footnotesize $^\dagger$DPCache releases Wan2.1 code for image-to-video only. These rows are our port to the text-to-video pipeline, recalibrated as in \cref{app:setup_details} and checked against the uncached reference, which it reproduces byte for byte with caching disabled.
\end{table}

\subsection{Per-prompt quality against achieved speedup}
\label{app:scatter}

\Cref{sec:results_image,sec:results_video} report means over the prompt set. \Cref{fig:speedup_scattered_flux,%
fig:speedup_scattered_hunyuan} resolve the same comparison per prompt, plotting every prompt at
its own achieved speedup rather than at the band mean, so the spread behind each reported mean is
visible. The FLUX figure shows all five methods of \cref{tab:flux_drawbench_main}. The
HunyuanVideo figure shows SeaCache, DPCache and \method, so TeaCache and TaylorSeer appear only on
FLUX. The clouds overlap heavily at every operating point, which is what the same prompts and
seeds decoded under different caches should produce, and the means separate anyway.

\begin{figure}[t]
  \centering
  \includegraphics[width=0.7\linewidth]{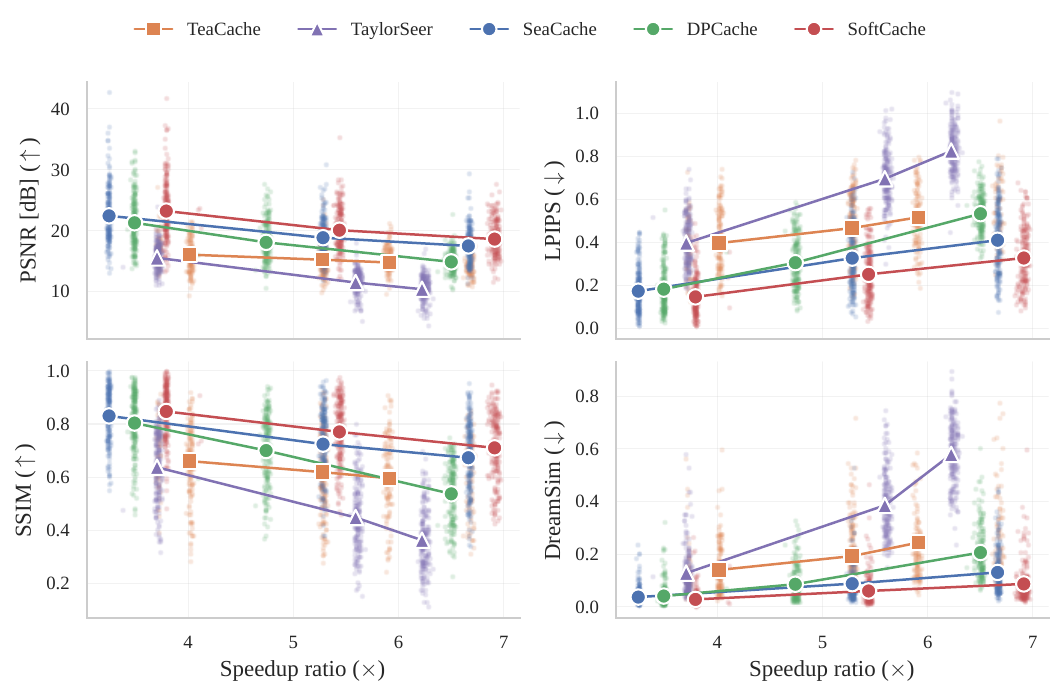}
  \caption{\textbf{FLUX DrawBench-200.} The four fidelity metrics against achieved speedup for
  TeaCache, TaylorSeer, SeaCache, DPCache and \method. Each faint dot is one prompt. Solid markers
  are per-method means, connected across the three speed bands of \cref{app:main_tables}.}
  \label{fig:speedup_scattered_flux}
\end{figure}

\begin{figure}[t]
  \centering
  \includegraphics[width=0.7\linewidth]{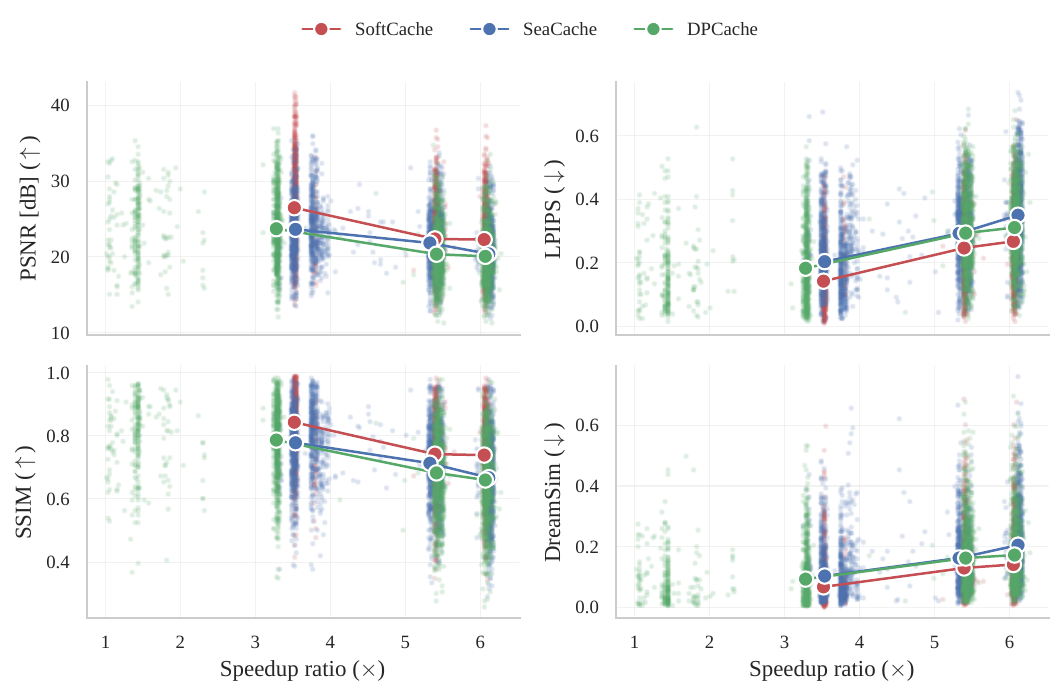}
  \caption{\textbf{HunyuanVideo VBench 480p.} The four fidelity metrics against achieved speedup.
  Each faint dot is one prompt. Solid markers are per-method means, connected across the three
  speed bands of \cref{app:main_tables}. \method holds the best mean PSNR, LPIPS, SSIM and
  DreamSim at all three.}
  \label{fig:speedup_scattered_hunyuan}
\end{figure}

\subsection{Reference-free metrics}
\label{app:unused_metrics}

Our main tables measure fidelity to the uncached sample generated from the same noise and prompt,
since a cache approximates the uncached sampler.
Reference-free metrics instead judge a generated image or video by its own quality,
independent of any other sample.
We report two of them here, ImageReward~\citep{xu2023imagereward} for images and
VBench~\citep{huang2023vbench} for video.

\begin{figure}[t]
  \centering
  \includegraphics{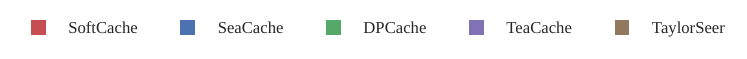}\\[2pt]
  \begin{subfigure}[b]{0.54\linewidth}
    \centering
    \includegraphics{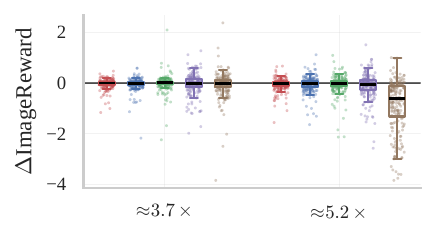}
    \caption{}
    \label{fig:imagereward_paired_a}
  \end{subfigure}\hfill
  \begin{subfigure}[b]{0.44\linewidth}
    \centering
    \includegraphics{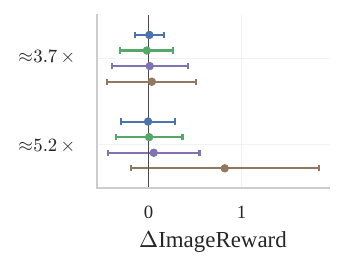}
    \caption{}
    \label{fig:imagereward_paired_b}
  \end{subfigure}
  \caption{ImageReward on DrawBench-200 with FLUX (one seed, $200$ prompts), paired by prompt.
(a)~Per-prompt change relative to the uncached image, for each cache at the $\approx\!3.7\times$
and $\approx\!5.2\times$ bands of \cref{tab:flux_drawbench_main}.
(b)~Mean per-prompt difference between \method and each baseline in the same band, with bars at
one standard deviation of the per-prompt differences.
ImageReward separates the methods only where the fidelity metrics already do: TaylorSeer at the
$\approx\!5.2\times$ band collapses, and every other difference is small against its spread.}
  \label{fig:imagereward_paired}
\end{figure}

ImageReward (package $1.5$, ImageReward-v1.0 weights) scores human preference and prompt
alignment.
We score FLUX on DrawBench-200 for the uncached sampler and for every method of
\cref{tab:flux_drawbench_main} at the $\approx\!3.7\times$ and $\approx\!5.2\times$ bands.
ImageReward separates the methods only where the fidelity metrics already do.
TaylorSeer at the $\approx\!5.2\times$ band loses $0.89$ against the uncached sampler, more than the across-prompt
standard deviation of $0.79$, on the run that \cref{tab:flux_drawbench_main} scores at
$11.44$~dB PSNR. Every other cache lowers the mean by $0.02$ to $0.12$
(\cref{fig:imagereward_paired}a).
Paired by prompt, \method scores above TaylorSeer at the $\approx\!5.2\times$ band on $81\%$ of prompts
(Wilcoxon signed-rank, $p < 10^{-20}$), and the same test finds no difference in the other seven
comparisons ($p \geq 0.14$), where the mean difference is below $0.06$
against a standard deviation of $0.49$ in the per-prompt difference and \method scores higher
on $45\%$ to $55\%$ of prompts (\cref{fig:imagereward_paired}b).
ImageReward therefore adds nothing to \cref{tab:flux_drawbench_main} that the fidelity metrics
do not already show, and we do not use it to rank caches.

VBench scores a video along $16$ quality and semantic dimensions.
We score the HunyuanVideo samples of \cref{tab:hunyuan_vbench_main} on all $16$, each on the
prompts VBench tags for it, which together cover all $946$ prompts.
The VBench protocol samples five videos per prompt, and we generate one, with seed $0$, within our
compute budget. Totals are therefore lower than published ones for every method, the uncached
model included ($0.791$ against the $0.809$ Total that \citet{cui2026dpcache} report), and we
compare our runs only with each other (\cref{tab:vbench16_quality,tab:vbench16_semantic}).

On the Total \method is ahead of DPCache by $0.001$ at the slowest band and $0.003$ at the
fastest, behind by $0.001$ in the middle, and ahead of SeaCache in all three.
DPCache is ahead of \method on aesthetic quality, imaging quality, object class and temporal
style in all three bands, by up to $0.039$, and \method is ahead on background consistency,
temporal flickering and appearance style in all three, by up to $0.016$, while the remaining
dimensions change sides between bands.
The two sets offset each other in the Total, and neither measures fidelity to the uncached clip,
which \cref{tab:hunyuan_vbench_main} separates by $2.78$~dB PSNR at the slowest band.

\begin{table}[ht]
\centering
\small
\setlength{\tabcolsep}{5pt}
\renewcommand{\arraystretch}{1.15}
\caption{\textbf{HunyuanVideo VBench quality dimensions.}}
\label{tab:vbench16_quality}
\resizebox{\linewidth}{!}{%
\begin{tabular}{lccccccccc}
\toprule
  Method & Speedup & Subj. ($\uparrow$) & Bkg. ($\uparrow$) & Flicker ($\uparrow$) & Smooth ($\uparrow$) & Aesth. ($\uparrow$) & Imag. ($\uparrow$) & Dynamic ($\uparrow$) & Quality ($\uparrow$) \\
\midrule
  HunyuanVideo & $1.00\times$ & 0.963 & 0.965 & 0.991 & 0.992 & 0.601 & 0.619 & 0.347 & 0.810 \\
\midrule
  SeaCache~\citep{chung2026seacache} & $3.53\times$ & 0.962 & 0.968 & 0.991 & 0.992 & 0.589 & 0.591 & 0.306 & 0.801 \\
  DPCache$^\dagger$~\citep{cui2026dpcache} & $3.28\times$ & 0.964 & 0.964 & 0.991 & 0.993 & 0.595 & 0.607 & 0.319 & 0.805 \\
  \cellcolor{gray!10}\method & \cellcolor{gray!10}$3.52\times$ & \cellcolor{gray!10}0.964 & \cellcolor{gray!10}0.968 & \cellcolor{gray!10}0.992 & \cellcolor{gray!10}0.993 & \cellcolor{gray!10}0.592 & \cellcolor{gray!10}0.597 & \cellcolor{gray!10}0.333 & \cellcolor{gray!10}0.805 \\
\midrule
  SeaCache~\citep{chung2026seacache} & $5.33\times$ & 0.960 & 0.970 & 0.992 & 0.992 & 0.575 & 0.544 & 0.306 & 0.791 \\
  DPCache$^\dagger$~\citep{cui2026dpcache} & $5.42\times$ & 0.961 & 0.959 & 0.989 & 0.991 & 0.589 & 0.597 & 0.292 & 0.797 \\
  \cellcolor{gray!10}\method & \cellcolor{gray!10}$5.39\times$ & \cellcolor{gray!10}0.960 & \cellcolor{gray!10}0.968 & \cellcolor{gray!10}0.991 & \cellcolor{gray!10}0.992 & \cellcolor{gray!10}0.587 & \cellcolor{gray!10}0.578 & \cellcolor{gray!10}0.333 & \cellcolor{gray!10}0.800 \\
\midrule
  SeaCache~\citep{chung2026seacache} & $6.11\times$ & 0.955 & 0.969 & 0.992 & 0.992 & 0.562 & 0.508 & 0.278 & 0.781 \\
  DPCache$^\dagger$~\citep{cui2026dpcache} & $6.07\times$ & 0.956 & 0.953 & 0.987 & 0.990 & 0.586 & 0.598 & 0.208 & 0.786 \\
  \cellcolor{gray!10}\method & \cellcolor{gray!10}$6.05\times$ & \cellcolor{gray!10}0.958 & \cellcolor{gray!10}0.969 & \cellcolor{gray!10}0.992 & \cellcolor{gray!10}0.992 & \cellcolor{gray!10}0.580 & \cellcolor{gray!10}0.564 & \cellcolor{gray!10}0.319 & \cellcolor{gray!10}0.796 \\
\bottomrule
\end{tabular}}
\vspace{2pt}
\footnotesize $^\dagger$One sample per prompt (seed $0$), for compute reasons, against five in the VBench protocol, so totals are not comparable with published scores: DPCache reports $0.8035$ and $0.8023$ against an uncached $0.8093$ under the five-sample protocol~\citep{cui2026dpcache}.
\end{table}

\begin{table}[ht]
\centering
\small
\setlength{\tabcolsep}{3pt}
\renewcommand{\arraystretch}{1.15}
\caption{\textbf{HunyuanVideo VBench semantic dimensions and total.}}
\label{tab:vbench16_semantic}
\resizebox{\linewidth}{!}{%
\begin{tabular}{lccccccccccc}
\toprule
  Method & Object ($\uparrow$) & Multi-obj. ($\uparrow$) & Action ($\uparrow$) & Color ($\uparrow$) & Spatial ($\uparrow$) & Scene ($\uparrow$) & Appear. ($\uparrow$) & Temporal ($\uparrow$) & Overall ($\uparrow$) & Semantic ($\uparrow$) & Total ($\uparrow$) \\
\midrule
  HunyuanVideo & 0.820 & 0.658 & 0.900 & 0.856 & 0.621 & 0.416 & 0.198 & 0.237 & 0.267 & 0.715 & 0.791 \\
\midrule
  SeaCache~\citep{chung2026seacache} & 0.860 & 0.683 & 0.910 & 0.888 & 0.603 & 0.403 & 0.200 & 0.237 & 0.264 & 0.723 & 0.785 \\
  DPCache$^\dagger$~\citep{cui2026dpcache} & 0.843 & 0.704 & 0.910 & 0.863 & 0.586 & 0.388 & 0.197 & 0.239 & 0.264 & 0.716 & 0.787 \\
  \cellcolor{gray!10}\method & \cellcolor{gray!10}0.832 & \cellcolor{gray!10}0.707 & \cellcolor{gray!10}0.920 & \cellcolor{gray!10}0.900 & \cellcolor{gray!10}0.600 & \cellcolor{gray!10}0.389 & \cellcolor{gray!10}0.199 & \cellcolor{gray!10}0.234 & \cellcolor{gray!10}0.263 & \cellcolor{gray!10}0.722 & \cellcolor{gray!10}\textbf{0.788} \\
\midrule
  SeaCache~\citep{chung2026seacache} & 0.779 & 0.606 & 0.890 & 0.846 & 0.574 & 0.355 & 0.206 & 0.227 & 0.261 & 0.687 & 0.771 \\
  DPCache$^\dagger$~\citep{cui2026dpcache} & 0.831 & 0.668 & 0.910 & 0.874 & 0.619 & 0.392 & 0.199 & 0.238 & 0.264 & 0.717 & \textbf{0.781} \\
  \cellcolor{gray!10}\method & \cellcolor{gray!10}0.792 & \cellcolor{gray!10}0.681 & \cellcolor{gray!10}0.910 & \cellcolor{gray!10}0.842 & \cellcolor{gray!10}0.572 & \cellcolor{gray!10}0.365 & \cellcolor{gray!10}0.202 & \cellcolor{gray!10}0.234 & \cellcolor{gray!10}0.264 & \cellcolor{gray!10}0.701 & \cellcolor{gray!10}0.780 \\
\midrule
  SeaCache~\citep{chung2026seacache} & 0.762 & 0.547 & 0.910 & 0.845 & 0.538 & 0.369 & 0.209 & 0.225 & 0.258 & 0.679 & 0.760 \\
  DPCache$^\dagger$~\citep{cui2026dpcache} & 0.813 & 0.633 & 0.910 & 0.871 & 0.606 & 0.403 & 0.200 & 0.237 & 0.264 & 0.711 & 0.771 \\
  \cellcolor{gray!10}\method & \cellcolor{gray!10}0.799 & \cellcolor{gray!10}0.608 & \cellcolor{gray!10}0.930 & \cellcolor{gray!10}0.831 & \cellcolor{gray!10}0.554 & \cellcolor{gray!10}0.332 & \cellcolor{gray!10}0.205 & \cellcolor{gray!10}0.229 & \cellcolor{gray!10}0.262 & \cellcolor{gray!10}0.688 & \cellcolor{gray!10}\textbf{0.774} \\
\bottomrule
\end{tabular}}
\vspace{2pt}
\footnotesize $^\dagger$One sample per prompt, see the note to \cref{tab:vbench16_quality}.
\end{table}

\section{Additional ablations and analysis}
\label{app:grp_ablations}
\subsection{Headroom and reuse error}
\label{app:headroom_drift}

\Cref{fig:jump_size_a} is measured on uncached FLUX trajectories, in the
setting of \cref{app:setup_details}, on $40$ DrawBench prompts.
Since nothing is reused, the residual is exact at every node. We can
therefore pair every node with every later one, treating the earlier node
as the anchor, which gives $1{,}128$ pairs per prompt. For each pair we
compute the reuse error $E_i$ and the accumulator $A_i$. The first and last
nodes are excluded as boundary effects.
Panel (a) keeps the pairs a hard cache would reuse, those with $A_i$ below
the threshold, and plots their error against the accumulator. Error grows
with the accumulator: a node about to trigger a refresh ($A_i\approx\tau$)
has $2.2\times$ the median error of one just after a refresh
($A_i\approx0$), although a threshold treats the two alike. This is the
factor quoted in \cref{sec:jump}.
Since both the error and the accumulator grow with the distance from the
anchor, we also check that $A_i$ is more than a step counter: at a fixed gap
$i-a$ of $2$ to $12$ nodes, $A_i$ still orders the pairs by error with a
Spearman correlation of about $0.8$, computed per prompt and averaged.

\subsection{Substituting the staleness probe}
\label{app:probe}

\method assumes nothing of the probe $\rho_i$ of \cref{eq:proxy} beyond
$\rho_i \geq 0$ and the reset of $A_i$ at a refresh, so either dynamic cache can
host it. We use SeaCache's filtered probe~\citep{chung2026seacache} throughout.
Here we swap in TeaCache's polynomial-rescaled probe~\citep{liu2025teacache} for
both the refresh decision and the headroom, hold the rest of \method fixed
($J{=}8$, $\beta{=}1$, tail re-spacing, the $\tau$ grid), and score both
configurations at matched wall-clock budget against both caches on DPG-50
(\cref{tab:ablation_probe}).

The policy transfers. On TeaCache's probe, \method improves on TeaCache at every
budget and degrades it at none: by $2.92$ and $1.32$~dB of PSNR at the two lower
budgets, and by LPIPS at the two higher ones, where PSNR is level at $0.29$ and
$0.25$~dB. The host
sets the ceiling. The same policy on TeaCache's probe lands $3.6$ to $4.2$~dB
below plain SeaCache, while on SeaCache's probe it is at or above SeaCache and
$3.9$ to $10.0$~dB above TeaCache. We therefore read this experiment as evidence
that jump scheduling and residual motion are portable, not as a comparison between
the two caches. It uses one seed and $50$ prompts on FLUX, and substitutes the gate
and the headroom together, so it does not separate which use of $A_i$ is
responsible.

\begin{table}[t]
  \centering
  \small
  \setlength{\tabcolsep}{5pt}
  \begin{tabular}{c ccc c cccc}
    \toprule
    & \multicolumn{3}{c}{\method, SeaCache probe} & & \multicolumn{4}{c}{\method, TeaCache probe} \\
    \cmidrule(lr){2-4} \cmidrule(lr){6-9}
    $\tau$ & Speedup & $\Delta$PSNR & $\Delta$PSNR & & Speedup & $\Delta$PSNR & $\Delta$PSNR & $\Delta$LPIPS \\
           &         & vs.\ SeaCache & vs.\ TeaCache & &        & vs.\ SeaCache & vs.\ TeaCache & vs.\ TeaCache \\
    \midrule
    0.3 & 2.95$\times$ & $+2.93$           & $+10.00$ & & 3.13$\times$ & $-3.61$           & $\mathbf{+2.92}$  & $+0.138$ \\
    0.5 & 3.80$\times$ & $+2.09$           & $+7.75$  & & 4.44$\times$ & $-3.77$           & $\mathbf{+1.32}$  & $+0.094$ \\
    0.8 & 5.43$\times$ & $+0.41$           & $+5.12$  & & 6.09$\times$ & $-4.16$           & $+0.29$           & $+0.079$ \\
    1.1 & 6.89$\times$ & $+0.04$           & $+3.94$  & & 7.18$\times$ & --                & $+0.25$           & $+0.037$ \\
    \bottomrule
  \end{tabular}
  \caption{\textbf{\method on either probe.} \method driven by the SeaCache probe
  (used throughout the paper) or the TeaCache probe, scored at matched wall-clock
  budget against both caches. Positive values favour \method. Means over 50 DPG
  prompts, FLUX.1-dev, 50 steps. A dash marks a budget beyond the SeaCache
  frontier, where the comparison would be extrapolated.}
  \label{tab:ablation_probe}
\end{table}

\subsection{Cap sweep and the blind-pick protocol}
\label{app:cap_dial}

The cap ladder of \cref{tab:ablation_design} fixes $J{=}8$ on DPG-50. Smaller caps ($J{=}3$,
$J{=}5$) are perceptually worse from $5\times$ on.
Larger caps are a wash through $5\times$, gaining PSNR and giving it back on LPIPS and DreamSim,
and win only at $7\times$, where they lose high-frequency texture (\cref{app:ablation_extra}).
$J{=}8$ is therefore the smallest cap that gives up nothing through $5\times$. The $7\times$ band
is handled by $\beta$ instead (\cref{sec:ablations}), which costs no texture. Two blind checks
below back this, both on DPG-50 only. In each, one author saw, per prompt, the uncached reference
and several unlabeled variants in a random order, with only the speedup shown.

\paragraph{\boldmath Reject test at $5\times$.} The six variants differ only in the cap,
$J\in\{1,3,5,8,12,32\}$, at matched $\approx5\times$. The rater rejected any variant visibly
worse than the reference, or marked the prompt ``can't tell''. The rule was fixed in advance:
keep $J{=}8$ if it has the lowest or tied-lowest reject rate, otherwise switch to the cap that is
strictly lower. Five prompts were ``can't tell''. On the remaining $45$, $J{=}8$ was never
rejected ($0/45$), against $2\%$ for $J{=}3$ and $J{=}5$, $22\%$ for $J{=}1$, $53\%$ for $J{=}12$
and $91\%$ for $J{=}32$, so the rule keeps $J{=}8$.

\paragraph{\boldmath Preference test at $7\times$.} Every cap $J\ge12$ and $\beta{=}0.5$ beat
\method at $\beta{=}1$ on both LPIPS and PSNR at this band (\cref{tab:ablation_design}).
The four variants, all at $6.7$--$6.9\times$, are \method at $J{=}8$, $\beta{=}1$, the same without
residual motion, $\beta{=}0.5$, and $J{=}20$. The rater picked the variant closest to the
reference, could pick several when no single one stood out, or marked ``can't tell''. Of the
$50$ prompts, $38$ received a pick and $12$ were ``can't tell''. $\beta{=}0.5$ was picked on $26$
prompts, the variant without residual motion on $15$, \method at $\beta{=}1$ on $13$ and $J{=}20$
on $8$. Head to head, $\beta{=}0.5$ beats \method at $\beta{=}1$ $19$ to $6$ (two-sided sign
test, $p{=}0.015$) and $J{=}20$ loses to it $21$ to $3$ ($p{<}0.001$). Picks showed no position
bias: the four shuffled slots received $14$, $15$, $14$ and $19$ picks.

Both checks are single-rater preference tests, not user studies.

\subsection{Additional ablation views}
\label{app:ablation_extra}

Every result here is FLUX on the $50$ DPG-Bench prompts of \cref{sec:ablations}, at the matched
wall-clock speedups of \cref{app:setup_details}.

\Cref{tab:ablation_loo} extends the leave-one-out of \cref{sec:ablations} to the $5\times$ band
and adds residual motion, with $\beta{=}1$ at $3\times$ and $\beta{=}0.5$ above it as in the
method we report. The extra band and the extra row do not change the ranking of \cref{sec:ablations}.
\method has the lowest LPIPS and DreamSim at every band, by $0.001$ LPIPS over the headroom
removal at $7\times$, and removing the jump costs the most PSNR at every band. Removing headroom
scaling scores a higher PSNR than \method at $5\times$ and $7\times$, as does removing residual
motion at $7\times$. This is the disagreement between PSNR and the perceptual metrics that
\cref{app:metric_reading} looks at.

\begin{table}[ht]
\centering
\small
\setlength{\tabcolsep}{5pt}
\renewcommand{\arraystretch}{1.15}
\caption{\textbf{Component leave-one-out (FLUX, DPG-50).} Each row removes one component of \method at matched wall-clock speedup, with $\beta{=}1$ at $3\times$ and $\beta{=}0.5$ above it.}
\label{tab:ablation_loo}
\begin{tabular}{lccccccccc}
\toprule
  \multirow{2}{*}{} & \multicolumn{3}{c}{LPIPS ($\downarrow$)} & \multicolumn{3}{c}{DreamSim ($\downarrow$)} & \multicolumn{3}{c}{PSNR ($\uparrow$)} \\
  \cmidrule(lr){2-4} \cmidrule(lr){5-7} \cmidrule(lr){8-10}
   & $3\times$ & $5\times$ & $7\times$ & $3\times$ & $5\times$ & $7\times$ & $3\times$ & $5\times$ & $7\times$ \\
\midrule
  w/o jumps ($J{=}1$) & 0.103 & 0.245 & 0.548 & 0.018 & 0.048 & 0.199 & 24.30 & 18.88 & 13.67 \\
  w/o headroom ($H_i{=}1$) & 0.122 & 0.215 & 0.330 & 0.016 & 0.041 & 0.082 & 25.70 & \textbf{21.36} & 18.17 \\
  w/o re-spacing ($m_i$ fixed) & 0.109 & 0.231 & 0.371 & 0.018 & 0.045 & 0.101 & 24.64 & 19.65 & 16.64 \\
  w/o residual motion ($\beta{=}0$) & 0.117 & 0.246 & 0.363 & 0.018 & 0.051 & 0.087 & 25.50 & 20.24 & \textbf{18.28} \\
  \cellcolor{gray!10}\method & \cellcolor{gray!10}\textbf{0.093} & \cellcolor{gray!10}\textbf{0.207} & \cellcolor{gray!10}\textbf{0.329} & \cellcolor{gray!10}\textbf{0.013} & \cellcolor{gray!10}\textbf{0.037} & \cellcolor{gray!10}\textbf{0.078} & \cellcolor{gray!10}\textbf{26.50} & \cellcolor{gray!10}20.61 & \cellcolor{gray!10}17.85 \\
\bottomrule
\end{tabular}
\end{table}

\Cref{tab:ablation_design} changes one setting at a time against \method at $\beta{=}1$. The
$\beta{=}0.5$ row is the band rule of \cref{sec:ablations}: it loses at $3\times$ and wins at
$7\times$. Caps below $8$ lose from $5\times$ on. Caps above $8$ trade the slower bands for the
fastest one, and what they gain there on the metrics they pay in fine texture: the correlation
of high-pass detail between the output and the uncached image falls from $0.32$ to between
$0.23$ and $0.30$ at $7\times$, while $\beta{=}0.5$ keeps it at $0.32$. Beyond $5\times$ the
reported method therefore lowers $\beta$ to $0.5$ rather than raising the cap, and
\cref{app:cap_dial} backs the choice with a blind comparison. The remaining rows change little.
A second-order term in \cref{eq:residual_motion} is no better than the straight line, at $J{=}8$
or at $J{=}20$. Removing the two jump guards of \cref{app:gate}, which hold a cached node to the
scheduled step early in sampling and when the probe changes sharply, moves LPIPS by at most
$0.022$. Spacing the rebuilt tail in the scheduler's shifted coordinate rather than uniformly in
$t$ (\cref{app:regrid_geometry}) moves it by at most $0.012$.

\begin{table}[ht]
\centering
\small
\setlength{\tabcolsep}{1.5pt}
\renewcommand{\arraystretch}{1.15}
\caption{\textbf{One setting at a time (FLUX, DPG-50).} Change against \method at $\beta{=}1$ at matched wall-clock speedup, signed so that positive is better.}
\label{tab:ablation_design}
\begin{tabular}{lcccccc}
\toprule
  \multirow{2}{*}{Variant} & \multicolumn{3}{c}{$-\Delta$LPIPS} & \multicolumn{3}{c}{$\Delta$PSNR} \\
  \cmidrule(lr){2-4} \cmidrule(lr){5-7}
   & $3\times$ & $5\times$ & $7\times$ & $3\times$ & $5\times$ & $7\times$ \\
\midrule
  $\beta{=}0.5$ & $-0.009$ & $+0.004$ & $+0.080$ & $-0.31$ & $+0.75$ & $+1.65$ \\
\midrule
  cap $J{=}3$ & $+0.005$ & $-0.020$ & $-0.085$ & $-0.47$ & $-0.63$ & $-1.28$ \\
  cap $J{=}5$ & $+0.004$ & $-0.014$ & $-0.067$ & $-0.50$ & $-0.37$ & $-1.02$ \\
  cap $J{=}12$ & $-0.006$ & $-0.014$ & $+0.021$ & $+0.83$ & $+0.18$ & $+0.47$ \\
  cap $J{=}20$ & $-0.025$ & $-0.005$ & $+0.050$ & $+0.68$ & $+1.19$ & $+1.12$ \\
  cap $J{=}32$ & $-0.051$ & $-0.023$ & $+0.065$ & $-0.08$ & $+1.47$ & $+2.30$ \\
\midrule
  residual motion, 2nd order & $-0.002$ & $-0.002$ & $-0.038$ & $-0.12$ & $-0.03$ & $-0.22$ \\
  residual motion, 2nd order, $J{=}20$ & $-0.026$ & $-0.004$ & $+0.045$ & $+0.62$ & $+1.18$ & $+1.08$ \\
\midrule
  no jump guards & $+0.006$ & $-0.022$ & $+0.002$ & $+0.59$ & $-0.21$ & $-0.66$ \\
\midrule
  tail, shifted spacing & $-0.012$ & $-0.002$ & $+0.003$ & $-0.26$ & $+0.08$ & $+0.06$ \\
\bottomrule
\end{tabular}
\end{table}

\Cref{tab:ablation_placement} asks whether it matters where the jumps fall. \method jumps
wherever the headroom allows, so every prompt gets its own positions. Each alternative instead
fixes in advance which grid nodes may jump, the same for every prompt: nodes spread evenly over
the trajectory, nodes near its end, or nodes drawn at random. A last variant never jumps. A
listed node jumps only if it is cached, so these variants jump less often and run slower than
\method, and each row is reported as its PSNR gain over SeaCache at that row's own speedup. For
the random lists we report the mean of five draws and the best one, chosen after scoring. Nodes
near the start are omitted as a list because almost none of them are cached. At $\tau{=}0.3$
\method leads every fixed list, by $1.2$~dB over the next best. At $\tau{=}0.8$ the fixed lists
trail by $0.1$ to $1.3$~dB, and never jumping costs $1.5$~dB. Because the fixed lists also jump
less often, the table shows that where the jumps fall matters, but it does not isolate the value
of choosing them per prompt.

\begin{table}[ht]
\centering
\small
\setlength{\tabcolsep}{5pt}
\renewcommand{\arraystretch}{1.15}
\caption{\textbf{Where the jumps fall (FLUX, DPG-50).} PSNR gain over SeaCache at each variant's own wall-clock speedup. Jumps is the mean number of jumps per image.}
\label{tab:ablation_placement}
\begin{tabular}{lcccccc}
\toprule
  \multirow{2}{*}{Placement} & \multicolumn{3}{c}{$\tau{=}0.3$} & \multicolumn{3}{c}{$\tau{=}0.8$} \\
  \cmidrule(lr){2-4} \cmidrule(lr){5-7}
   & Speedup & Jumps & $\Delta$PSNR & Speedup & Jumps & $\Delta$PSNR \\
\midrule
  \cellcolor{gray!10}\method & \cellcolor{gray!10}$2.91\times$ & \cellcolor{gray!10}5.0 & \cellcolor{gray!10}$+3.08$ & \cellcolor{gray!10}$5.34\times$ & \cellcolor{gray!10}5.9 & \cellcolor{gray!10}$+0.35$ \\
  Evenly spaced & $2.56\times$ & 2.0 & $+0.92$ & $4.74\times$ & 3.4 & $-0.26$ \\
  Near the end & $2.57\times$ & 3.0 & $+1.88$ & $5.25\times$ & 4.0 & $+0.10$ \\
  Random (best of 5) & $2.43\times$ & 1.0 & $+1.69$ & $5.28\times$ & 5.0 & $+0.22$ \\
  Random (mean of 5) & $2.49\times$ & 2.0 & $+0.98$ & $5.23\times$ & 4.0 & $-0.99$ \\
  No jumps & $2.31\times$ & 0.0 & $+1.79$ & $4.23\times$ & 0.0 & $-1.16$ \\
\bottomrule
\end{tabular}
\end{table}

\section{Qualitative results}
\label{app:grp_qual}
\subsection{What the fidelity metrics measure}
\label{app:metric_reading}

\begin{figure}[t]
  \centering
  \newlength{\cwframew}\setlength{\cwframew}{0.20\linewidth}%
  \newlength{\cwframesep}\setlength{\cwframesep}{4pt}%
  \newcommand{\cwlabsize}{\footnotesize}%
  \newcommand{\cwlabline}{0.5pt}%
  \newcommand{\cwlabsep}{3pt}%
  \newcommand{\cwrowsep}{4pt}%
  \newcommand{\cwmetsep}{3pt}%
  \newcommand{\cwpsnrsep}{\quad}%
  \newcommand{\cwaftermet}{7pt}%
  \renewcommand{\filmstriph}{4.5}%
  \renewcommand{\filmholew}{3.2pt}%
  \renewcommand{\filmholeh}{2.2pt}%
  \renewcommand{\filmnholes}{21}%
  \renewcommand{\filmstripcol}{black!85}%
  \renewcommand{\zoominsetfrac}{0.42}%
  \setlength{\zoompop}{2.8pt}%
  \setlength{\zoompopx}{\zoompop}\setlength{\zoompopy}{\zoompop}%
  \def\ztx{0.74}\def\zty{0.42}\def\ztw{0.18}\def\zth{0.24}\def\ztc{north west}%
  \newlength{\cwstripw}\setlength{\cwstripw}{3\cwframew}\addtolength{\cwstripw}{2\cwframesep}%
  \setlength{\tabcolsep}{0pt}%
  \renewcommand{\arraystretch}{0}%
  \def\zoommarks{}%
  \newcommand{\cwpath}[2]{figures/psnr_lies_crosswalk/#1_c#2.jpg}%
  \newcommand{\cwstrip}[1]{\marktile{\cwframew}{\cwpath{#1}{0}}\hspace{\cwframesep}%
    \zoomtile{\cwframew}{\cwpath{#1}{1}}{\ztx}{\zty}{\ztw}{\zth}{\ztc}\hspace{\cwframesep}%
    \marktile{\cwframew}{\cwpath{#1}{3}}}%
  \newlength{\cwrowh}%
  \settoheight{\cwrowh}{\includegraphics[width=\cwframew]{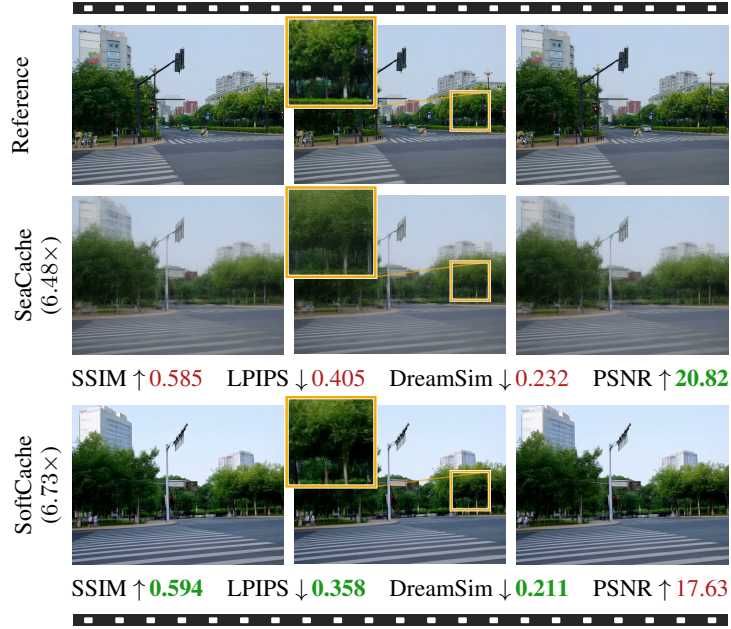}}%
  \newcommand{\cwmid}[1]{\raisebox{\dimexpr0.5\cwrowh-0.5\height\relax}{#1}}%
  \newcommand{\cwrot}[1]{\cwmid{\rotatebox{90}{{\cwlabsize #1}}}}%
  \newcommand{\cwlab}[2]{\cwrot{\shortstack[c]{#1\\[\cwlabline]($#2\times$)}}}%
  \newcommand{\cwlabref}[1]{\cwrot{\shortstack[c]{#1\\[\cwlabline]\strut}}}%
  \newcommand{\mwin}[2]{#1\,\textcolor{darkgreen}{\textbf{#2}}}%
  \newcommand{\mlose}[2]{#1\,\textcolor{darkred}{#2}}%
  \newcommand{\cwmetrics}[1]{& \makebox[\cwstripw]{\cwlabsize\rule{0pt}{2.2ex}#1}}%
  \ifdefined\tikzexternaldisable\tikzexternaldisable\fi
  \begin{tabular}{@{}c@{\hspace{\cwlabsep}}c@{}}
    & \filmsprocket{\cwstripw} \\[\cwrowsep]
    \cwlabref{Reference} & \cwstrip{reference} \\[\cwrowsep]
    \cwlab{SeaCache}{6.48} & \cwstrip{seacache} \\[\cwmetsep]
    \cwmetrics{\mlose{SSIM $\uparrow$}{0.585}\quad \mlose{LPIPS $\downarrow$}{0.405}\quad
      \mlose{DreamSim $\downarrow$}{0.232}\cwpsnrsep \mwin{PSNR $\uparrow$}{20.82}} \\[\cwaftermet]
    \cwlab{\method}{6.73} & \cwstrip{ours} \\[\cwmetsep]
    \cwmetrics{\mwin{SSIM $\uparrow$}{0.594}\quad \mwin{LPIPS $\downarrow$}{0.358}\quad
      \mwin{DreamSim $\downarrow$}{0.211}\cwpsnrsep \mlose{PSNR $\uparrow$}{17.63}} \\[\cwaftermet]
    & \filmsprocket{\cwstripw} \\
  \end{tabular}
  \ifdefined\tikzexternalenable\tikzexternalenable\fi
    \caption{\textbf{Higher PSNR, blurrier frames.} Four frames of one HunyuanVideo 480p clip at
    a similar speedup, with green marking the better value on each metric. SeaCache leads on PSNR, yet
    the inset shows its foliage blurred where \method keeps the leaves. SSIM, LPIPS and DreamSim rank
    the clip the other way.}
  \label{fig:psnr_lies_crosswalk}
\end{figure}

PSNR is a per-pixel error, so it rewards an output that drifts toward a smooth conditional mean
and penalises one that keeps sharp detail slightly out of place. At high speedups a blurred output
can therefore outscore a sharper one that looks closer to the reference.
\Cref{fig:psnr_lies_crosswalk} shows such a case. On one HunyuanVideo clip at a similar speedup,
SeaCache leads \method by $3.19$~dB in PSNR, yet its foliage is visibly blurred (inset), and SSIM,
LPIPS and DreamSim all favour \method. We therefore report PSNR in every table but never decide on
it alone, and we back the ablation choices with blind visual comparisons (\cref{app:cap_dial}).

\subsection{Qualitative comparisons across prompts}
\label{app:qualitative_grids}

\Cref{fig:qual_flux_6x} compares all methods against the uncached reference on further DrawBench
prompts at the $\approx\!6.6\times$ FLUX band of \cref{tab:flux_drawbench}. The speedup under each
header is the method's DrawBench-200 mean from \cref{tab:flux_drawbench}. \Cref{tab:prompts_image}
gives the full prompt for every row of \cref{fig:qual_flux_main,fig:qual_flux_6x}.

TaylorSeer degrades on every row at this speed. The other baselines fail in two recurring ways.
Some change the content, adding a second trunk to the elephant (elephant row), drawing two carrots
(carrot row), or blending the giraffe with a different animal (giraffe row). Others keep the
content but lose texture, smoothing the peel (pear row) or covering the tree behind the elephant
with a high-frequency pattern (elephant row). \method keeps the content of the reference and gives
the most realistic textures (pear, red-car, bear and four-cars rows).
\method is the closest method to the reference on all four metrics on every row except the pear,
where SeaCache leads PSNR, SSIM and LPIPS with its smoother peel while \method still leads DreamSim.

\expandafter\def\csname qf6@ours\endcsname{6.91}
\expandafter\def\csname qf6@seacache\endcsname{6.66}
\expandafter\def\csname qf6@dpcache\endcsname{6.50}
\expandafter\def\csname qf6@teacache\endcsname{5.91}
\expandafter\def\csname qf6@taylorseer\endcsname{6.22}

\begin{figure}[p]
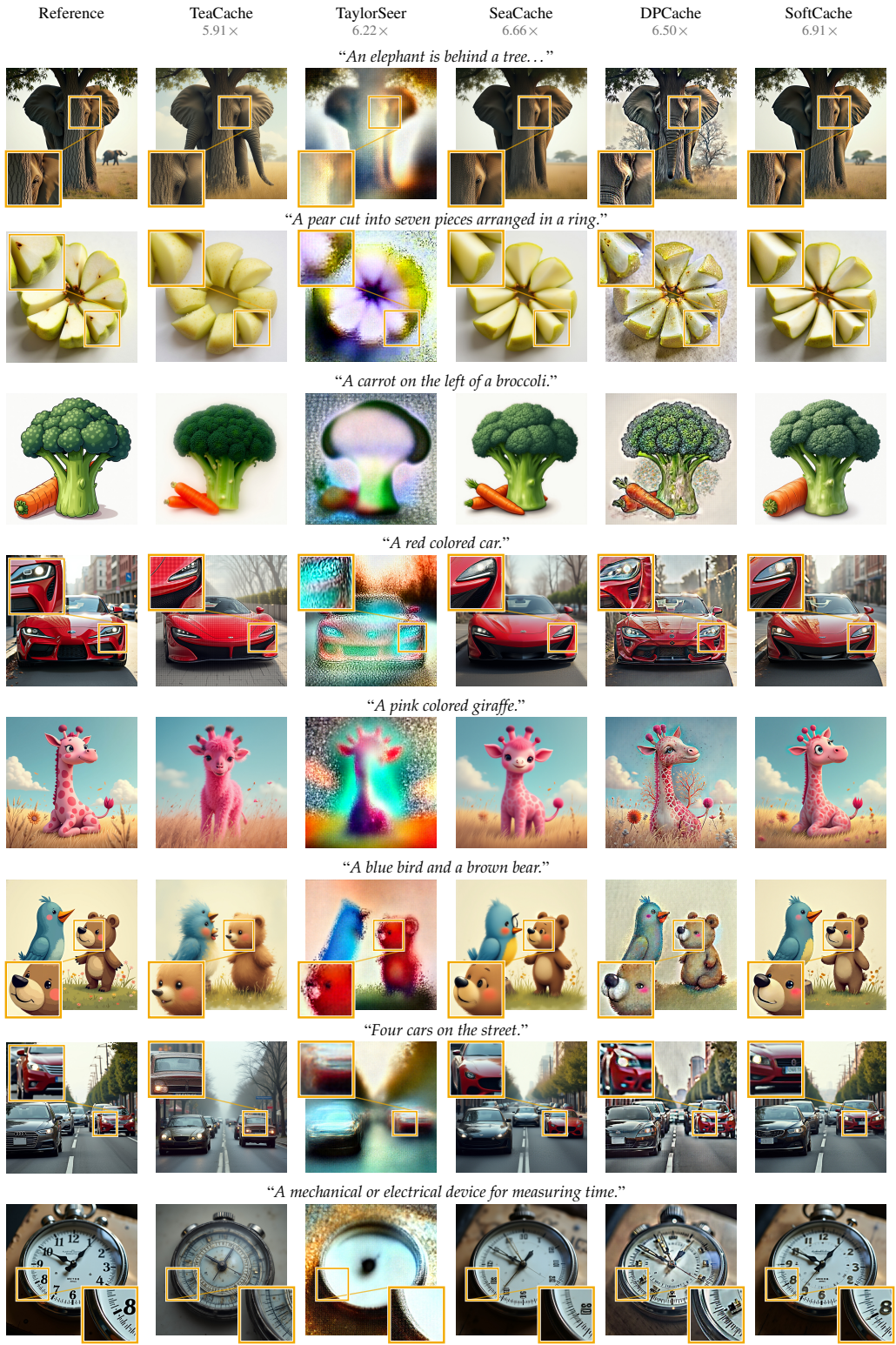

  \centering
  \newcommand{\gpanelw}{0.142}%
  \newcommand{\gpromptfont}{\scriptsize}%
  \newcommand{\gpromptsep}{2pt}%
  \newcommand{\glabsize}{\scriptsize}%
  \newcommand{\gheadline}{0.5pt}%
  \newcommand{\gcolsep}{4pt}%
  \newcommand{\gheadsep}{13pt}%
  \newcommand{\growsep}{5pt}%
  \renewcommand{\zoominsetfrac}{0.42}%
  \setlength{\zoompop}{3pt}%
  \setlength{\tabcolsep}{\gcolsep}%
  \renewcommand{\arraystretch}{0}%
  \newcommand{\gzoomon}[2]{\expandafter\def\csname gzoomon@#1\endcsname{#2}}%
  \newcommand{\gzoomroi}[6]{\expandafter\def\csname gzoomroi@#1\endcsname{%
    \def\ztx{#2}\def\zty{#3}\def\ztw{#4}\def\zth{#5}\def\ztc{#6}}}%
  \newcommand{\gzoomat}[4]{\expandafter\def\csname gzoomat@#1@#2\endcsname{\def\ztx{#3}\def\zty{#4}}}%
  \gzoomon{drawbench_098}{1}%
  \gzoomroi{drawbench_098}{0.4822}{0.2172}{0.2378}{0.2378}{south west}%
  \gzoomon{drawbench_100}{1}%
  \gzoomroi{drawbench_100}{0.6009}{0.6099}{0.2643}{0.2643}{north west}%
  \gzoomon{drawbench_128}{0}%
  \gzoomroi{drawbench_128}{0.05}{0.60}{0.28}{0.28}{south east}%
  \gzoomon{drawbench_000}{1}%
  \gzoomroi{drawbench_000}{0.7078}{0.5244}{0.2118}{0.2118}{north west}%
  \gzoomon{drawbench_012}{0}%
  \gzoomroi{drawbench_012}{0.36}{0.12}{0.22}{0.22}{south east}%
  \gzoomon{drawbench_016}{1}%
  \gzoomroi{drawbench_016}{0.5183}{0.3147}{0.2267}{0.2267}{south west}%
  \gzoomon{drawbench_038}{1}%
  \gzoomroi{drawbench_038}{0.6655}{0.5321}{0.1796}{0.1796}{north west}%
  \gzoomon{drawbench_082}{1}%
  \gzoomroi{drawbench_082}{0.0828}{0.4889}{0.242}{0.242}{south east}%
  \def\gNW{north west}\def\gNE{north east}%
  \newcommand{\gcell}[2]{\def\garm{#2}\def\gref{reference}\def\gzoff{0}%
    \ifcsname gzoomon@#1\endcsname\edef\gzon{\csname gzoomon@#1\endcsname}\else\def\gzon{0}\fi
    \ifx\gzon\gzoff
      \includegraphics[width=\gpanelw\linewidth]{figures/qual_flux_6x/#1/#2.jpg}%
    \else
      \csname gzoomroi@#1\endcsname
      \ifcsname gzoomat@#1@#2\endcsname\csname gzoomat@#1@#2\endcsname\fi
      \setlength{\zoompopy}{\zoompop}%
      \ifx\ztc\gNW\setlength{\zoompopy}{0pt}\fi\ifx\ztc\gNE\setlength{\zoompopy}{0pt}\fi
      \ifx\garm\gref\setlength{\zoompopx}{0pt}\else\setlength{\zoompopx}{\zoompop}\fi
      \zoomtile{\gpanelw\linewidth}{figures/qual_flux_6x/#1/#2.jpg}{\ztx}{\zty}{\ztw}{\zth}{\ztc}%
    \fi}%
  \newcommand{\gprompt}[1]{\multicolumn{6}{@{}c@{}}{\parbox{\dimexpr6\dimexpr\gpanelw\linewidth\relax+10\dimexpr\gcolsep\relax\relax}{\centering\gpromptfont\gridprompt{#1}}} \\[\gpromptsep]}%
  \newcommand{\ghbox}[1]{\hbox to \gpanelw\linewidth{\hss#1\hss}}%
  \newcommand{\glabref}[1]{\vtop{\ghbox{\glabsize #1}}}%
  \newcommand{\glab}[2]{\vtop{\ghbox{\glabsize\strut #1}\kern\gheadline\ghbox{\tiny\color{black!55}\csname qf6@#2\endcsname$\times$}}}%
  \newcommand{\grow}[1]{%
    \gcell{#1}{reference} & \gcell{#1}{teacache} & \gcell{#1}{taylorseer} &
    \gcell{#1}{seacache} & \gcell{#1}{dpcache} & \gcell{#1}{ours}}%
  \ifdefined\tikzexternaldisable\tikzexternaldisable\fi
  \begin{tabular}{@{}cccccc@{}}
    \glabref{Reference} & \glab{TeaCache}{teacache} & \glab{TaylorSeer}{taylorseer} &
    \glab{SeaCache}{seacache} & \glab{DPCache}{dpcache} & \glab{\method}{ours} \\[\gheadsep]
    \gprompt{An elephant is behind a tree\dots}
    \grow{drawbench_098} \\[\growsep]
    \gprompt{A pear cut into seven pieces arranged in a ring.}
    \grow{drawbench_100} \\[\growsep]
    \gprompt{A carrot on the left of a broccoli.}
    \grow{drawbench_128} \\[\growsep]
    \gprompt{A red colored car.}
    \grow{drawbench_000} \\[\growsep]
    \gprompt{A pink colored giraffe.}
    \grow{drawbench_012} \\[\growsep]
    \gprompt{A blue bird and a brown bear.}
    \grow{drawbench_016} \\[\growsep]
    \gprompt{Four cars on the street.}
    \grow{drawbench_038} \\[\growsep]
    \gprompt{A mechanical or electrical device for measuring time.}
    \grow{drawbench_082} \\
  \end{tabular}
  \ifdefined\tikzexternalenable\tikzexternalenable\fi
  \caption{\textbf{FLUX at the $\approx\!6.6\times$ band, eight DrawBench prompts.} Each method's achieved speedup over DrawBench-200 is under its header (\cref{tab:flux_drawbench}).}
  \label{fig:qual_flux_6x}
\end{figure}

\Cref{fig:supp_zoom_flux} returns to the slower $\approx\!5.2\times$ band of
\cref{fig:qual_flux_main} on two prompts. \method is the closest method to the reference, and
the enlarged crops show this most clearly on the frame of the clock, which \method reproduces
more faithfully than any baseline. On the second prompt, of two cats and two dogs, the reference
itself draws only one dog, so what counts is the animals it does draw. Some baselines change them,
morphing one of the cats, while \method keeps the two cats and the dog of
the reference.

\expandafter\def\csname sgf@6.6x@ours\endcsname{6.91}
\expandafter\def\csname sgf@6.6x@seacache\endcsname{6.66}
\expandafter\def\csname sgf@6.6x@dpcache\endcsname{6.50}
\expandafter\def\csname sgf@6.6x@teacache\endcsname{5.91}
\expandafter\def\csname sgf@6.6x@taylorseer\endcsname{6.22}
\expandafter\def\csname sgf@5.2x@ours\endcsname{5.44}
\expandafter\def\csname sgf@5.2x@seacache\endcsname{5.28}
\expandafter\def\csname sgf@5.2x@dpcache\endcsname{4.74}
\expandafter\def\csname sgf@5.2x@teacache\endcsname{5.28}
\expandafter\def\csname sgf@5.2x@taylorseer\endcsname{5.59}

\begin{figure}[t]
  \centering
  \newcommand{\zpanelw}{0.160}%
  \newcommand{\zlabsize}{\scriptsize}%
  \newcommand{\zpromptfont}{\scriptsize}%
  \newcommand{\zheadline}{0.5pt}%
  \newcommand{\zcolsep}{1.2pt}%
  \newcommand{\zheadsep}{10pt}%
  \newcommand{\zpromptsep}{2pt}%
  \newcommand{\zrowsep}{1.2pt}%
  \newcommand{\zexsep}{6pt}%
  \setlength{\tabcolsep}{\zcolsep}%
  \renewcommand{\arraystretch}{0}%
  \newcommand{\zfull}[2]{\input{figures/supp_zoom_flux/#1/crop}\unskip\begin{tikzpicture}%
    \node[anchor=south west,inner sep=0] (img) at (0,0)%
      {\includegraphics[width=\zpanelw\linewidth]{figures/supp_zoom_flux/#1/#2.jpg}};%
    \begin{scope}[x={(img.south east)},y={(img.north west)}]%
      \draw[red,line width=.6pt] (\zboxx,{1-\zboxy}) rectangle ({\zboxx+\zboxs},{1-\zboxy-\zboxs});%
    \end{scope}\end{tikzpicture}}%
  \newcommand{\zcrop}[2]{\includegraphics[width=\zpanelw\linewidth]%
    {figures/supp_zoom_flux/#1/#2_crop.jpg}}%
  \newcommand{\zrows}[1]{%
    \input{figures/supp_zoom_flux/#1/crop}\unskip\begin{tikzpicture}%
    \node[anchor=south west,inner sep=0] (img) at (0,0)%
      {\includegraphics[width=\zpanelw\linewidth]{figures/supp_zoom_flux/#1/reference.jpg}};%
    \begin{scope}[x={(img.south east)},y={(img.north west)}]%
      \draw[red,line width=.6pt] (\zboxx,{1-\zboxy}) rectangle ({\zboxx+\zboxs},{1-\zboxy-\zboxs});%
    \end{scope}\end{tikzpicture} & \input{figures/supp_zoom_flux/#1/crop}\unskip\begin{tikzpicture}%
    \node[anchor=south west,inner sep=0] (img) at (0,0)%
      {\includegraphics[width=\zpanelw\linewidth]{figures/supp_zoom_flux/#1/teacache.jpg}};%
    \begin{scope}[x={(img.south east)},y={(img.north west)}]%
      \draw[red,line width=.6pt] (\zboxx,{1-\zboxy}) rectangle ({\zboxx+\zboxs},{1-\zboxy-\zboxs});%
    \end{scope}\end{tikzpicture} & \input{figures/supp_zoom_flux/#1/crop}\unskip\begin{tikzpicture}%
    \node[anchor=south west,inner sep=0] (img) at (0,0)%
      {\includegraphics[width=\zpanelw\linewidth]{figures/supp_zoom_flux/#1/taylorseer.jpg}};%
    \begin{scope}[x={(img.south east)},y={(img.north west)}]%
      \draw[red,line width=.6pt] (\zboxx,{1-\zboxy}) rectangle ({\zboxx+\zboxs},{1-\zboxy-\zboxs});%
    \end{scope}\end{tikzpicture} &
    \input{figures/supp_zoom_flux/#1/crop}\unskip\begin{tikzpicture}%
    \node[anchor=south west,inner sep=0] (img) at (0,0)%
      {\includegraphics[width=\zpanelw\linewidth]{figures/supp_zoom_flux/#1/seacache.jpg}};%
    \begin{scope}[x={(img.south east)},y={(img.north west)}]%
      \draw[red,line width=.6pt] (\zboxx,{1-\zboxy}) rectangle ({\zboxx+\zboxs},{1-\zboxy-\zboxs});%
    \end{scope}\end{tikzpicture} & \input{figures/supp_zoom_flux/#1/crop}\unskip\begin{tikzpicture}%
    \node[anchor=south west,inner sep=0] (img) at (0,0)%
      {\includegraphics[width=\zpanelw\linewidth]{figures/supp_zoom_flux/#1/dpcache.jpg}};%
    \begin{scope}[x={(img.south east)},y={(img.north west)}]%
      \draw[red,line width=.6pt] (\zboxx,{1-\zboxy}) rectangle ({\zboxx+\zboxs},{1-\zboxy-\zboxs});%
    \end{scope}\end{tikzpicture} & \input{figures/supp_zoom_flux/#1/crop}\unskip\begin{tikzpicture}%
    \node[anchor=south west,inner sep=0] (img) at (0,0)%
      {\includegraphics[width=\zpanelw\linewidth]{figures/supp_zoom_flux/#1/ours.jpg}};%
    \begin{scope}[x={(img.south east)},y={(img.north west)}]%
      \draw[red,line width=.6pt] (\zboxx,{1-\zboxy}) rectangle ({\zboxx+\zboxs},{1-\zboxy-\zboxs});%
    \end{scope}\end{tikzpicture} \\[\zrowsep]
    \zcrop{#1}{reference} & \zcrop{#1}{teacache} & \zcrop{#1}{taylorseer} &
    \zcrop{#1}{seacache} & \zcrop{#1}{dpcache} & \zcrop{#1}{ours}}%
  \newcommand{\zprompt}[1]{\multicolumn{6}{@{}c@{}}{\parbox{\dimexpr6\dimexpr\zpanelw\linewidth\relax+10\dimexpr\zcolsep\relax\relax}{\centering\zpromptfont\gridprompt{#1}}} \\[\zpromptsep]}%
  \newcommand{\zhbox}[1]{\hbox to \zpanelw\linewidth{\hss#1\hss}}%
  \newcommand{\zlabref}[1]{\vtop{\zhbox{\zlabsize #1}}}%
  \newcommand{\zlab}[2]{\vtop{\zhbox{\zlabsize\strut #1}\kern\zheadline\zhbox{\tiny\color{black!55}\csname sgf@5.2x@#2\endcsname$\times$}}}%
  \ifdefined\tikzexternaldisable\tikzexternaldisable\fi
  \begin{tabular}{@{}cccccc@{}}
    \zlabref{Reference} & \zlab{TeaCache}{teacache} & \zlab{TaylorSeer}{taylorseer} &
    \zlab{SeaCache}{seacache} & \zlab{DPCache}{dpcache} & \zlab{\method}{ours} \\[\zheadsep]
    \zprompt{A single clock is sitting on a table.}
    \zrows{clock} \\[\zexsep]
    \zprompt{Two cats and two dogs sitting on the grass.}
    \zrows{drawbench_049} \\
  \end{tabular}
  \ifdefined\tikzexternalenable\tikzexternalenable\fi
  \caption{\textbf{FLUX on DrawBench at $\approx\!5.2\times$ speedup, two prompts enlarged.} The row
  under each prompt enlarges the region in its red box, the frame of the clock and the middle cat,
  where \method stays closest to the reference.}
  \label{fig:supp_zoom_flux}
\end{figure}
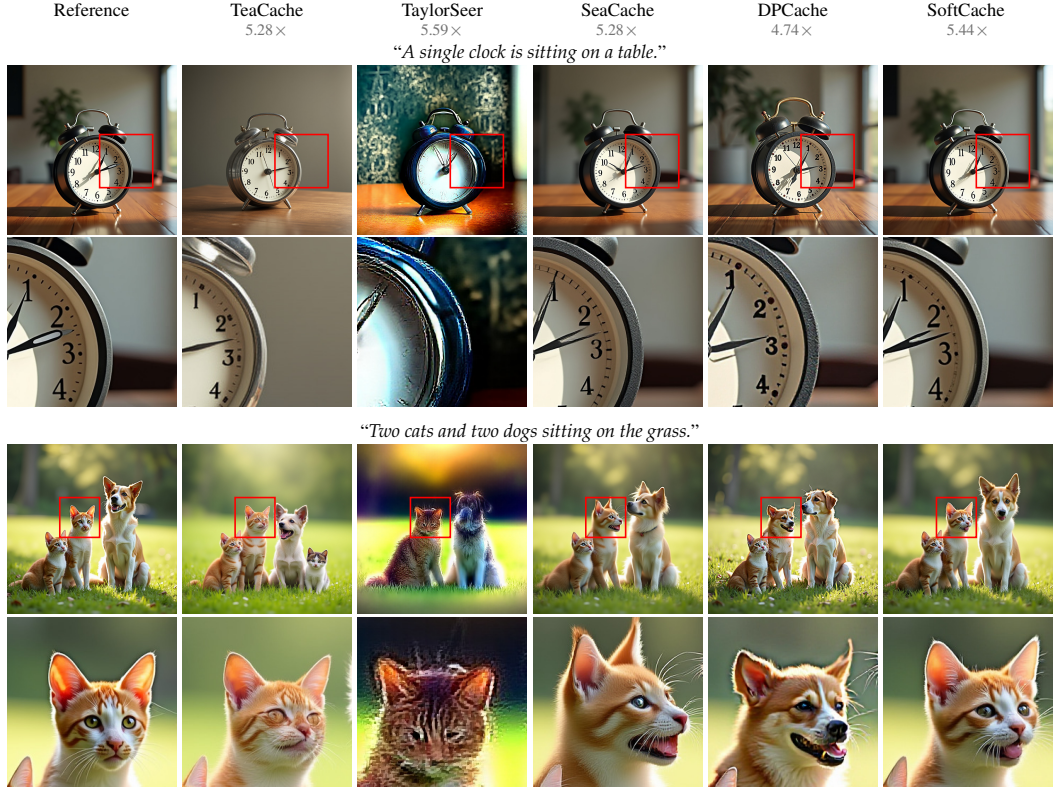

The video figures show three frames per clip, with zoom insets and red ellipses marking defects.
\Cref{fig:supp_grid_hunyuan_5x} adds two HunyuanVideo clips at the $\approx\!5.4\times$ band of
\cref{fig:grid_hunyuan_main}. On the sweeping clip SeaCache blurs the whole scene and DPCache
changes the content, drawing a second broom (ellipse), while \method stays sharp and keeps a
single broom. On the zebra clip the inset shows DPCache smearing the back of the zebra's head
and SeaCache blurring the background far more than the other methods, while \method keeps both
close to the reference.

\expandafter\def\csname gh@seacache\endcsname{5.33}
\expandafter\def\csname gh@dpcache\endcsname{5.42}
\expandafter\def\csname gh@ours\endcsname{5.39}

\begin{figure}[ht]
  \centering
  \newlength{\shpanelw}\setlength{\shpanelw}{0.15\linewidth}%
  \newlength{\shframesep}\setlength{\shframesep}{4pt}%
  \newcommand{\hlabsize}{\scriptsize}%
  \newcommand{\hlabline}{0.5pt}%
  \newcommand{\hlabsep}{3pt}%
  \newcommand{\hclipsep}{4pt}%
  \newcommand{\hrowsep}{4pt}%
  \renewcommand{\filmstriph}{4.5}%
  \renewcommand{\filmholew}{3.2pt}%
  \renewcommand{\filmholeh}{2.2pt}%
  \renewcommand{\filmnholes}{8}%
  \renewcommand{\filmstripcol}{black!85}%
  \renewcommand{\zoominsetfrac}{0.42}%
  \setlength{\zoompop}{2.8pt}%
  \setlength{\zoompopx}{\zoompop}\setlength{\zoompopy}{\zoompop}%
  \newlength{\shtripw}\setlength{\shtripw}{3\shpanelw}\addtolength{\shtripw}{2\shframesep}%
  \setlength{\tabcolsep}{0pt}%
  \renewcommand{\arraystretch}{0}%
  \def\hroiA{vbad168}%
  \newcommand{\hroi}[1]{%
    \def\hroiid{#1}%
    \ifx\hroiid\hroiA %
      \def\ztf{-1}%
    \else %
      \def\ztf{0}\def\ztx{0.32}\def\zty{0.18}\def\ztw{0.18}\def\zth{0.24}\def\ztc{north east}%
    \fi}%
  \newcommand{\hmark}[7]{\expandafter\def\csname hmark@#1@#2@#3\endcsname{\defectellipse{#4}{#5}{#6}{#7}}}%
  \hmark{vbad168}{dpcache}{1}{0.775}{0.61}{0.17}{0.11}%
  \newcommand{\hcell}[3]{\hroi{#1}%
    \ifcsname hmark@#1@#2@#3\endcsname\expandafter\let\expandafter\zoommarks\csname hmark@#1@#2@#3\endcsname\else\def\zoommarks{}\fi
    \ifnum#3=\ztf
      \zoomtile{\shpanelw}{figures/supp_grid_hunyuan_5x/#1/#2_f#3.jpg}{\ztx}{\zty}{\ztw}{\zth}{\ztc}%
    \else\marktile{\shpanelw}{figures/supp_grid_hunyuan_5x/#1/#2_f#3.jpg}\fi}%
  \newcommand{\htrip}[2]{\hcell{#1}{#2}{0}\hspace{\shframesep}%
    \hcell{#1}{#2}{1}\hspace{\shframesep}\hcell{#1}{#2}{2}}%
  \newlength{\shrowh}%
  \settoheight{\shrowh}{\includegraphics[width=\shpanelw]%
    {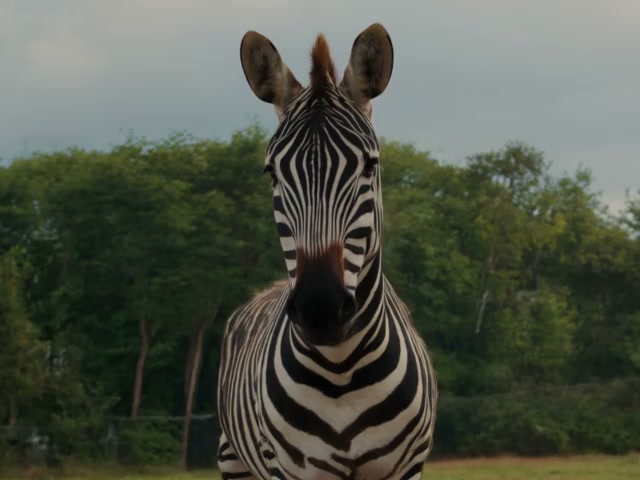}}%
  \newcommand{\hrot}[1]{\raisebox{\dimexpr0.5\shrowh-0.5\height\relax}%
    {\rotatebox{90}{{\hlabsize #1}}}}%
  \newcommand{\hlab}[2]{\hrot{\shortstack[c]{#1\\[\hlabline](#2)}}}%
  \newcommand{\hlabref}[1]{\hrot{\shortstack[c]{#1\\[\hlabline]\strut}}}%
  \newcommand{\hpromptfont}{\footnotesize}%
  \newcommand{\hprompt}[1]{\parbox{\shtripw}{\centering\hpromptfont\gridprompt{#1}}}%
  \newcommand{\hfilmbar}{ & \filmsprocket{\shtripw} & \filmsprocket{\shtripw}}%
  \ifdefined\tikzexternaldisable\tikzexternaldisable\fi
  \begin{tabular}{@{}c@{\hspace{\hlabsep}}c@{\hspace{\hclipsep}}c@{}}
    & \hprompt{A person is sweeping floor} & \hprompt{a zebra} \\[\hrowsep]
    \hfilmbar \\[\hrowsep]
    \hlabref{Reference} & \htrip{vbad168}{reference} & \htrip{vbad351}{reference} \\[\hrowsep]
    \hlab{SeaCache}{$\csname gh@seacache\endcsname\times$} & \htrip{vbad168}{seacache} & \htrip{vbad351}{seacache} \\[\hrowsep]
    \hlab{DPCache}{$\csname gh@dpcache\endcsname\times$} & \htrip{vbad168}{dpcache} & \htrip{vbad351}{dpcache} \\[\hrowsep]
    \hlab{\method}{$\csname gh@ours\endcsname\times$} & \htrip{vbad168}{ours} & \htrip{vbad351}{ours} \\[\hrowsep]
    \hfilmbar \\
  \end{tabular}
  \ifdefined\tikzexternalenable\tikzexternalenable\fi
  \caption{\textbf{HunyuanVideo on VBench at $\approx\!5.4\times$ speedup, two further clips.}
  Same band, methods and frames as \cref{fig:grid_hunyuan_main}. The red ellipse marks a second
  broom in DPCache, and the inset enlarges the back of the zebra's head.}
  \label{fig:supp_grid_hunyuan_5x}
\end{figure}

\Cref{fig:supp_grid_hunyuan} repeats the comparison at the slower $\approx\!3.5\times$ band. On
the harp clip the inset shows that \method keeps the most detail in the hand, and in the last
frame DPCache hallucinates a second hand next to the one on the strings (ellipse), so the player
has three hands. On the grooming clip \method keeps the dog closest to the reference. SeaCache
distorts the dog's face, and DPCache changes the action and leaves the face less visible.

\expandafter\def\csname ghs@seacache\endcsname{3.53}
\expandafter\def\csname ghs@dpcache\endcsname{3.28}
\expandafter\def\csname ghs@ours\endcsname{3.52}

\begin{figure}[ht]
  \centering
  \newlength{\hspanelw}\setlength{\hspanelw}{0.15\linewidth}%
  \newlength{\hsframesep}\setlength{\hsframesep}{4pt}%
  \newcommand{\hslabsize}{\scriptsize}%
  \newcommand{\hslabline}{0.5pt}%
  \newcommand{\hslabsep}{3pt}%
  \newcommand{\hsclipsep}{4pt}%
  \newcommand{\hsrowsep}{4pt}%
  \renewcommand{\filmstriph}{4.5}%
  \renewcommand{\filmholew}{3.2pt}%
  \renewcommand{\filmholeh}{2.2pt}%
  \renewcommand{\filmnholes}{8}%
  \renewcommand{\filmstripcol}{black!85}%
  \renewcommand{\zoominsetfrac}{0.42}%
  \setlength{\zoompop}{2.8pt}%
  \setlength{\zoompopx}{\zoompop}\setlength{\zoompopy}{\zoompop}%
  \newlength{\hstripw}\setlength{\hstripw}{3\hspanelw}\addtolength{\hstripw}{2\hsframesep}%
  \setlength{\tabcolsep}{0pt}%
  \renewcommand{\arraystretch}{0}%
  \def\hsroiA{vbad165}%
  \newcommand{\hsroi}[1]{%
    \def\hsroiid{#1}%
    \ifx\hsroiid\hsroiA %
      \def\ztf{0}\def\ztx{0.74}\def\zty{0.33}\def\ztw{0.18}\def\zth{0.24}\def\ztc{north west}%
    \else %
      \def\ztf{0}\def\ztx{0.62}\def\zty{0.46}\def\ztw{0.18}\def\zth{0.24}\def\ztc{south west}%
    \fi}%
  \newcommand{\hsmark}[7]{\expandafter\def\csname hsmark@#1@#2@#3\endcsname{\defectellipse{#4}{#5}{#6}{#7}}}%
  \hsmark{vbad165}{dpcache}{2}{0.64}{0.69}{0.17}{0.22}%
  \newcommand{\hscell}[3]{\hsroi{#1}%
    \ifcsname hsmark@#1@#2@#3\endcsname\expandafter\let\expandafter\zoommarks\csname hsmark@#1@#2@#3\endcsname\else\def\zoommarks{}\fi
    \ifnum#3=\ztf
      \zoomtile{\hspanelw}{figures/supp_grid_hunyuan/#1/#2_f#3.jpg}{\ztx}{\zty}{\ztw}{\zth}{\ztc}%
    \else\marktile{\hspanelw}{figures/supp_grid_hunyuan/#1/#2_f#3.jpg}\fi}%
  \newcommand{\hstrip}[2]{\hscell{#1}{#2}{0}\hspace{\hsframesep}%
    \hscell{#1}{#2}{1}\hspace{\hsframesep}\hscell{#1}{#2}{2}}%
  \newlength{\hsrowh}%
  \settoheight{\hsrowh}{\includegraphics[width=\hspanelw]%
    {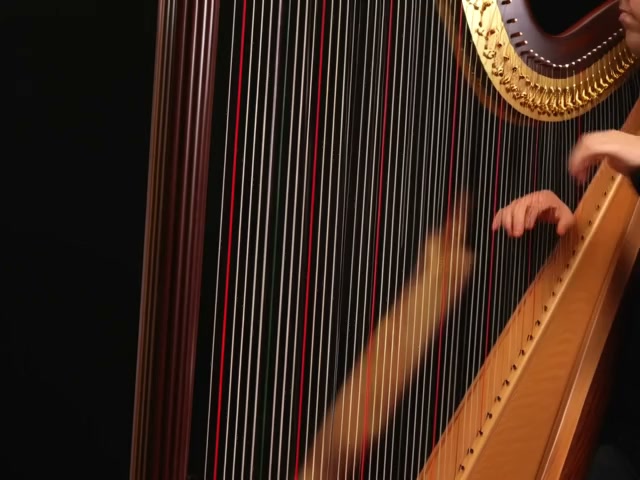}}%
  \newcommand{\hsrot}[1]{\raisebox{\dimexpr0.5\hsrowh-0.5\height\relax}%
    {\rotatebox{90}{{\hslabsize #1}}}}%
  \newcommand{\hslab}[2]{\hsrot{\shortstack[c]{#1\\[\hslabline](#2)}}}%
  \newcommand{\hslabref}[1]{\hsrot{\shortstack[c]{#1\\[\hslabline]\strut}}}%
  \newcommand{\hspromptfont}{\footnotesize}%
  \newcommand{\hsprompt}[1]{\parbox{\hstripw}{\centering\hspromptfont\gridprompt{#1}}}%
  \newcommand{\hsfilmbar}{ & \filmsprocket{\hstripw} & \filmsprocket{\hstripw}}%
  \ifdefined\tikzexternaldisable\tikzexternaldisable\fi
  \begin{tabular}{@{}c@{\hspace{\hslabsep}}c@{\hspace{\hsclipsep}}c@{}}
    & \hsprompt{A person is playing harp} & \hsprompt{A person is grooming dog} \\[\hsrowsep]
    \hsfilmbar \\[\hsrowsep]
    \hslabref{Reference} & \hstrip{vbad165}{reference} & \hstrip{vbad220}{reference} \\[\hsrowsep]
    \hslab{SeaCache}{$\csname ghs@seacache\endcsname\times$} & \hstrip{vbad165}{seacache} & \hstrip{vbad220}{seacache} \\[\hsrowsep]
    \hslab{DPCache}{$\csname ghs@dpcache\endcsname\times$} & \hstrip{vbad165}{dpcache} & \hstrip{vbad220}{dpcache} \\[\hsrowsep]
    \hslab{\method}{$\csname ghs@ours\endcsname\times$} & \hstrip{vbad165}{ours} & \hstrip{vbad220}{ours} \\[\hsrowsep]
    \hsfilmbar \\
  \end{tabular}
  \ifdefined\tikzexternalenable\tikzexternalenable\fi
  \caption{\textbf{HunyuanVideo on VBench at $\approx\!3.5\times$ speedup.} Frames 1, 33 and 65 of
  65, with a zoom inset of the boxed region on the first frame. The red ellipse marks a
  hallucinated second hand in DPCache.}
  \label{fig:supp_grid_hunyuan}
\end{figure}

\Cref{fig:grid_wan} repeats the comparison on Wan2.1 at $\approx\!3.7\times$. Across the
shown frames, SeaCache and DPCache lose fine detail. Edges soften, high-frequency texture
fades, and small structures that are sharp in the reference become smeared. \method
retains this detail and stays visually closest to the reference throughout. The inset on
the palace pool shows the largest gap, where both baselines distort the pool far more than
\method.

\expandafter\def\csname gw@seacache\endcsname{3.73}
\expandafter\def\csname gw@dpcache\endcsname{3.60}
\expandafter\def\csname gw@ours\endcsname{3.83}

\begin{figure}[t]
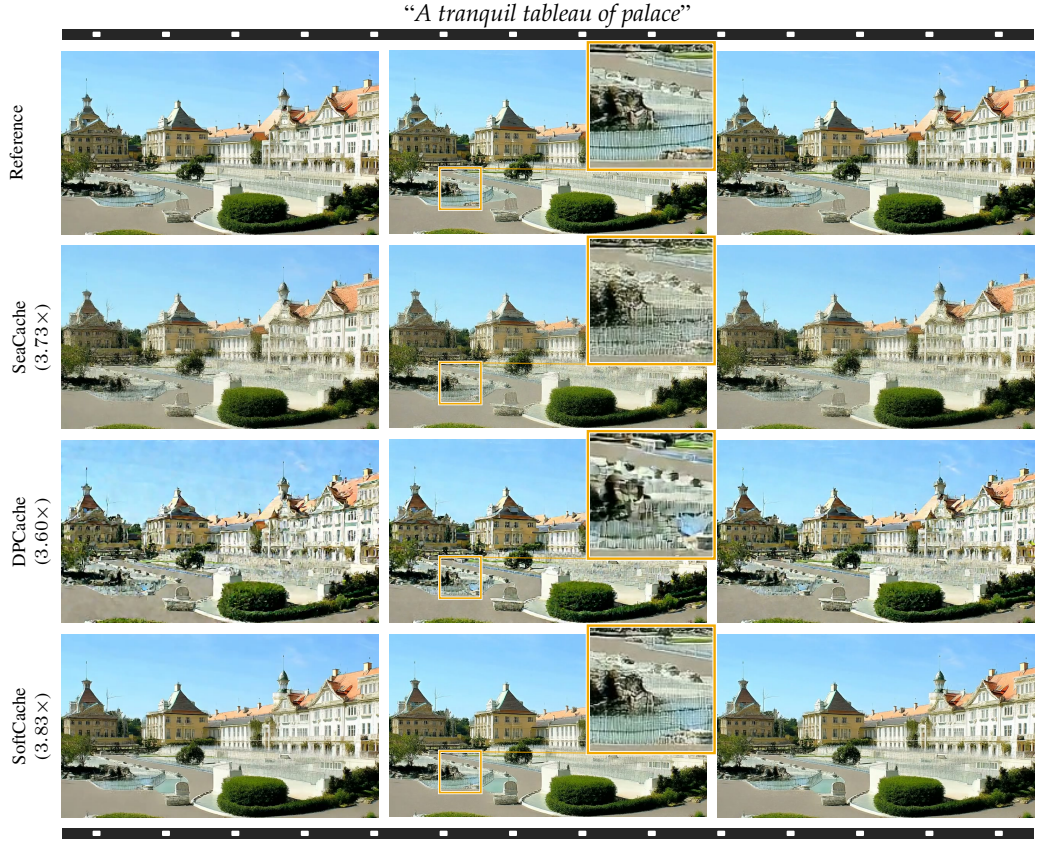

  \centering
  \newlength{\wpanelw}\setlength{\wpanelw}{0.3\textwidth}%
  \newlength{\wframesep}\setlength{\wframesep}{4pt}%
  \newcommand{\wlabsize}{\scriptsize}%
  \newcommand{\wpromptfont}{\footnotesize}%
  \newcommand{\wlabline}{0.5pt}%
  \newcommand{\wlabsep}{3pt}%
  \newcommand{\wrowsep}{4pt}%
  \newcommand{\wheadsep}{2pt}%
  \renewcommand{\filmstriph}{4}%
  \renewcommand{\filmholew}{3pt}%
  \renewcommand{\filmholeh}{2pt}%
  \renewcommand{\filmnholes}{14}%
  \renewcommand{\filmstripcol}{black!85}%
  \renewcommand{\zoominsetfrac}{0.40}%
  \setlength{\zoompop}{3pt}%
  \setlength{\zoompopx}{\zoompop}\setlength{\zoompopy}{\zoompop}%
  \newlength{\wtripw}\setlength{\wtripw}{3\wpanelw}\addtolength{\wtripw}{2\wframesep}%
  \setlength{\tabcolsep}{0pt}%
  \renewcommand{\arraystretch}{0}%
  \newcommand{\wroi}[1]{%
    \def\ztx{0.1600}\def\zty{0.6400}\def\ztw{0.1298}\def\zth{0.2250}\def\ztc{north east}}%
  \newcommand{\wcell}[3]{\ifnum#3=1 \wroi{#1}%
    \zoomtile{\wpanelw}{figures/grid_wan/#1/#2_f#3.jpg}{\ztx}{\zty}{\ztw}{\zth}{\ztc}\else
    \includegraphics[width=\wpanelw]{figures/grid_wan/#1/#2_f#3.jpg}\fi}%
  \newcommand{\wtrip}[2]{\wcell{#1}{#2}{0}\hspace{\wframesep}%
    \wcell{#1}{#2}{1}\hspace{\wframesep}\wcell{#1}{#2}{2}}%
  \newlength{\wrowh}%
  \settoheight{\wrowh}{\wcell{vbad015}{reference}{0}}%
  \newcommand{\wrot}[1]{\raisebox{\dimexpr0.5\wrowh-0.5\height\relax}%
    {\rotatebox{90}{{\wlabsize #1}}}}%
  \newcommand{\wlab}[2]{\wrot{\shortstack[c]{#1\\[\wlabline](${#2}\times$)}}}%
  \newcommand{\wlabref}[1]{\wrot{\shortstack[c]{#1\\[\wlabline]\strut}}}%
  \newcommand{\wfilmbar}{ & \filmsprocket{\wtripw}}%
  \newcommand{\wside}[2]{\begin{tabular}[t]{@{}c@{\hspace{\wlabsep}}c@{}}
    & {\wpromptfont\gridprompt{#2}} \\[\wheadsep]
    \wfilmbar \\[\wrowsep]
    \wlabref{Reference} & \wtrip{#1}{reference} \\[\wrowsep]
    \wlab{SeaCache}{\csname gw@seacache\endcsname} & \wtrip{#1}{seacache} \\[\wrowsep]
    \wlab{DPCache}{\csname gw@dpcache\endcsname} & \wtrip{#1}{dpcache} \\[\wrowsep]
    \wlab{\method}{\csname gw@ours\endcsname} & \wtrip{#1}{ours} \\[\wrowsep]
    \wfilmbar \\
  \end{tabular}}%
  \ifdefined\tikzexternaldisable\tikzexternaldisable\fi
  \wside{vbad015}{A tranquil tableau of palace}
  \ifdefined\tikzexternalenable\tikzexternalenable\fi
  \caption{\textbf{Wan2.1 on VBench at $\approx\!3.7\times$ speedup.} Frames 1, 41 and 81 of 81,
  with each method's mean speedup at left (\cref{tab:wan_vbench}) and a zoom inset of the pool on
  the middle frame. SeaCache and DPCache distort the pool, while \method stays closest to the
  reference.}
  \label{fig:grid_wan}
\end{figure}

\Cref{fig:supp_grid_hunyuan720} runs HunyuanVideo at its native $1280\times720$, 129-frame
resolution, instead of the $640\times480$, 65-frame setting of the tables, against SeaCache at
$\approx\!4.5\times$ (median speedup over the 720p prompts). \method does not reproduce the
reference's motion exactly but stays consistent with it, while SeaCache has the corgi start a
distorted flip in the middle frame that never happens in the reference. By the last frame SeaCache
also softens the corgi's face and the grass, while \method keeps both sharp.

\expandafter\def\csname ghk@seacache\endcsname{4.53}
\expandafter\def\csname ghk@ours\endcsname{4.53}

\begin{figure}[ht]
  \centering
  \newlength{\hkpanelw}\setlength{\hkpanelw}{0.3\linewidth}%
  \newlength{\hkframesep}\setlength{\hkframesep}{4pt}%
  \newcommand{\hklabsize}{\footnotesize}%
  \newcommand{\hklabline}{0.5pt}%
  \newcommand{\hklabsep}{3pt}%
  \newcommand{\hkrowsep}{4pt}%
  \renewcommand{\filmstriph}{4.5}%
  \renewcommand{\filmholew}{3.2pt}%
  \renewcommand{\filmholeh}{2.2pt}%
  \renewcommand{\filmnholes}{32}%
  \renewcommand{\filmstripcol}{black!85}%
  \renewcommand{\zoominsetfrac}{0.42}%
  \setlength{\zoompop}{2.8pt}%
  \setlength{\zoompopx}{0pt}\setlength{\zoompopy}{\zoompop}%
  \newlength{\hkstripw}\setlength{\hkstripw}{3\hkpanelw}\addtolength{\hkstripw}{2\hkframesep}%
  \setlength{\tabcolsep}{0pt}%
  \renewcommand{\arraystretch}{0}%
  \def\ztx{0.40}\def\zty{0.26}\def\ztw{0.16}\def\zth{0.284}\def\ztc{north east}%
  \def\zoommarks{}%
  \newcommand{\hkcell}[2]{\ifnum#2=2
      \zoomtile{\hkpanelw}{figures/supp_grid_hunyuan720/vbad623/#1_f#2.jpg}{\ztx}{\zty}{\ztw}{\zth}{\ztc}%
    \else\marktile{\hkpanelw}{figures/supp_grid_hunyuan720/vbad623/#1_f#2.jpg}\fi}%
  \newcommand{\hkstrip}[1]{\hkcell{#1}{0}\hspace{\hkframesep}%
    \hkcell{#1}{1}\hspace{\hkframesep}\hkcell{#1}{2}}%
  \newlength{\hkrowh}%
  \settoheight{\hkrowh}{\includegraphics[width=\hkpanelw]%
    {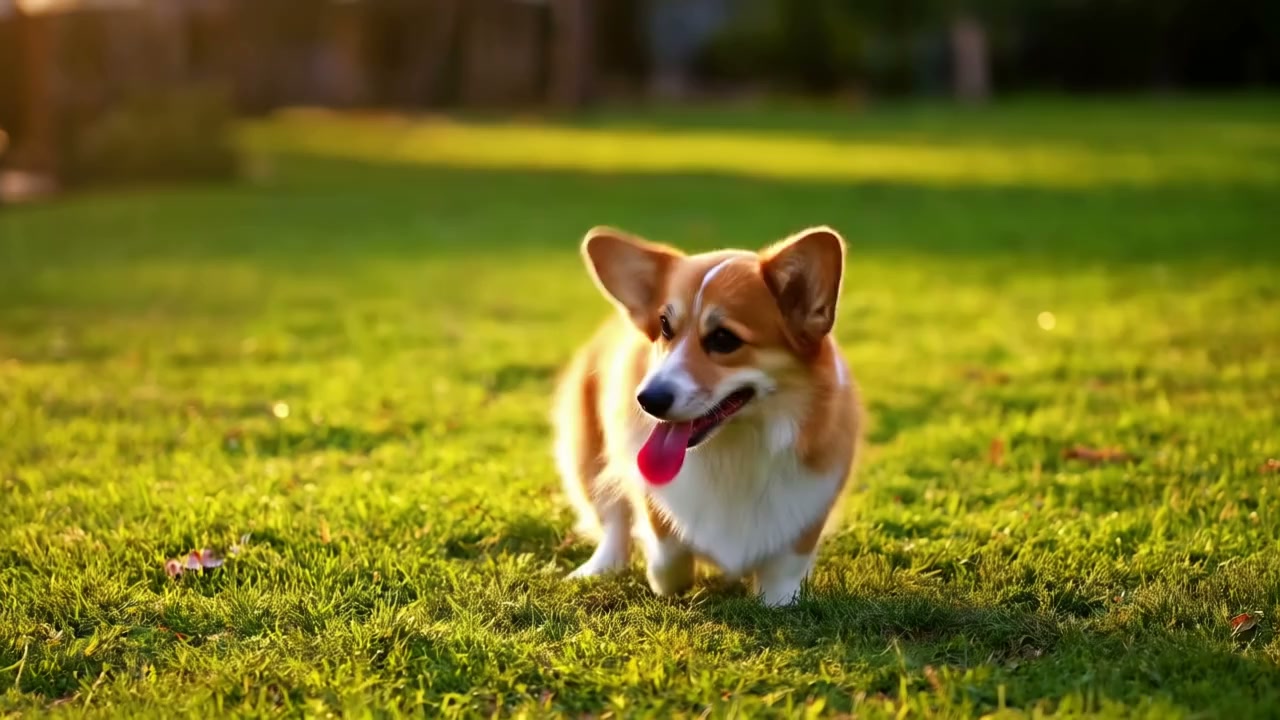}}%
  \newcommand{\hkrot}[1]{\raisebox{\dimexpr0.5\hkrowh-0.5\height\relax}%
    {\rotatebox{90}{{\hklabsize #1}}}}%
  \newcommand{\hklab}[2]{\hkrot{\shortstack[c]{#1\\[\hklabline](#2)}}}%
  \newcommand{\hklabref}[1]{\hkrot{\shortstack[c]{#1\\[\hklabline]\strut}}}%
  \newcommand{\hkprompt}[1]{\parbox{\hkstripw}{\centering\footnotesize\gridprompt{#1}}}%
  \ifdefined\tikzexternaldisable\tikzexternaldisable\fi
  \begin{tabular}{@{}c@{\hspace{\hklabsep}}c@{}}
    & \hkprompt{A cute happy Corgi playing in park, sunset, in super slow motion} \\[\hkrowsep]
    & \filmsprocket{\hkstripw} \\[\hkrowsep]
    \hklabref{Reference} & \hkstrip{reference} \\[\hkrowsep]
    \hklab{SeaCache}{$\csname ghk@seacache\endcsname\times$} & \hkstrip{seacache} \\[\hkrowsep]
    \hklab{\method}{$\csname ghk@ours\endcsname\times$} & \hkstrip{ours} \\[\hkrowsep]
    & \filmsprocket{\hkstripw} \\
  \end{tabular}
  \ifdefined\tikzexternalenable\tikzexternalenable\fi
  \caption{\textbf{HunyuanVideo at native 720p, $\approx\!4.5\times$ speedup.} Frames 1, 65 and 129
  of 129, the last frame carrying a zoom inset of the corgi's face.}
  \label{fig:supp_grid_hunyuan720}
\end{figure}

\begin{table}[ht]
\centering
\small
\setlength{\tabcolsep}{5pt}
\renewcommand{\arraystretch}{1.15}
\caption{\textbf{Prompts of the image qualitative figures, verbatim.}}
\label{tab:prompts_image}
\begin{tabular}{llp{0.55\linewidth}}
\toprule
  Prompt id & Category & Text \\
\midrule
  \multicolumn{3}{l}{\textit{\cref{fig:qual_flux_main}}} \\
  \texttt{drawbench\_133} & Positional & A zebra to the right of a fire hydrant. \\
  \texttt{drawbench\_079} & Descriptions & A device consisting of a circular canopy of cloth on a folding metal frame supported by a central rod, used as protection against rain or sometimes sun. \\
  \texttt{drawbench\_027} & Conflicting & A bird scaring a scarecrow. \\
\midrule
  \multicolumn{3}{l}{\textit{\cref{fig:qual_flux_6x}}} \\
  \texttt{drawbench\_098} & Gary Marcus et~al. & An elephant is behind a tree. You can see the trunk on one side and the back legs on the other. \\
  \texttt{drawbench\_100} & Gary Marcus et~al. & A pear cut into seven pieces arranged in a ring. \\
  \texttt{drawbench\_128} & Positional & A carrot on the left of a broccoli. \\
  \texttt{drawbench\_000} & Colors & A red colored car. \\
  \texttt{drawbench\_012} & Colors & A pink colored giraffe. \\
  \texttt{drawbench\_016} & Colors & A blue bird and a brown bear. \\
  \texttt{drawbench\_038} & Counting & Four cars on the street. \\
  \texttt{drawbench\_082} & Descriptions & A mechanical or electrical device for measuring time. \\
\midrule
  \multicolumn{3}{l}{\textit{\cref{fig:supp_zoom_flux}}} \\
  \texttt{drawbench\_061} & DALL-E & A single clock is sitting on a table. \\
  \texttt{drawbench\_049} & Counting & Two cats and two dogs sitting on the grass. \\
\midrule
  \multicolumn{3}{l}{\textit{\cref{fig:qual_ablation_pianist}}} \\
  \texttt{dpg\_whoops26} & Dense (DPG-Bench) & A man with bright red boxing gloves awkwardly attempts to play a glossy black grand piano in the center of a room. Around him, the floor is a polished hardwood, reflecting the soft overhead lighting. Despite the gloves, he appears concentrated, engaged in his unique challenge, with sheet music propped up on the piano's stand. \\
\bottomrule
\end{tabular}
\end{table}

\begin{table}[ht]
\centering
\small
\setlength{\tabcolsep}{5pt}
\renewcommand{\arraystretch}{1.15}
\caption{\textbf{Prompts of the video qualitative figures, verbatim.}}
\label{tab:prompts_video}
\begin{tabular}{lp{0.3\linewidth}p{0.4\linewidth}}
\toprule
  Prompt id & VBench dimension & Text \\
\midrule
  \multicolumn{3}{l}{\textit{\cref{fig:grid_hunyuan_main}}} \\
  \texttt{vbad077} & multiple\_objects & a dog and a horse \\
  \texttt{vbad687} & overall\_consistency, aesthetic\_quality, imaging\_quality & An astronaut feeding ducks on a sunny afternoon, reflection from the water. \\
\midrule
  \multicolumn{3}{l}{\textit{\cref{fig:supp_grid_hunyuan_5x}}} \\
  \texttt{vbad168} & human\_action & A person is sweeping floor \\
  \texttt{vbad351} & object\_class & a zebra \\
\midrule
  \multicolumn{3}{l}{\textit{\cref{fig:grid_wan}}} \\
  \texttt{vbad015} & temporal\_flickering & A tranquil tableau of palace \\
\midrule
  \multicolumn{3}{l}{\textit{\cref{fig:supp_grid_hunyuan}}} \\
  \texttt{vbad165} & human\_action & A person is playing harp \\
  \texttt{vbad220} & human\_action & A person is grooming dog \\
\midrule
  \multicolumn{3}{l}{\textit{\cref{fig:supp_grid_hunyuan720}}} \\
  \texttt{vbad623} & temporal\_style & A cute happy Corgi playing in park, sunset, in super slow motion \\
\midrule
  \multicolumn{3}{l}{\textit{\cref{fig:psnr_lies_crosswalk}}} \\
  \texttt{vbad799} & scene, background\_consistency & crosswalk \\
\bottomrule
\end{tabular}
\end{table}

\end{document}